\documentclass{article}
\pdfoutput=1

\usepackage{arxiv}

\usepackage[utf8]{inputenc} 
\usepackage[T1]{fontenc}    
\usepackage{hyperref}       
\usepackage{url}            
\usepackage{booktabs}       
\usepackage{amsfonts}       
\usepackage{nicefrac}       
\usepackage{microtype}      

\usepackage{amsmath,amssymb,amsfonts}
\usepackage{algorithmic}
\usepackage{graphicx}
\usepackage{textcomp}
\usepackage{tabularx, booktabs} 
\usepackage{multirow}
\usepackage{upgreek}
\usepackage[caption]{subfig}
\usepackage[linesnumbered,ruled,vlined]{algorithm2e}

\usepackage{color}

\title{Radial basis function kernel optimization for Support Vector Machine classifiers\thanks{This work has been submitted to the IEEE for possible publication. Copyright may be transferred without notice, after which this version may no longer be accessible}}

\author{%
  Karl~Thurnhofer-Hemsi, Ezequiel~L\'opez-Rubio, Miguel~A.~Molina-Cabello\\
  Department of Computer Languages and Computer Science \\
  University of M\'alaga \\
  Bulevar Louis Pasteur, 35, 29071 M\'alaga, Spain \\
  Biomedic Research Institute of M\'alaga (IBIMA) \\
  C/ Doctor Miguel D\'iaz Recio, 28, 29010, M\'alaga, Spain \\
  \texttt{\{karlkhader,ezeqlr,miguelangel\}@lcc.uma.es} \\
  \And
  Kayvan~Najarian \\
  Department of Computational Medicine and Bioinformatics \\
  University of Michigan \\
  Ann Arbor, MI, USA \\
  \texttt{kayvan@med.umich.edu}
}

\begin{document}

\maketitle

\begin{abstract}
Support Vector Machines (SVMs) are still one of the most popular and precise classifiers. The Radial Basis Function (RBF) kernel has been used in SVMs to separate among classes with considerable success. However, there is an intrinsic dependence on the initial value of the kernel hyperparameter. In this work, we propose OKSVM, an algorithm that automatically learns the RBF kernel hyperparameter and adjusts the SVM weights simultaneously. The proposed optimization technique is based on a gradient descent method. We analyze the performance of our approach with respect to the classical SVM for classification on synthetic and real data. Experimental results show that OKSVM performs better irrespective of the initial values of the RBF hyperparameter.


\end{abstract}

\section{Introduction}

Since the inception of SVMs \cite{Cortes1995273}, the interest for this kind of supervised learning method has only grown over the years \cite{Cristianini2000}, so that it has become a well established tool both for classification and regression \cite{Brereton2010230}. SVMs are regarded as the most prominent exemplar of kernel methods, which solve complex machine learning problems by using linear estimation methods on a high dimensional feature space \cite{Hofmann20081171}. They are intensely employed in a myriad of applications, including object segmentation \cite{Xing2016234}, video surveillance \cite{Zhao201529}, drug discovery \cite{Heikamp201493}, and cancer genomics \cite{Huang201841}.

The SVM framework models a classification problem as a maximum margin optimization problem, where the decision boundary that has the largest distance (margin) to separate the training points of different classes is searched. There is a primal form of the optimization problem, where the weights to be optimized are associated with the input features, i.e., there is one weight per each input feature. There is also a dual form, where the weights are associated with the training samples, i.e., one weight per each training sample. In the dual form, the weights are Lagrange multipliers of a suitable Lagrangian function. The fewer variables to be optimized, the easier the optimization problem, so dual formulations are preferred for classification tasks with many input features \cite{Chauhan2019803}.

The values of the kernel hyperparameters of SVMs are typically determined by cross validation on a grid of candidate values \cite{VanGestel20045}. This is a crude procedure that can only yield a rough approximation of the optimal value since there is no provision to fine tune the results. In other words, this is a case of uninformed search, since the information from previous trials of possible kernel hyperparameters is not employed to boost the search. In particular, the correct tuning of the spread hyperparameter $\gamma$ of the RBF kernel is essential for SVM classification performance \cite{Wang2003643}.

Optimization of general Gaussian kernels for SVMs has been proposed \cite{LAANAYA20111511,Wang20151045}, although our attention is focused on RBF kernels here. It must be noted that RBF kernels can be regarded as a restricted version of general Gaussian ones, where the Gaussian matrix is constrained to be the unit matrix multiplied by a scale factor. Learning the kernel matrix is also possible \cite{Lanckriet2004} even for non positive semidefinite kernel matrices \cite{Luss200997}. The optimization of the RBF kernel hyperparameter has been previously done by minimizing an upper bound of the leave one out error \cite{Keerthi1031955}, while our approach directly optimizes the Lagrangian function of the dual form. Other strategies to optimize the RBF kernel hyperparameter are based on the distances among training samples and the tightness of the decision boundary \cite{Xiao201475}.

In this work, we consider the dual form of the Support Vector Machine model. A new method is proposed to optimize the radial basis function kernel hyperparameter as well as the dual formulation Lagrange multipliers are adapted. This is attained by alternating two kinds of steps. First, the RBF kernel parameter is held fixed while the Lagrange multipliers are optimized, as in the standard SVM approach. Then the Lagrange multipliers are kept fixed while the RBF kernel hyperparameter is optimized. Both kinds of steps are carried out by gradient descent optimization. Adequate control of the length of the gradient descent steps ensures the stability of this optimization scheme.

The structure of this paper is as follows. First, the proposed methodology for SVM radial basis kernel optimization is detailed in Section \ref{sec:Methodology}. Then the experimental design and the obtained results are reported in Section \ref{sec:Results}. Finally, Section \ref{sec:Conclusions} is devoted to conclusions.

\section{Methodology} \label{sec:Methodology}

In this section, our RBF kernel optimization methodology is presented. Subsection \ref{subsec:Support-Vector-Machines} reviews some important concepts of SVMs, while the derivation of our proposal is done in Subsection \ref{subsec:RBF-kernel-optimization}. Finally, Subsection \ref{subsec:Computational-implementation} explains some computational considerations that are fundamental to the successful implementation of our method.


\subsection{Support Vector Machines}\label{subsec:Support-Vector-Machines}

Support Vector Machines were initially developed for binary classification problems. Given a set of training patterns $\mathcal{T}=\{\mathbf{x}_i\in\mathbb{R}^n,\,i=1,\dots,N\}$, and their corresponding labels from two classes $y_i\in\{-1,1\},\,i=1,\dots,N$, the classification problem is formulated as
$
    y(\mathbf{x})=\mathbf{w}^T\phi(\mathbf{x})+b
$,
where $\phi$ is the feature-space transformation function, and $b$ is the linear classification bias.

SVM searches the optimal hyper-plane that has a maximum margin between the nearest positive and negative samples. This search is expressed as:
\begin{equation}\label{eq:minSVM}
    \arg \min_{\mathbf{w},b}  \frac{1}{2}\left\lVert\mathbf{w}\right\rVert^2, \qquad 
    \text{subject to: } y_i(\mathbf{w}^T\phi(\mathbf{x})+b)\geq 1
\end{equation}
The introduction of the Lagrange multipliers $\boldsymbol{\alpha}=\{\alpha_i\}_{i=1,\dots,N}$ converts the problem (\ref{eq:minSVM}) into a maximization problem with respect to $\boldsymbol{\alpha}$, as explained in \cite{bishop2006pattern}. However, when the problem is very noisy, even known that kernels can represent non-linear decision boundaries, the problem may become very hard computationally to be solved. The best way to face hard problems consists on the introduction of control parameters that allow the violation of the margin constraints \cite{Cortes1995273}. This is called as the soft-margin problem, for which the optimization problem can be expressed using the following dual formulation:
\begin{equation}\label{eq:dualSVM}
    \max_{\boldsymbol{\alpha}} D_\gamma(\boldsymbol{\alpha}) = \sum_{i=1}^{N}{\alpha_i}-\frac{1}{2}\sum_{i=1}^{N}\sum_{j=1}^{N}{\alpha_i \alpha_j y_i y_j k_\gamma\left(\mathbf{x}_i,\mathbf{x}_j\right)}\quad
    \text{subject to:}\left\{
	       \begin{array}{ll}
    		 0\leq \alpha_i\leq C  & \forall\, i \\
    		 \sum_{i=1}y_i \alpha_i = 0 & \forall\, i
	       \end{array}
	     \right.
\end{equation}
where $k_\gamma$ denotes the radial basis function kernel:
\begin{equation}\label{eq:kernel}
    k_\gamma(\mathbf{x},\mathbf{y})=\exp\left(-\gamma||\mathbf{x}-\mathbf{y}||^2\right)
\end{equation}

The parameter $C$ acts as a regularization term, and it controls the allowed misclassification level for the training samples. Note that small values of $C$ makes the optimizer to look for an hyperplane with a large-margin separation, which may misclassify some points, and large values of $C$ will look for a smaller-margin to classify better all the training points. As shown in \cite{Lanckriet2004}, for any fixed kernel $k_\gamma$ the quantity $\max_{\boldsymbol{\alpha}} D_\gamma(\boldsymbol{\alpha})$ is an upper bound on misclassification probability.

After solving the classification problem, the computed multipliers $\alpha_i^*$ (and then $b^*$) allow the determination of the class of any other test sample $x\in\mathbb{R}^n$ by applying the function
\begin{equation}
    f(\mathbf{x})=\text{sign}\left(\sum_{i\in \mathcal{S}}y_i\alpha^*_i k_\gamma(\mathbf{x},\mathbf{x}_i)+b^*\right)
\end{equation}
where $\mathcal{S}$ is the set of the indices of the support vectors.

\subsection{RBF kernel optimization}\label{subsec:RBF-kernel-optimization}

In this work we propose a method that learns the $\gamma$  hyperparameter of the radial basis function kernel.
From the considerations mentioned in Subsection \ref{subsec:Support-Vector-Machines}, it follows that the quantity $\max_{\boldsymbol{\alpha}} D_\gamma(\boldsymbol{\alpha})$ is an upper bound on misclassification probability. In light of this, we propose to find an optimal value for the kernel hyperparameter $\gamma$ by minimizing $\max_{\boldsymbol{\alpha}} D_\gamma(\boldsymbol{\alpha})$ with respect to $\gamma$. Thus, $D_\gamma(\boldsymbol{\alpha})$ has to be maximized with respect to $\boldsymbol{\alpha}$, but also be minimized with respect to the kernel hyperparameter $\gamma$, subject to the same constrains defined in (\ref{eq:dualSVM}).

Therefore, the problem is stated as the following double optimization:
\begin{equation}\label{eq:dualSVM_gamma}
    \min_\gamma\left(\max_{\boldsymbol{\alpha}} D_\gamma(\boldsymbol{\alpha})\right) \quad
    \text{subject to:}\left\{
	       \begin{array}{ll}
    		 0\leq \alpha_i\leq C  & \forall\, i=1,\dots,N \\
    		 \sum_{i=1}y_i \alpha_i = 0 & \forall\, i=1,\dots,N
	       \end{array}
	     \right.
\end{equation}
We propose that the minimization with respect to $\gamma$ is carried out by the gradient descent method, where the quantity $D_\gamma(\boldsymbol{\alpha})$ for the current value of $\boldsymbol{\alpha}$ is taken as a suitable approximation of $\max_{\boldsymbol{\alpha}} D_\gamma(\boldsymbol{\alpha})$. Therefore, traditional steps of maximization with respect to $\boldsymbol{\alpha}$ are interleaved with steps of minimization with respect to $\gamma$. This interleaving is necessary since the restriction to find the best $\gamma$ value that minimizes $D_\gamma(\boldsymbol{\alpha})$ at each step would cause jumps in the minimization process when the said value is recalculated, causing non-convergence. Thus, the calculation of the gradient of $D_\gamma$ with respect to $\gamma$ is needed for the update step of the gradient descent method:
\begin{equation}\label{eq:gradient}
    \begin{split}
    \frac{\partial D_\gamma(\boldsymbol{\alpha})}{\partial \gamma}=\frac{1}{2}\sum_{i=1}^{N}\sum_{j=1}^{N}{\alpha_i \alpha_j y_i y_j ||\mathbf{x}_i-\mathbf{x}_j||^2 \exp\left(-\gamma||\mathbf{x}_i-\mathbf{x}_j||^2\right)}\\[-5pt]
    =\frac{1}{2}\sum_{i=1}^{N}\sum_{j=1}^{N}{\alpha_i \alpha_j y_i y_j ||\mathbf{x}_i-\mathbf{x}_j||^2k_\gamma(\mathbf{x}_i,\mathbf{x}_j)}
\end{split}
\end{equation}

Therefore, the update step of the gradient descent adaptation rule is formulated as:
\begin{equation}\label{eq:GDrule}
        \gamma_{t+1}=\gamma_t - \eta \frac{\partial D_{\gamma_t}(\boldsymbol{\alpha})}{\partial \gamma_t}
    \end{equation}
where $t$ is the time step and $\eta$ is the learning rate. A detailed description of the method, named as \textit{OKSVM}, is presented in Algorithm \ref{algo:OKSVM}. Note that the regularization term $C$ is an input hyperparameter and it is not optimized by our algorithm.

The learning rate parameter $\eta$ of the gradient descend method is adapted following the well-known bold driver adaptation method:
\begin{itemize}
    \item If the minimum was not reached, then the learning rate is increased by a factor $\zeta^+$: $\eta=\zeta^+\eta$.
    \item Else, if the minimum was reached but not the convergence of the algorithm, or the actual $\eta$ is too high, then the learning rate is decreased by a factor $\zeta^-$: $\eta=\zeta^-\eta$.
\end{itemize}
The choice of the update factors are determined using a validation set and they were fixed to $\zeta^+=1.01$ and $\zeta^-=0.1$. Likewise, the initial value of the learning rate was set to $\eta=0.01$.

\IncMargin{1em}
\begin{algorithm}
\SetCommentSty{textit}
\DontPrintSemicolon 
\KwIn{The training data $\{(x_i,y_i),\,i=1,\dots,N\}$, the initial  $\gamma_0$, and the chosen value of $C$}
\KwOut{The trained SVM model}
$\gamma_f \gets \gamma_0, \quad$
$opt \gets \text{false}, \quad$
$t \gets 0$\;
Compute the initial Lagrange multipliers $\boldsymbol{\alpha}^0$ and the cost function $D_{\gamma_0}(\boldsymbol{\alpha}^0)$ associated to the initial kernel $k_{\gamma_0}$ using Eq. (\ref{eq:dualSVM})\;
\Repeat{$opt$ \textbf{ or } $\gamma_{t+1} > \gamma_{MAX}$ \textbf{ or } $WS=5$}{
    $\gamma_{t+1} \gets \gamma_t - \eta \frac{\partial D_{\gamma_t}(\boldsymbol{\alpha})}{\partial \gamma_t}$ \tcc*{gradient step using Eq. (\ref{eq:GDrule})}
    \uIf{$\gamma_{t+1}>0$}{
      Compute the new weights $\boldsymbol{\alpha}^{t+1}$ that maximize $D_{\gamma_{t+1}}(\boldsymbol{\alpha}^{t+1})$ using Eq. (\ref{eq:dualSVM})\;
      
        \uIf(\tcc*[f]{minimum reached}){$D_{\gamma_{t+1}}(\boldsymbol{\alpha}^{t+1})> D_{\gamma_{t}}(\boldsymbol{\alpha}^{t})$}{
            $\gamma_{t+1} \gets \gamma_t$\;
            \lIf(\tcc*[f]{check convergence}){$|\gamma_{t+1}-\gamma_{f}|<\varepsilon$}{$opt \gets \text{true}$}
            \lElse{
                $\gamma_{f} \gets \gamma_t, \quad$ 
                $\eta \gets \zeta^-\eta$
                }
        }
        \lElseIf(\tcc*[f]{stagnation condition}){$D_{\gamma_{t+1}}(\boldsymbol{\alpha}^{t+1})= D_{\gamma_{t}}(\boldsymbol{\alpha}^{t})$}{
            $WS \gets WS+1$
        }
        \lElse{
            $\eta \gets \zeta^+\eta, \quad$
            $WS \gets 0$
        }
    }
    \lElse{
      $\gamma_{f} \gets \gamma_t, \quad$
      $\eta \gets \zeta^-\eta$
    }
    $t \gets t+1$
}
\Return{$\boldsymbol{\alpha}^{t}$}\;
\caption{{\sc OKSVM}}
\label{algo:OKSVM}
\end{algorithm}
\DecMargin{1em}

\subsection{Computational implementation}\label{subsec:Computational-implementation}

In the experiments, it has been found that $D_\gamma(\boldsymbol{\alpha})$ sometimes exhibits horizontal asymptotes with respect to the $\gamma$ hyperparameter. This leads to many adjusts of $\gamma$ with tiny enhancements of $D_\gamma(\boldsymbol{\alpha})$. If left uncontrolled, this process results in numerical errors when the value of $\gamma$ grows too large. In order to avoid this situation step 4 of the algorithm is proposed, which stops the optimization whenever a horizontal asymptote has been detected. A safe value $\gamma_{MAX}$ is defined for this purpose, which ensures that no numerical errors arise if $\gamma$ is within the interval $\left[0,\gamma_{MAX}\right]$.

Two additional stagnation conditions are controlled by their respective checks. Line 11 of the algorithm deals with plateaus in the optimization landscape of $\gamma$. That is, if the value of the objective function $D_\gamma(\boldsymbol{\alpha})$ has not changed for the last five steps ($WS=5$), then the optimization procedure is halted. Line 9 is in charge of checking whether a vanishing gradient condition has been reached. In other words, if the gradient (\ref{eq:gradient}) is so small that the changes of $\gamma$ are insignificant, then the optimization is terminated. A tolerance $\varepsilon$ for $\gamma$ is defined accordingly.


The value of the mentioned control parameters are determined using a validation set and they were fixed to $\gamma_{MAX}=1000$ and $\varepsilon=0.01$.

\section{Experimental results} \label{sec:Results}
This section reports the experiments carried out. First of all, the methods are described in Subsection \ref{sec:Methods}. Then, the selected measures to compare the methods are detailed in Subsection \ref{sec:Measures}. After that, experiments with different synthetic and real datasets are reported in Subsections \ref{sec:Synthetic_datasets} and \ref{sec:Real_datasets}. More experimental results and the source code are attached as supplementary materials.

\subsection{Methods}\label{sec:Methods}

The proposed optimized kernel method is noted as \textit{OKSVM}. It has been developed in Python by using libraries such as the scikit-learn library, and it is based on an SVM that implements the soft-margin kernel SVM with the Sequential Minimal Optimisation (SMO) algorithm and includes explicit kernel functions \cite{Mathiew2018}, which adapts perfectly to our methodology. Then, \textit{OKSVM} was compared with this \textit{SVM} implementation. The reported experiments have been carried out on a 64-bit Personal Computer with an eight-core Intel i7 3.60 GHz CPU, 32 GB RAM, and an NVIDIA Titan X GPU.

Two different studies have been carried out to establish the influence of the hyperparameters $C$ and $\gamma$ in both \textit{SVM} and \textit{OKSVM} methods. The first experiment comprises the analysis of the performance when the hyperparameters $C$ and $\gamma$ are fixed with the same value for both methods. The second experiment optimizes the hyperparameters $C$ and $\gamma$ during a previous step by doing a grid search. 

\subsection{Measures}\label{sec:Measures}

In order to compare the performance of the methods, a well-known measure has been selected: the F1-score (noted as $F1$), which represents the percentage of hits of the system. Accuracy ($Acc$), Area Under Curve ($AUC$), Precision ($PR$) and Recall ($RC$) are also computed. All these measures provide values in the interval $[0,1]$, where the higher is better.
Although the Accuracy ($Acc$) is also a well-known and widely employed measure, it is invalidated since it could be biased in case the dataset is sensibly unbalanced. This way, $F1$ can give a better classifier's performance understanding.



Additionally, the performance difference of a method against the other has also been proposed as a measure. This measure (noted as $F1_\text{diff}$) has been computed as the difference of the performance of both considered methods and multiplied by a factor of 100 for the sake of clarity:
\begin{equation}
F1_\text{diff} = 100 \cdot (F1_\text{OKSVM} - F1_\text{SVM})
\end{equation}


Moreover, the wins-losses ratio (noted as \textit{wlr}) of \textit{OKSVM} against \textit{SVM} has also been computed as a measure to analyze in more detail the performance of both methods. Its definition is as follows:
\begin{equation}\label{eq:WinsLossesRatio}
wlr = 100\cdot \frac{1}{R}\sum_{i=1}^{R}{{wl}_{i}},
\qquad
wl=\begin{cases}
1 & \textrm{if }F1_\text{diff}>0\\
-1 & \textrm{if }F1_\text{diff}<0
\end{cases}
\end{equation}
\subsection{Synthetic datasets}\label{sec:Synthetic_datasets}

Synthetic datasets have been generated using the \textit{make\_classification} function of \textit{sklearn} package to determine the goodness of \textit{OKSVM} due to the wide range of properties of the datasets used in the comparison. This function allows us to generate a random binary classification problem with specific properties determined by several parameters. This way, the number of samples that compose each generated synthetic dataset has been fixed to 200. These samples are classified into two possible classes equally balanced. Besides, a fixed number of one cluster per class has been considered.

From the available options, the separation ($sep$) between classes has been tuned in order to test different classification problems where both classes can be linearly (larger values of $sep$) or not linearly (lower values of $sep$) classified. The tuned configurations for the separation parameter are $sep=\{0.6, 0.8, 1.0, 1.2, 1.4\}$. Lower values generate very overlapped datasets that are so complex to be well classified. In the same way, larger values produce datasets that can be easily classified. No noise was introduced in the datasets. Additionally, a different number of features (i.e., the dimension, noted as $dim$) for each sample of the dataset are also considered in this work. The tuned configurations for the number of features are $dim=\{2, 3, 4, 5, 6, 7, 8\}$. Furthermore, the tuned configurations of the parameters $C$ and $\gamma$ are $C=\{0.5, 1.0, 1.5\}$ and $\gamma=\{0.1, 0.5, 0.9, 1.3, 1.7, 2.1\}$, respectively. Experiments for each tuned configuration have been randomly repeated 100 times ($R=100$) using a testing set size of 50\%, and the average performance is used to establish the efficiency comparisons.



\subsubsection{Fixed hyperparameters comparison}

These experiments compare the performance of \textit{SVM} and \textit{OKSVM} methods when they have the same fixed hyperparameter configuration. \figurename\,\ref{fig:heatmap_Fm_comparison} exhibits the performance of each method. As can be observed, the higher the value of $\gamma$, the lower the performance of \textit{SVM} method, especially when the dimension $dim$ of the dataset is higher. Note that \textit{OKSVM} always achieves the best performance for a fixed value of the separation of the clusters in the dataset ($sep$), which means that the optimization of $\gamma$ was successful. Meanwhile, $C$ seems to do not have a high impact on performance. Regarding the effect of the dataset properties in the performance, the higher $sep$, the easier the classification task, as it was expected. Additionally, the higher the value of $dim$, the worse the performance of both methods. However, the \textit{OKSVM} performance is not deteriorated as strongly as \textit{SVM}.

\setlength\tabcolsep{1.5pt} 
\newcolumntype{C}{ >{\centering\arraybackslash} m{2.8cm} }
\newcolumntype{D}{ >{\centering\arraybackslash} m{1.5cm} }

\begin{figure}[t]
\centering
	\begin{tabular}{c|D|CCCc}
	    $dim$ & Method & $C=0.5$ & $C=1.0$ & $C=1.5$ \\
	    \midrule
		\multirow{2}{*}[-3em]{2} & OKSVM & \includegraphics[width=1\linewidth]{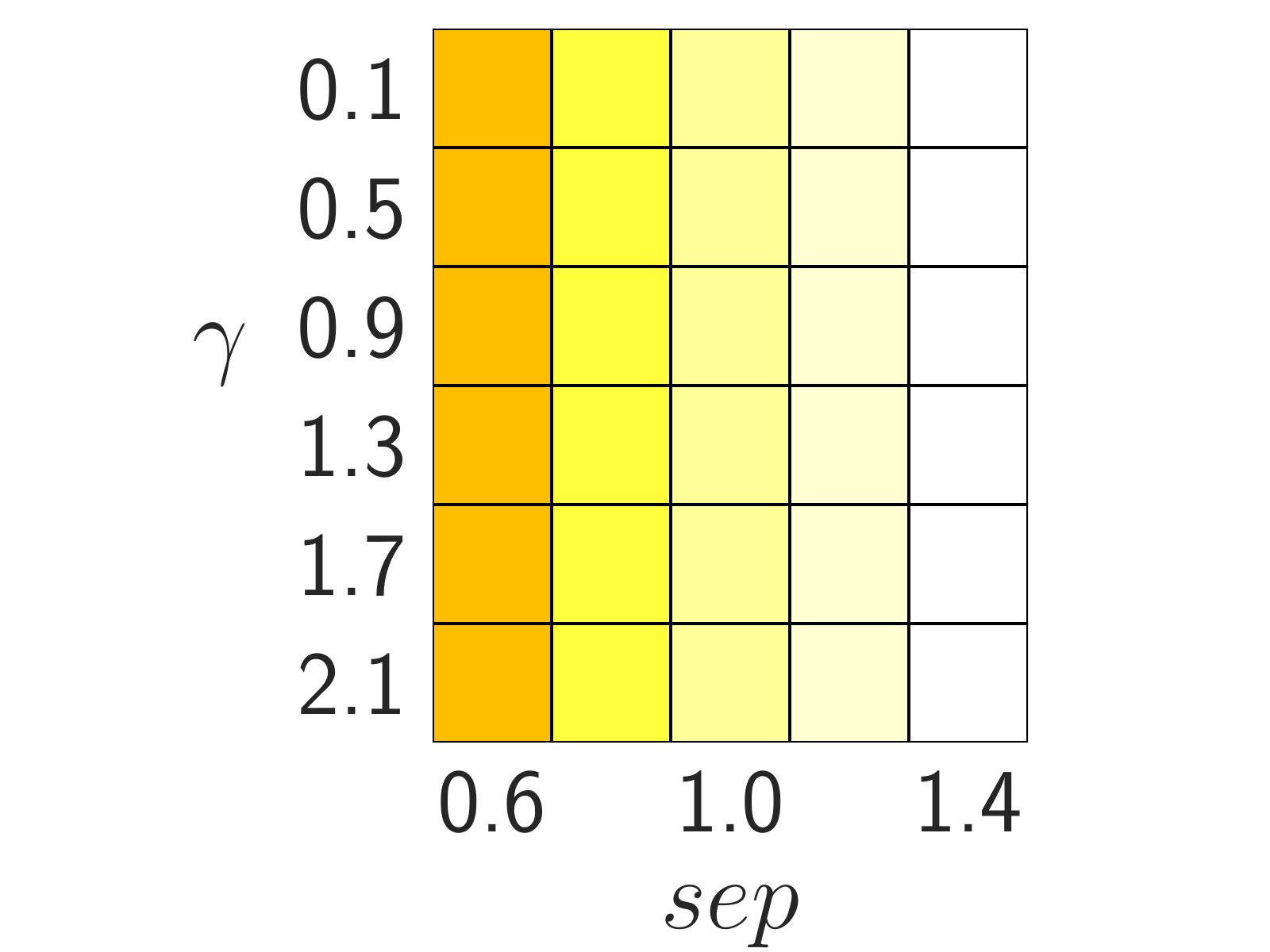} &
		\includegraphics[width=1\linewidth]{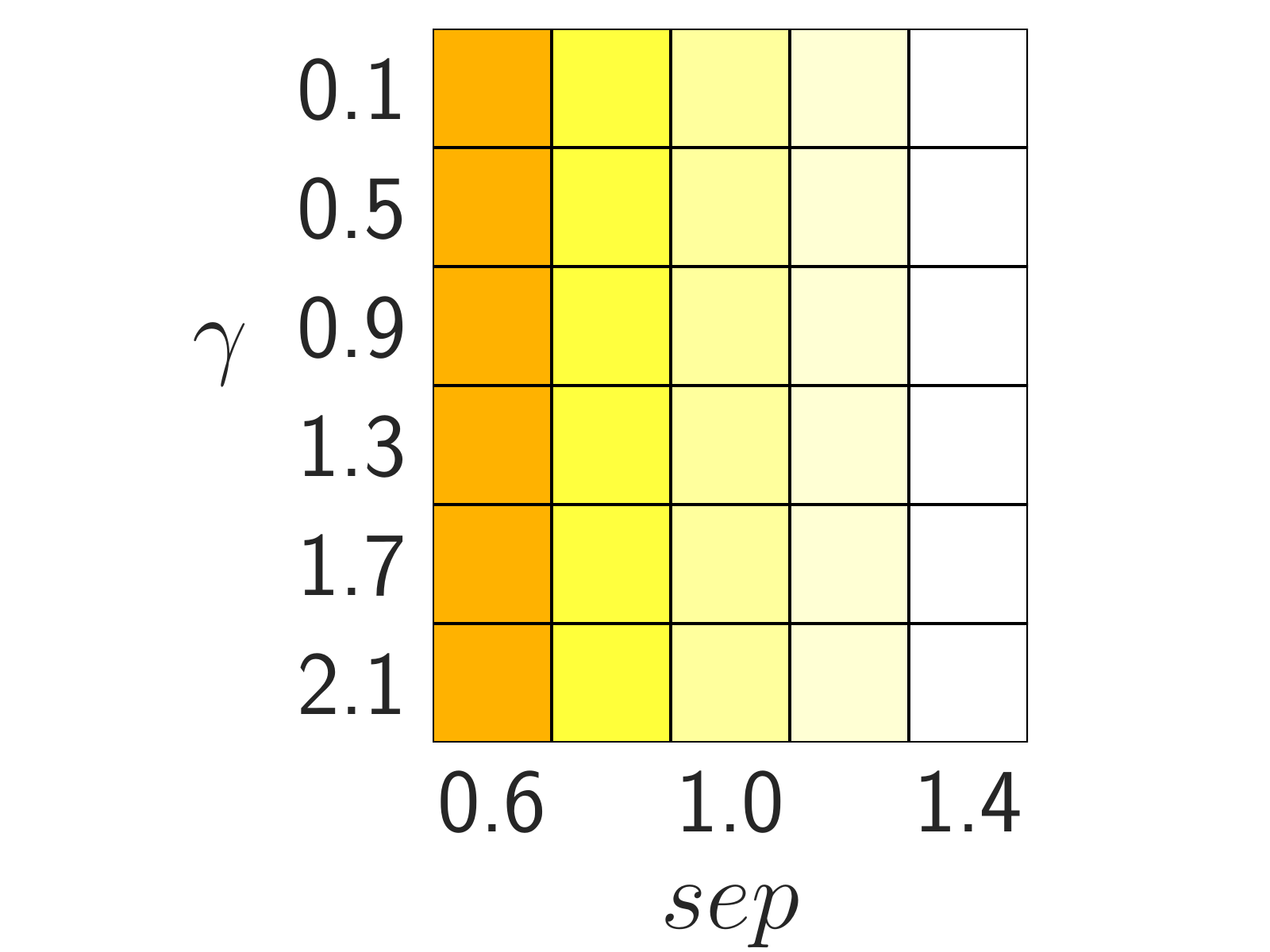} &
		\includegraphics[width=1\linewidth]{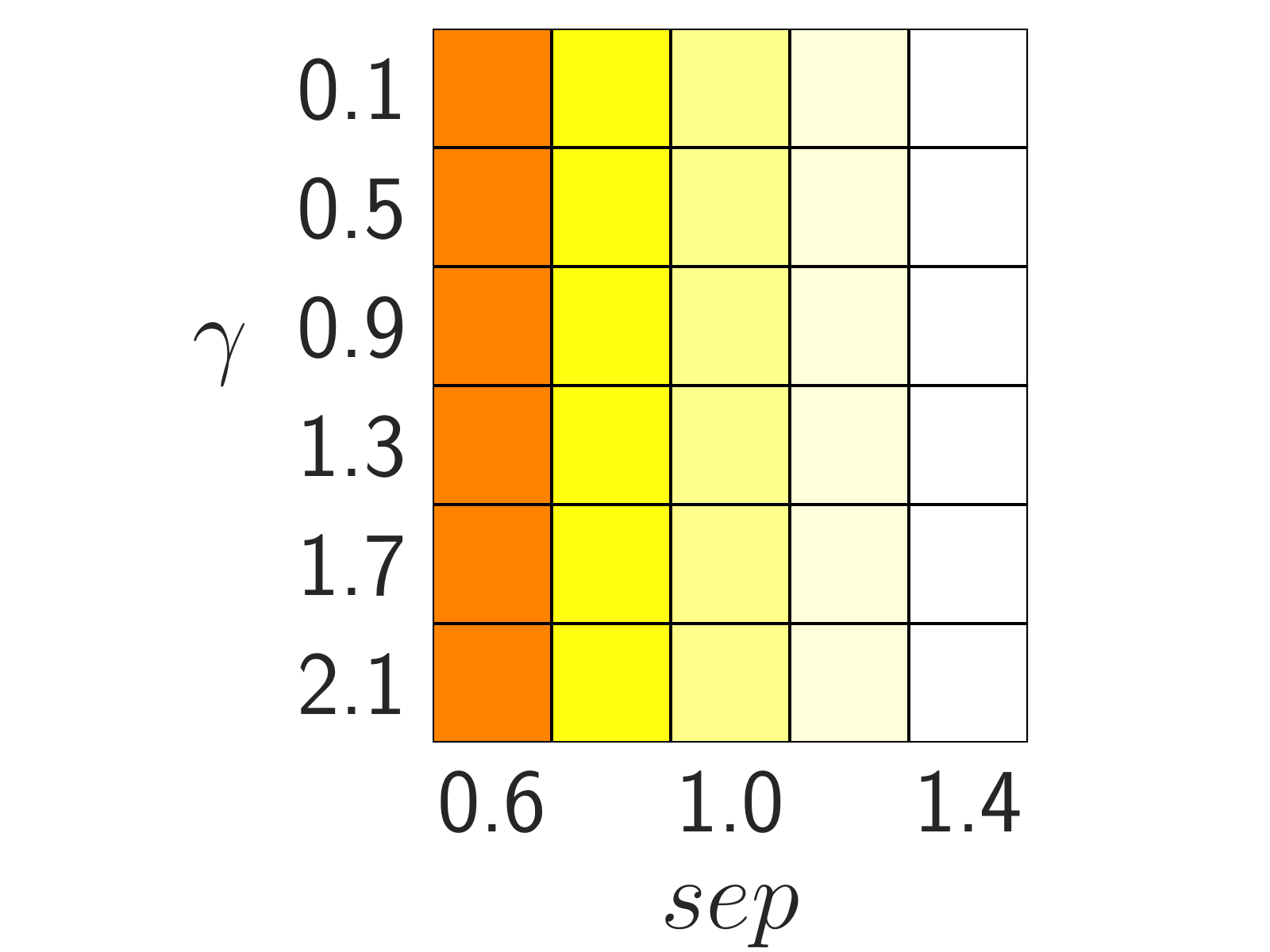} & \multirow{6}{*}[-5em]{\includegraphics[width=0.075\linewidth]{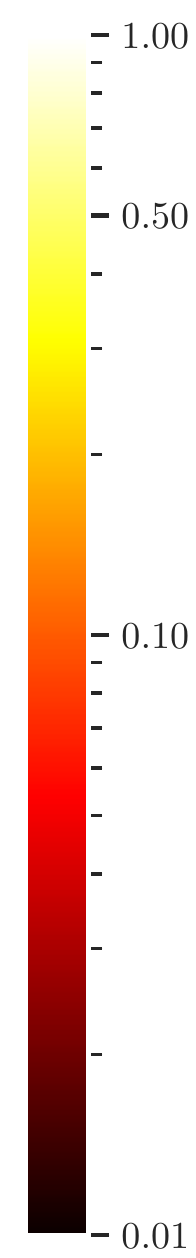}}\\
		& SVM  & \includegraphics[width=1\linewidth]{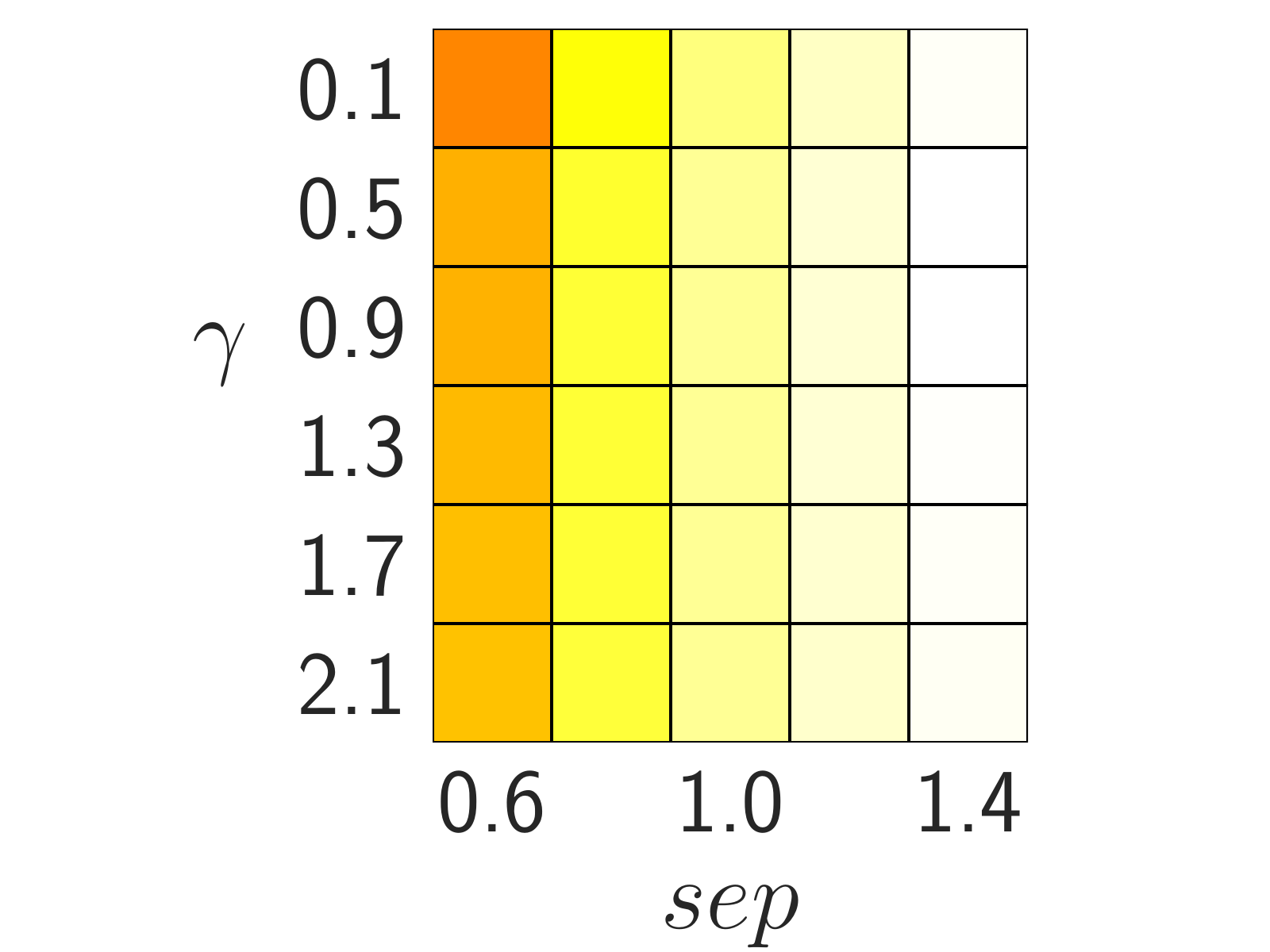} &
		\includegraphics[width=1\linewidth]{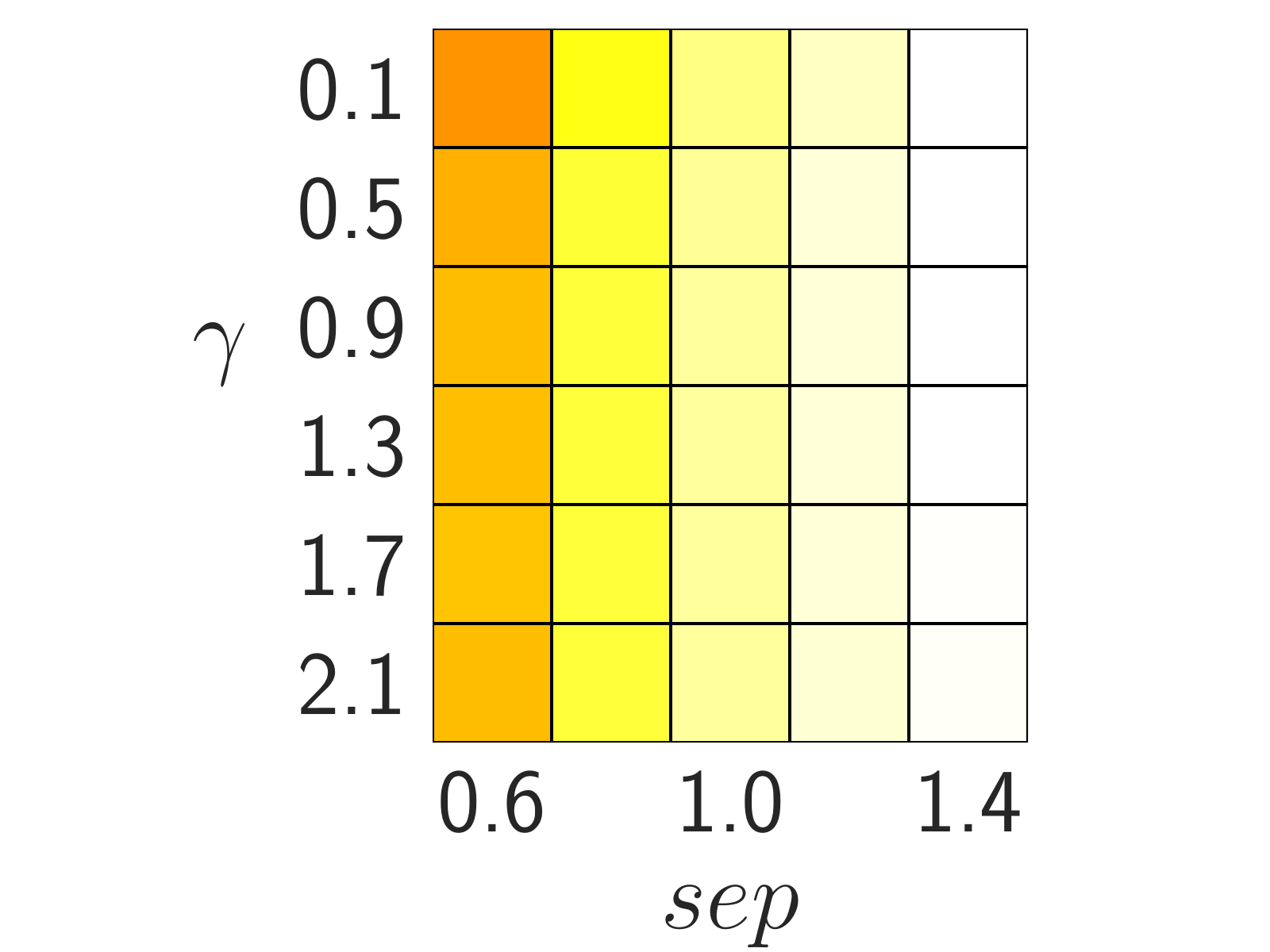} &
		\includegraphics[width=1\linewidth]{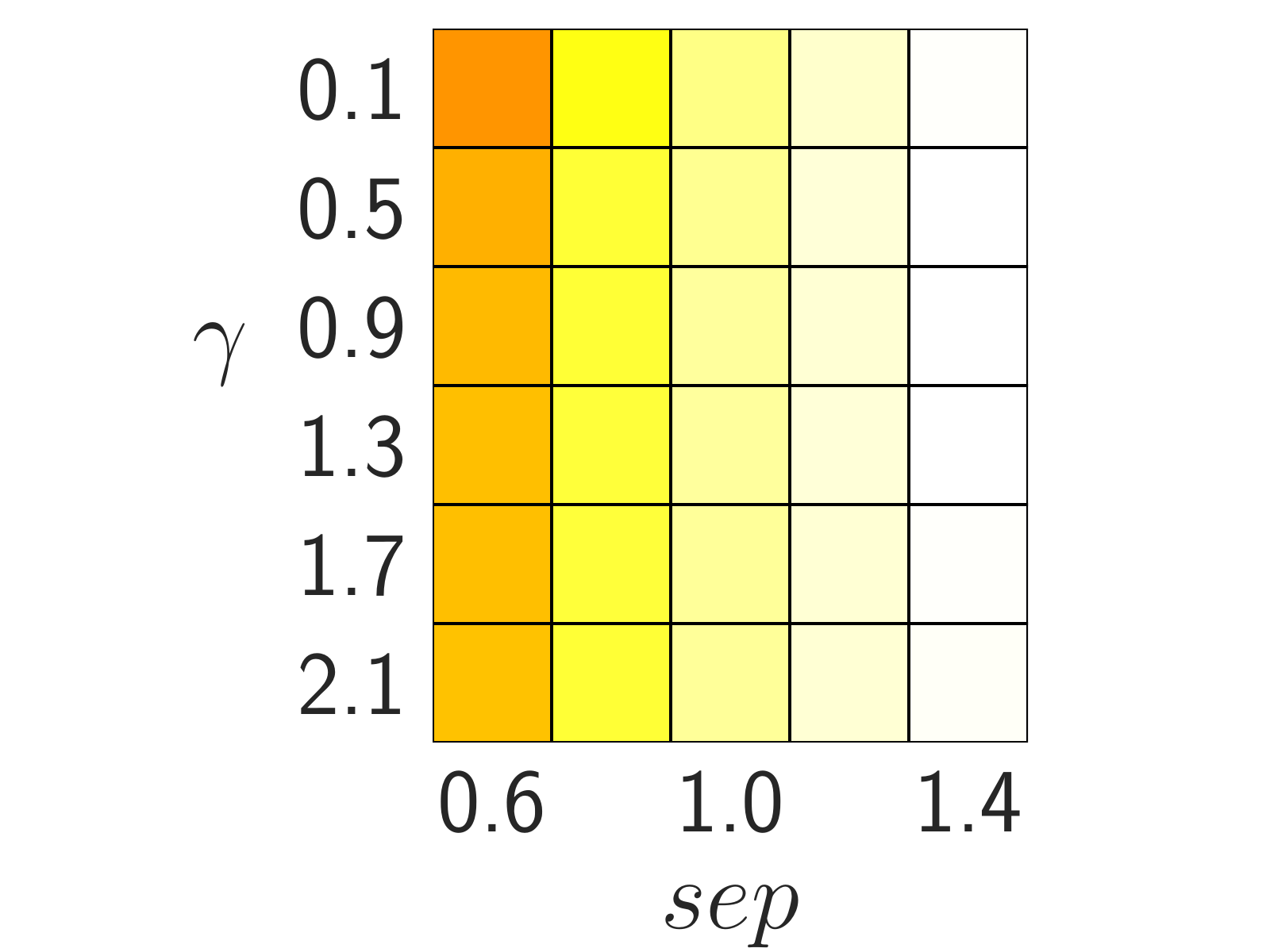} & \\
		\cmidrule{1-5}
		\multirow{2}{*}[-3em]{4} & OKSVM & \includegraphics[width=1\linewidth]{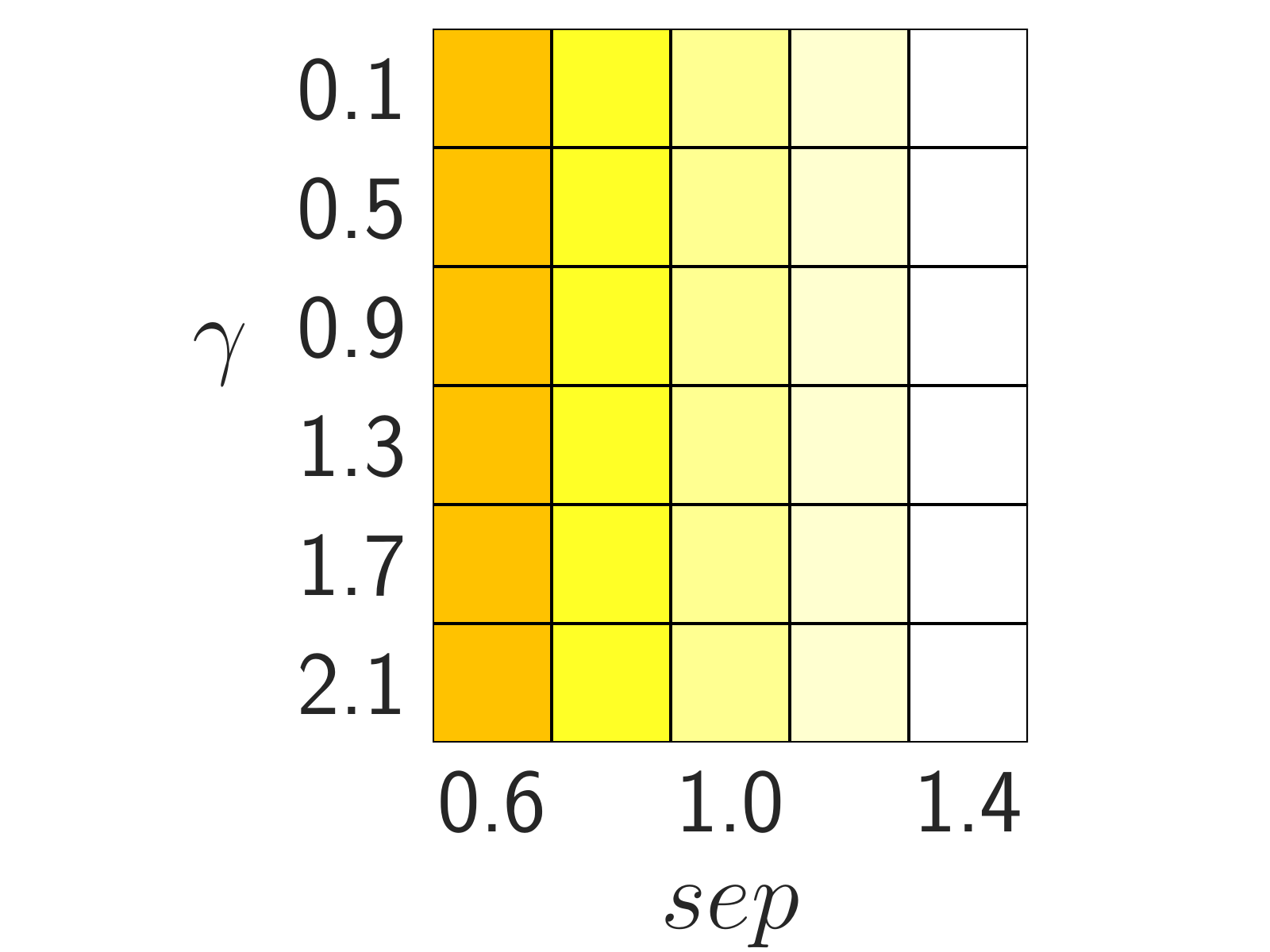} &
		\includegraphics[width=1\linewidth]{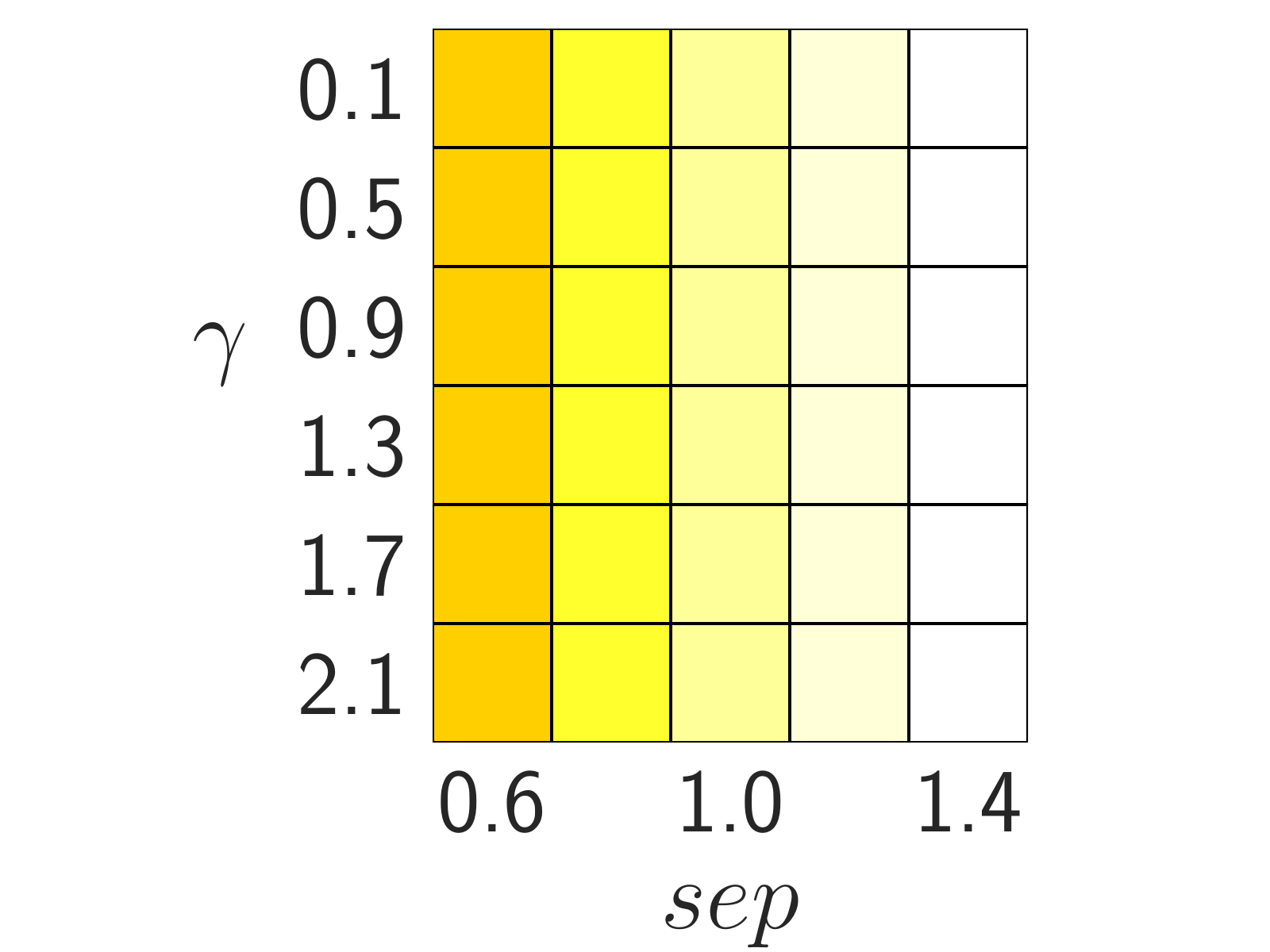} &
		\includegraphics[width=1\linewidth]{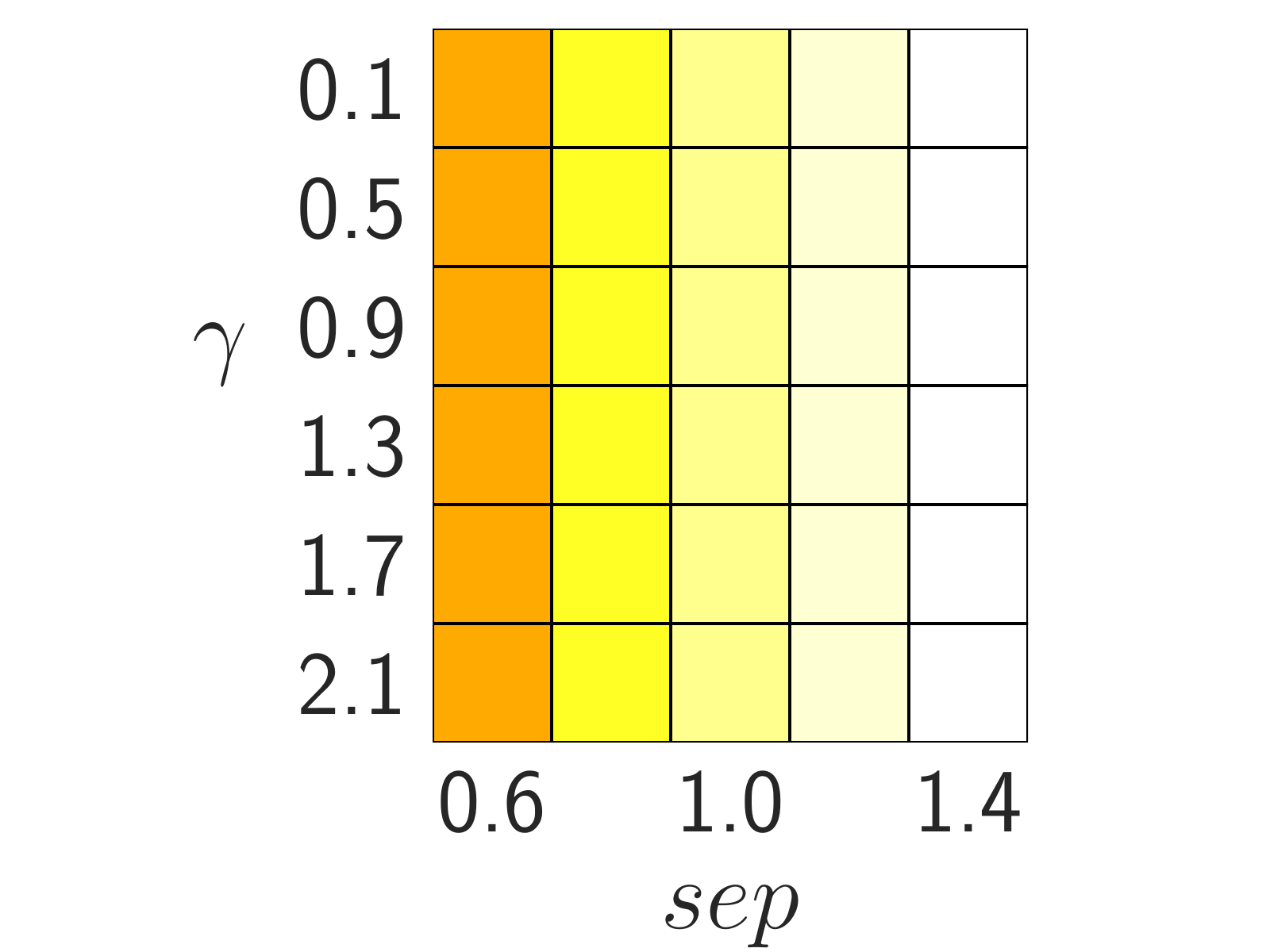} & \\
		 & SVM  & \includegraphics[width=1\linewidth]{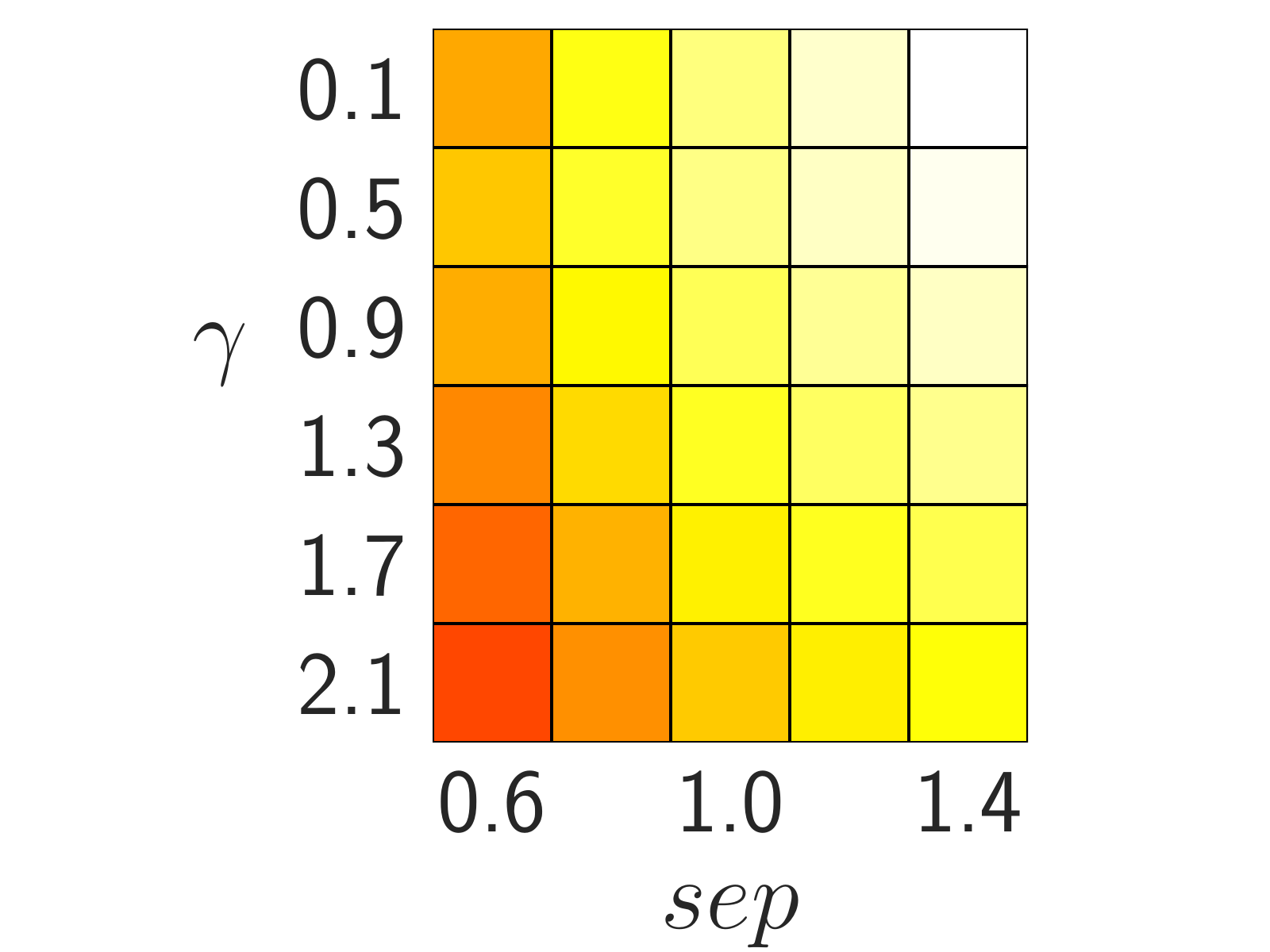} &
		\includegraphics[width=1\linewidth]{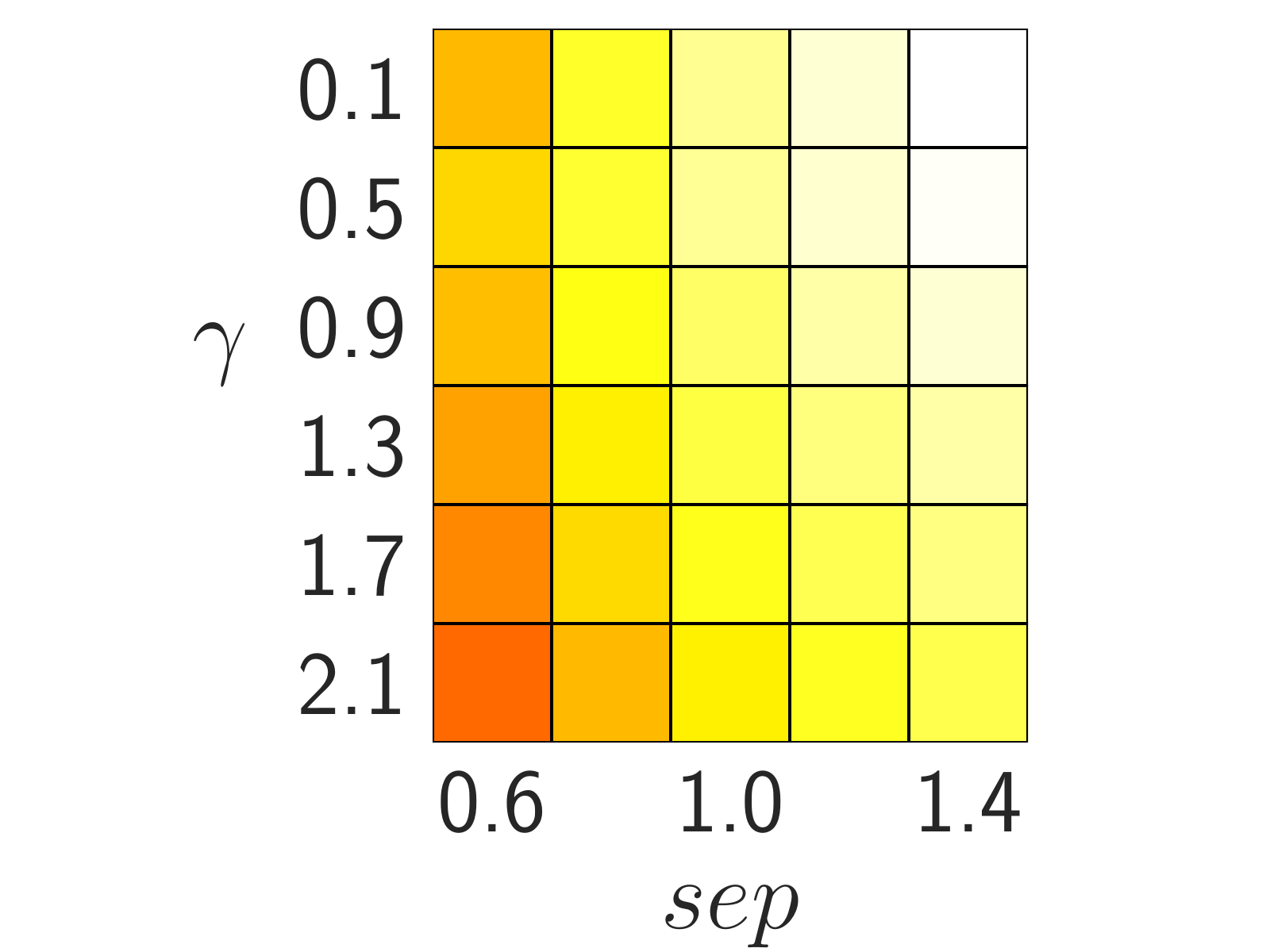} &
		\includegraphics[width=1\linewidth]{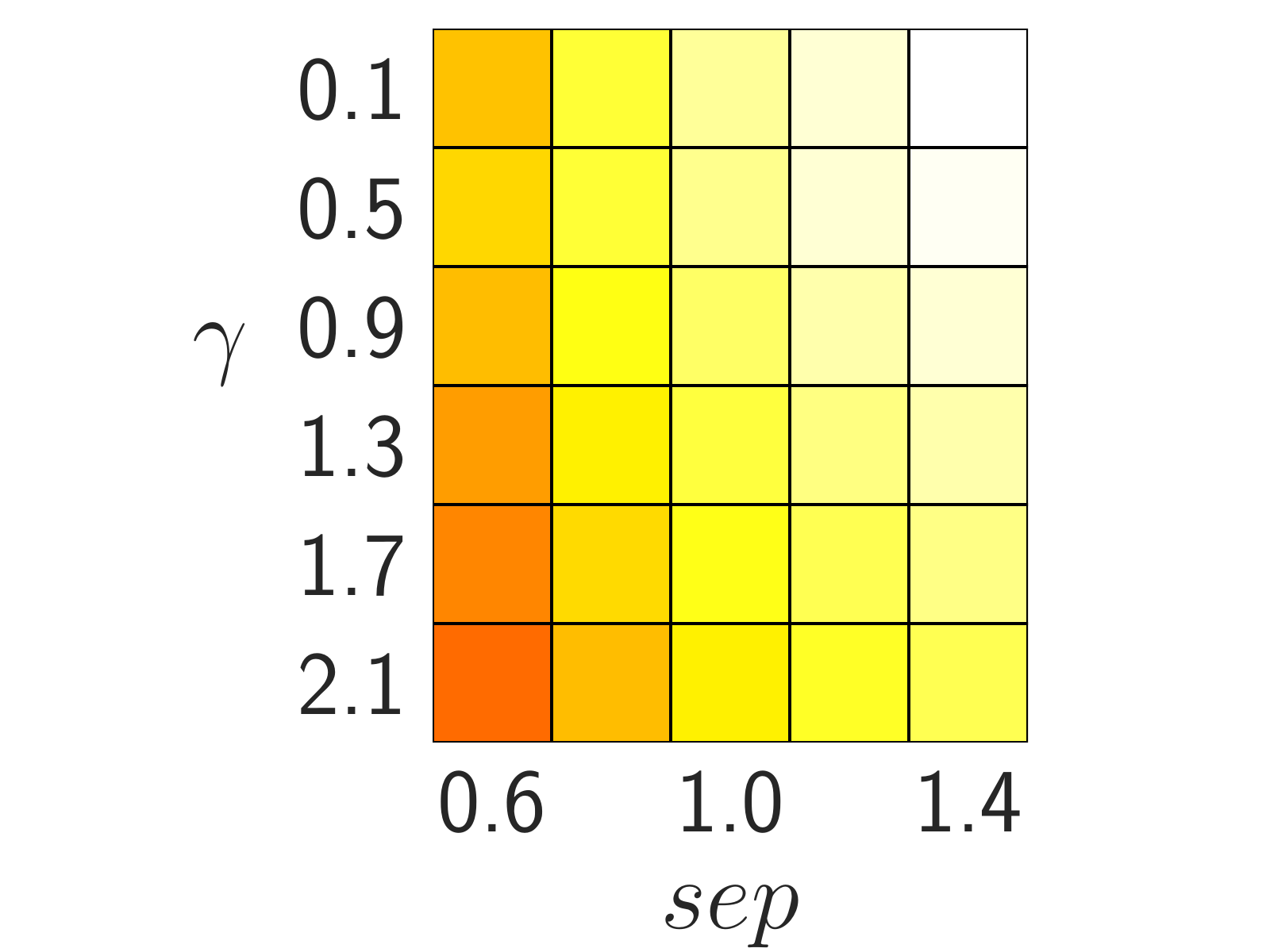} & \\
		\cmidrule{1-5}
		\multirow{2}{*}[-3em]{8} & OKSVM & \includegraphics[width=1\linewidth]{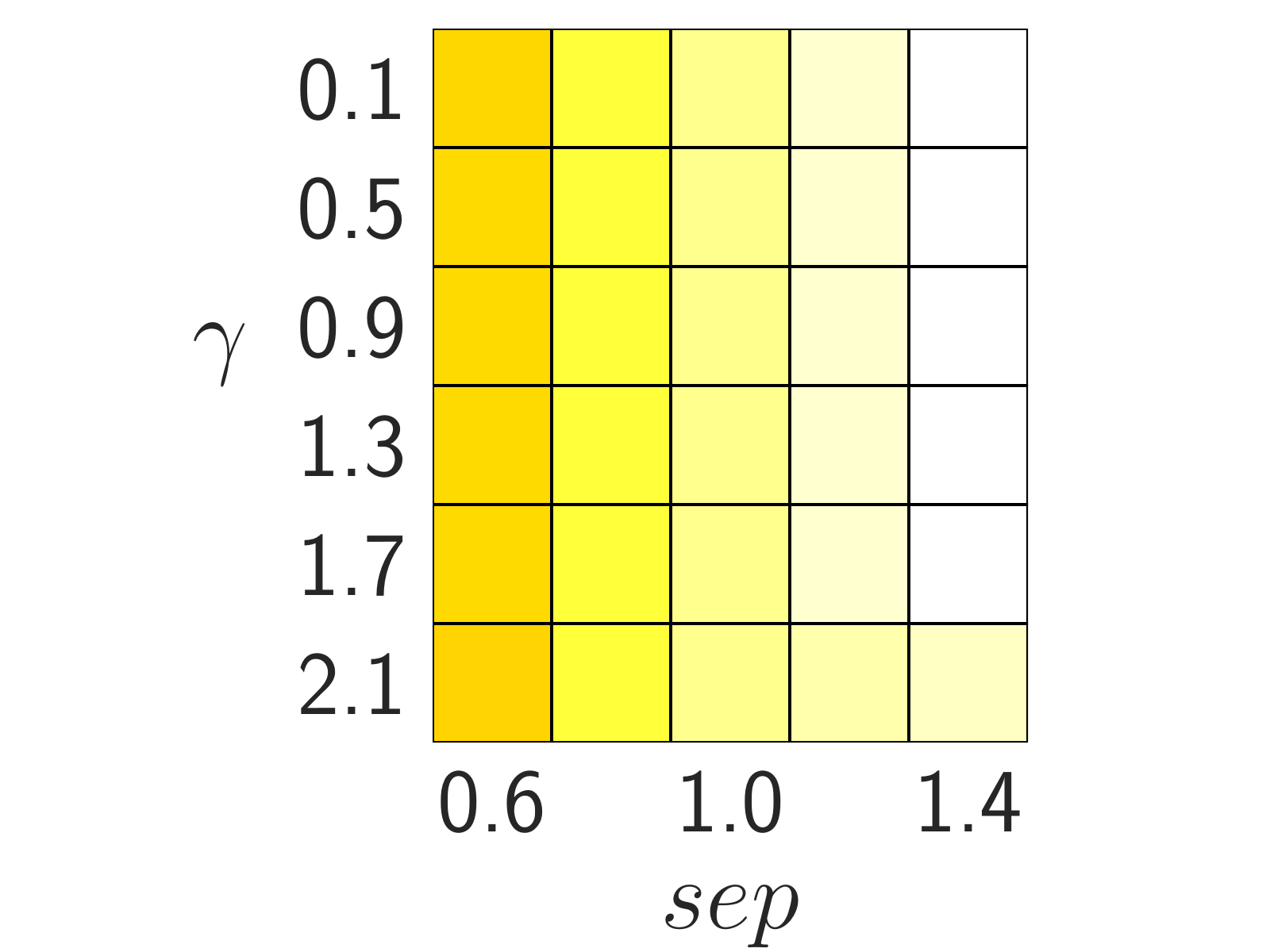} &
		\includegraphics[width=1\linewidth]{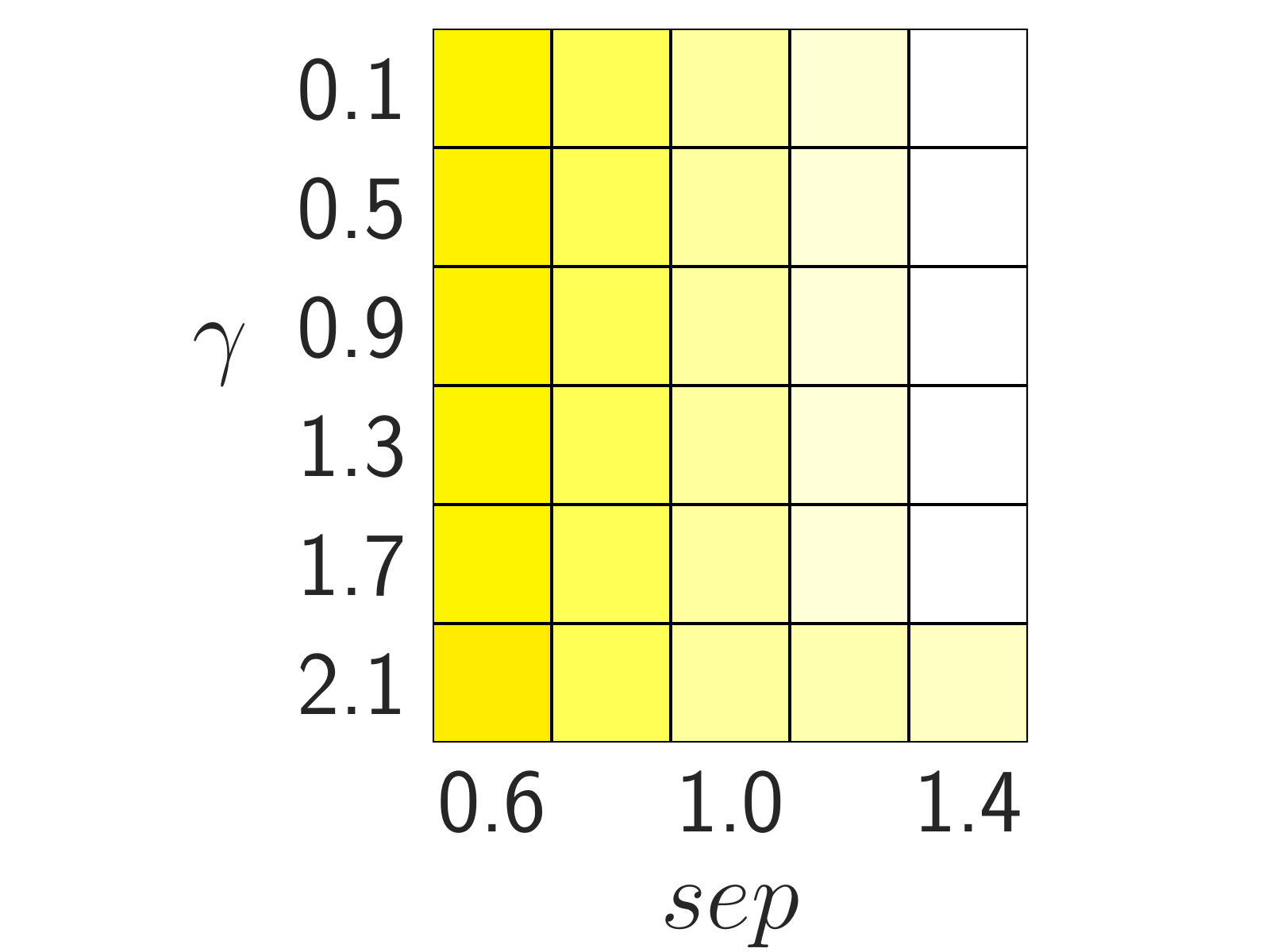} &
		\includegraphics[width=1\linewidth]{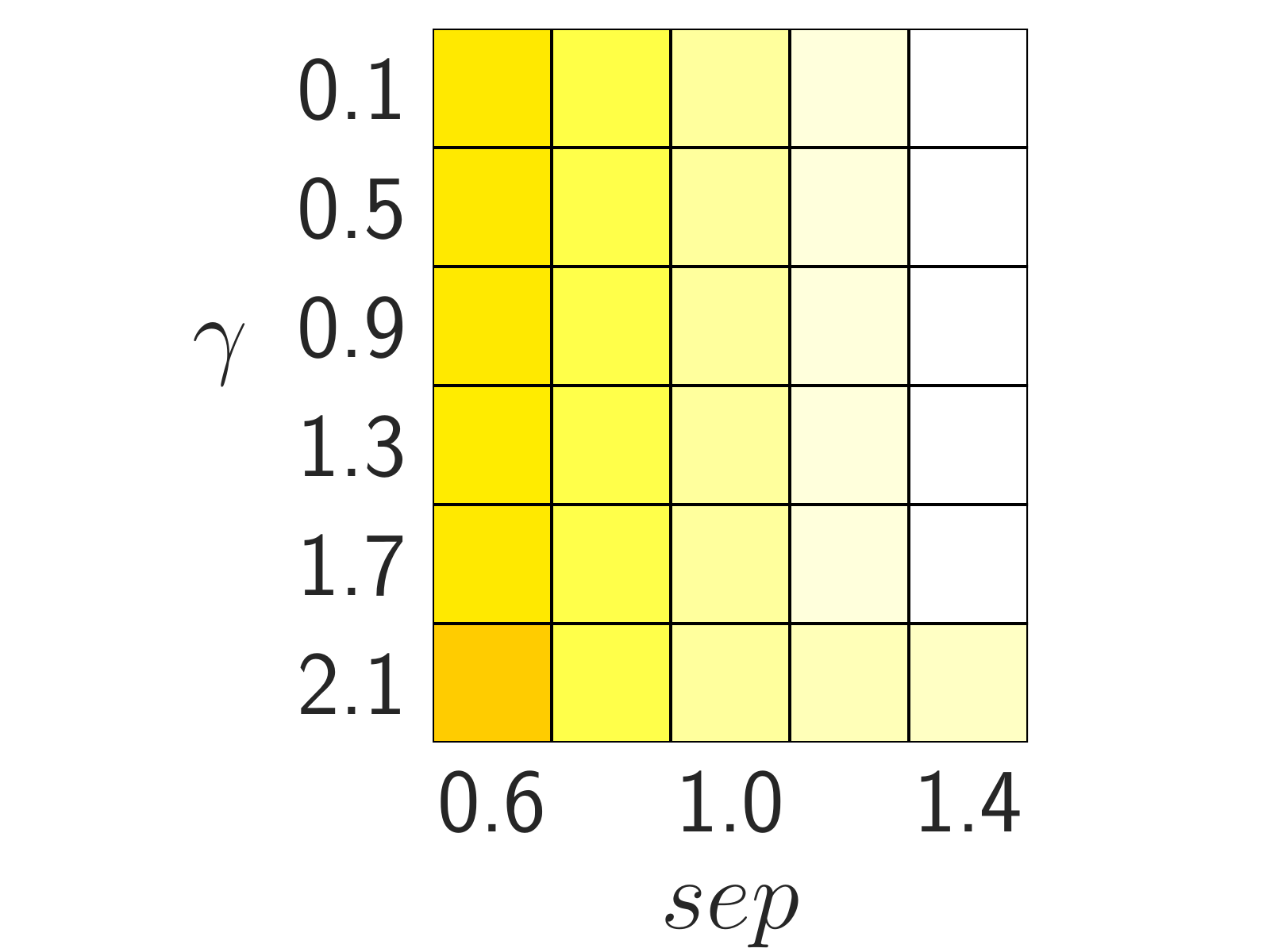} & \\
		 & SVM  & \includegraphics[width=1\linewidth]{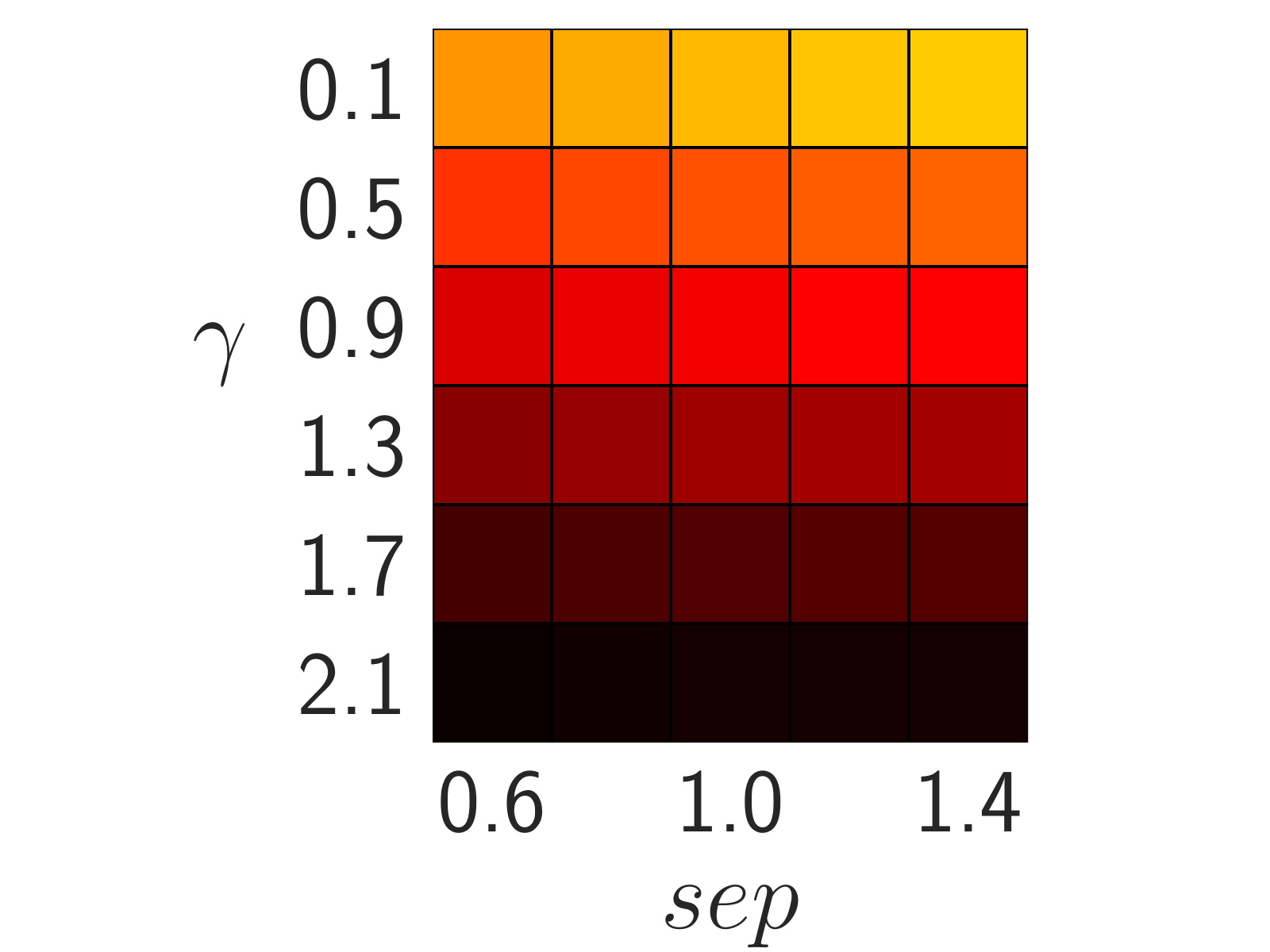} &
		\includegraphics[width=1\linewidth]{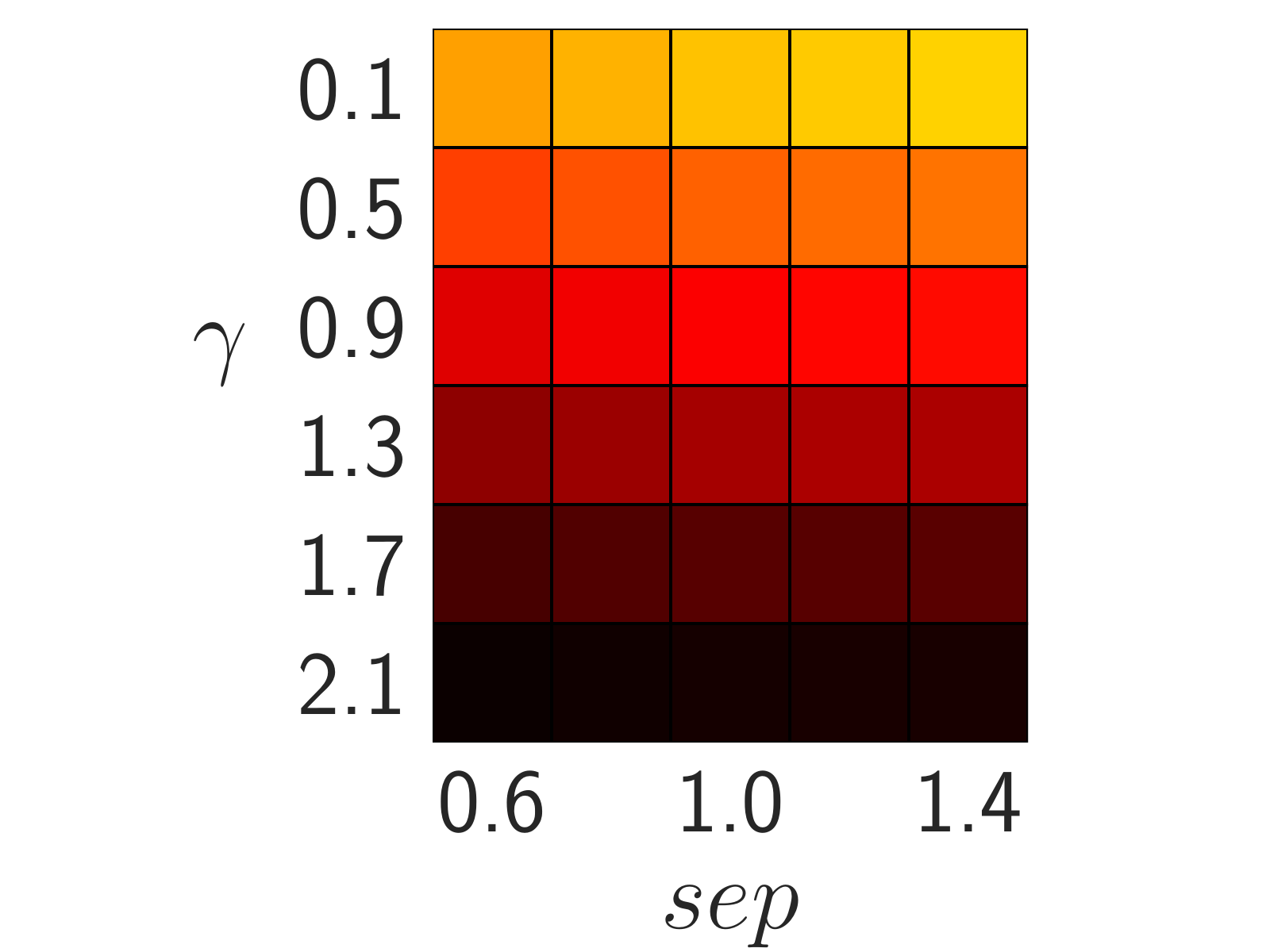} &
		\includegraphics[width=1\linewidth]{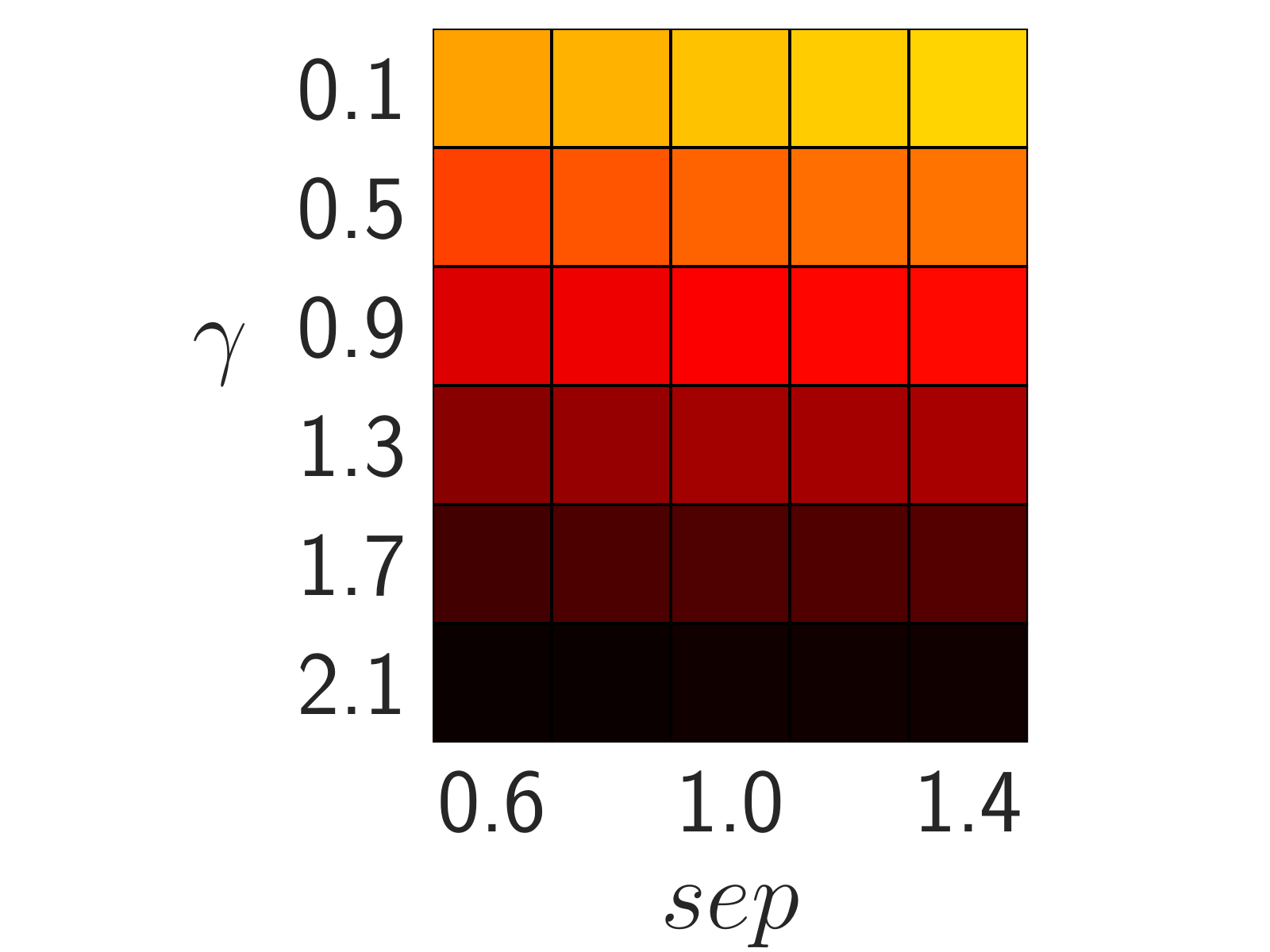} & \\
	\end{tabular}
\caption{$F1$ heatmaps of \textit{OKSVM} and \textit{SVM}. The average performance of 100 runs on the test set for each configuration is represented varying $\gamma$ and $sep$. Lighter tones are better.}
\label{fig:heatmap_Fm_comparison}
\end{figure}

\begin{figure}[t]
\centering
	\begin{tabular}{D|CCCc}
	    $dim$ & $C=0.5$ & $C=1.0$ & $C=1.5$ & \\
	    \midrule
		2 & \includegraphics[width=1\linewidth]{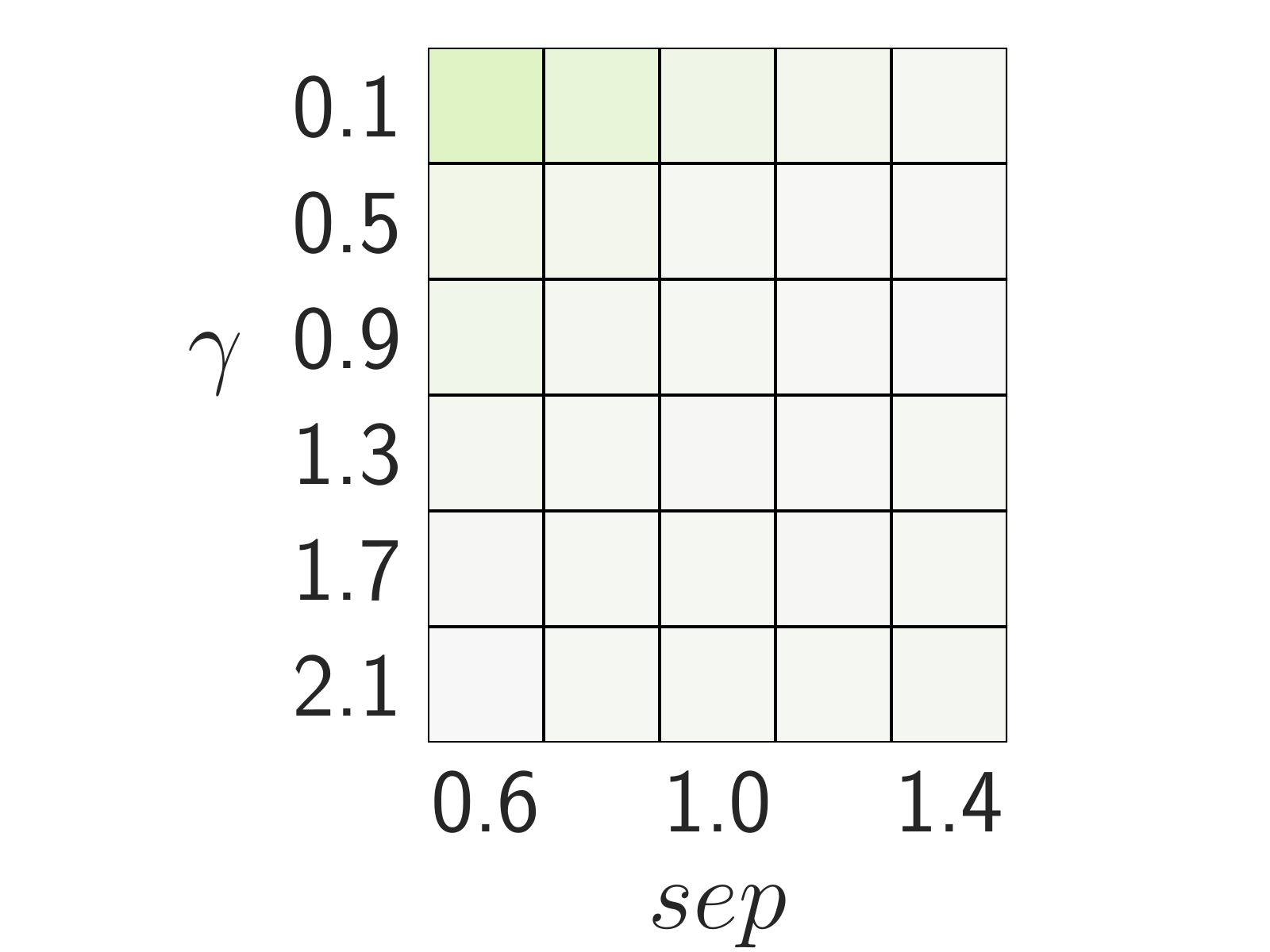} &
		\includegraphics[width=1\linewidth]{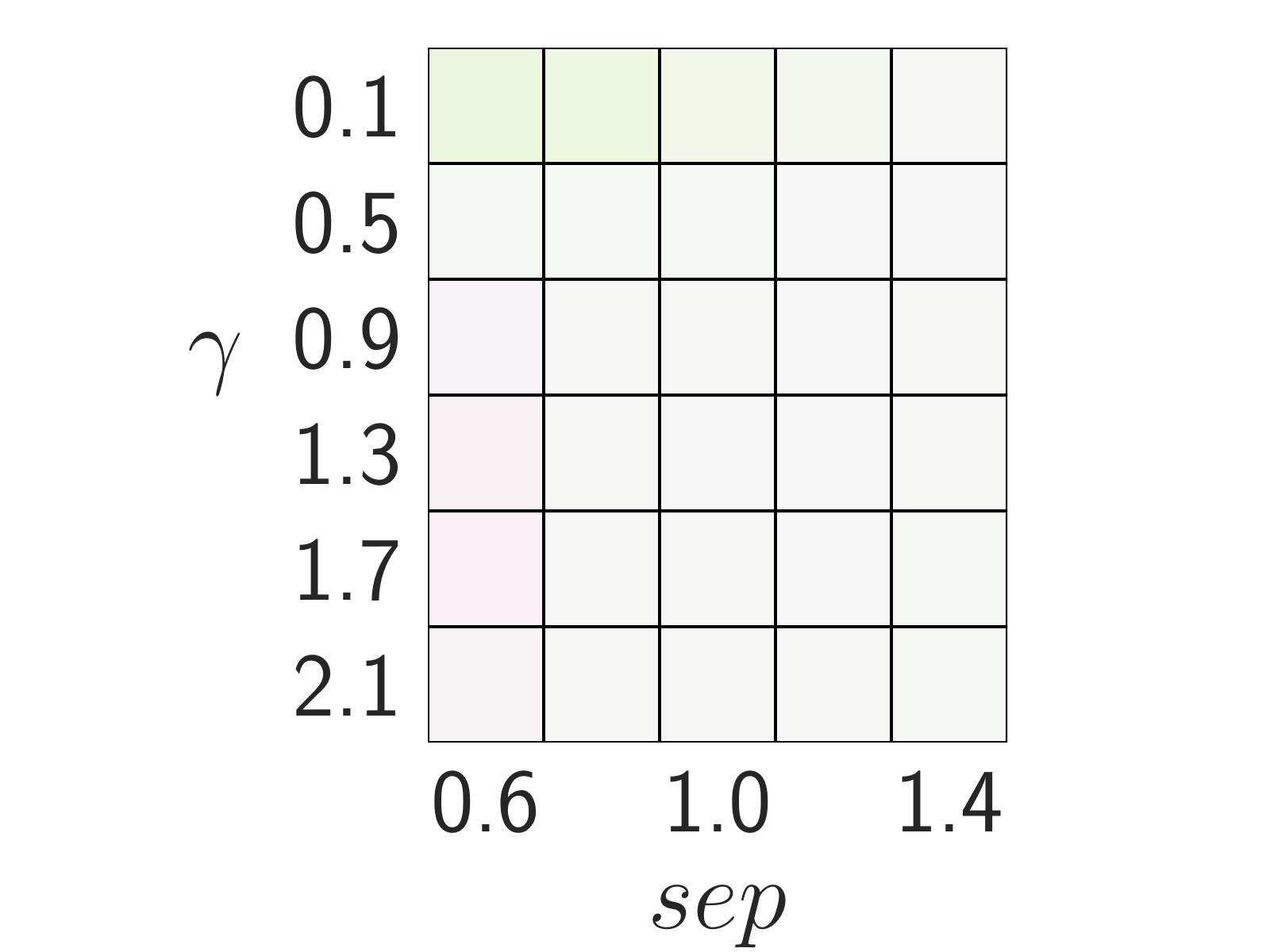} &
		\includegraphics[width=1\linewidth]{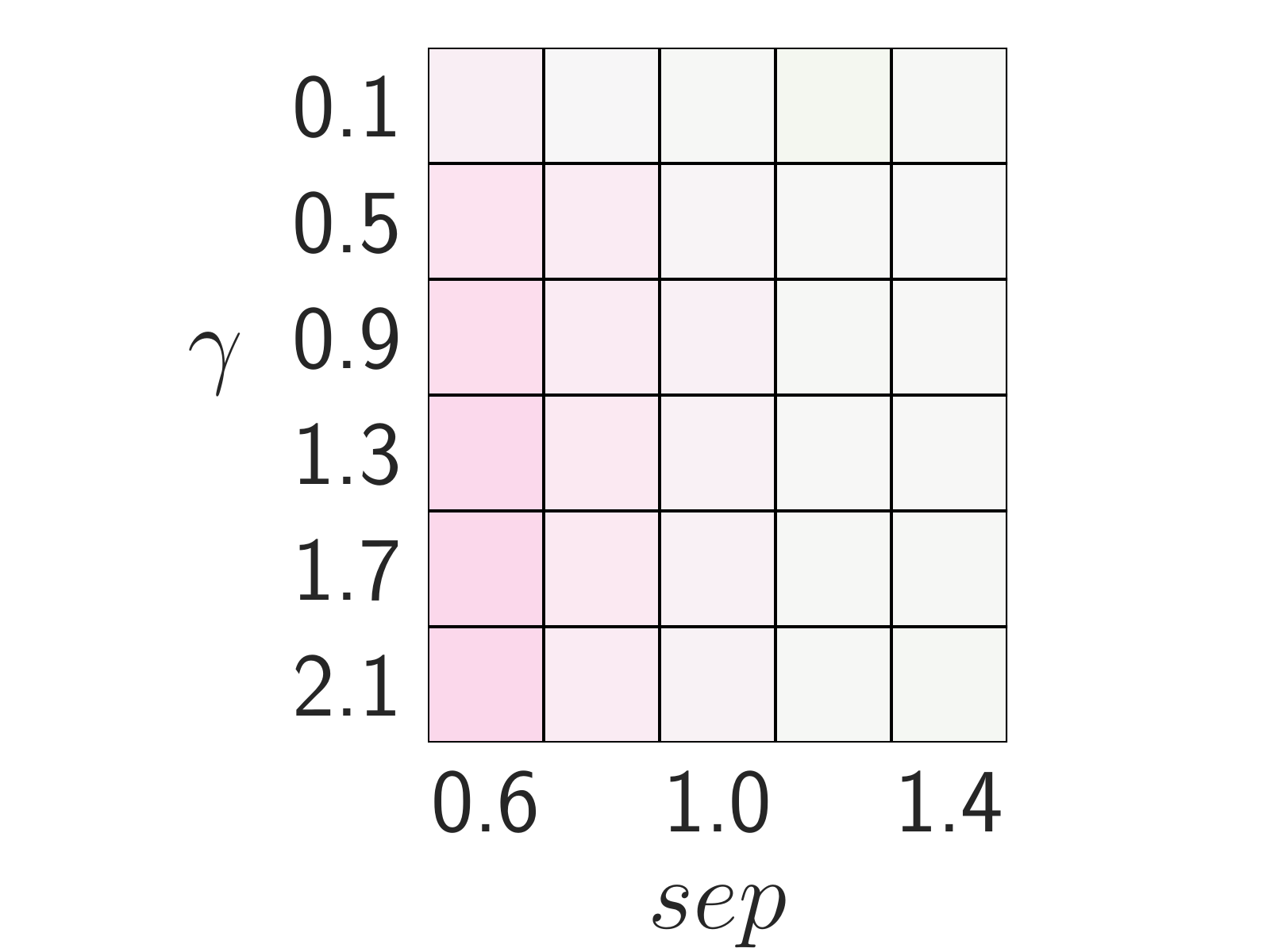} & \multirow{3}{*}[2em]{\includegraphics[width=0.07\linewidth]{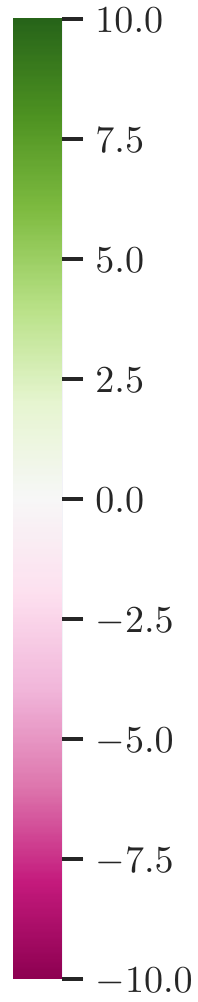}}\\
		4 & \includegraphics[width=1\linewidth]{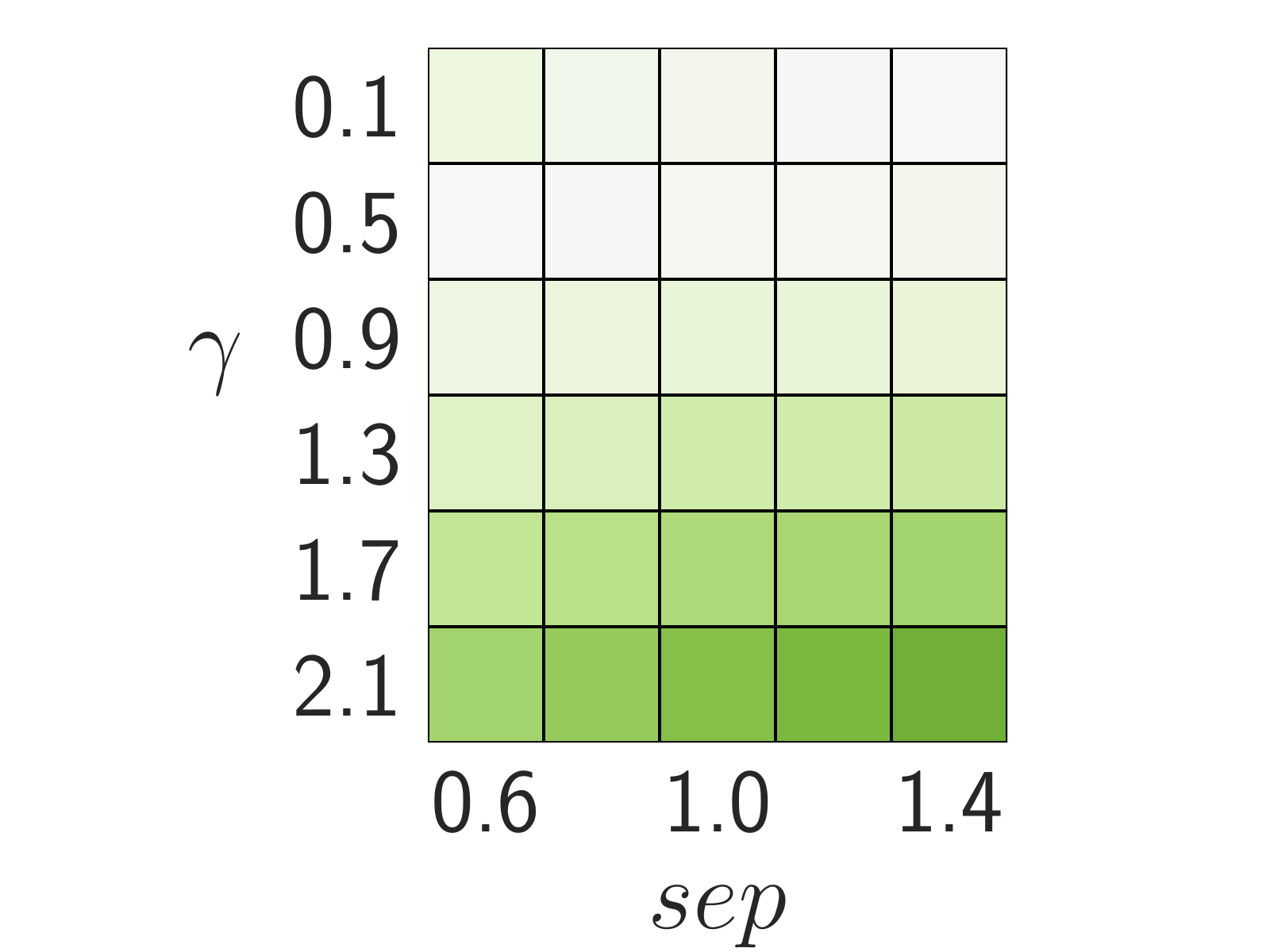} &
		\includegraphics[width=1\linewidth]{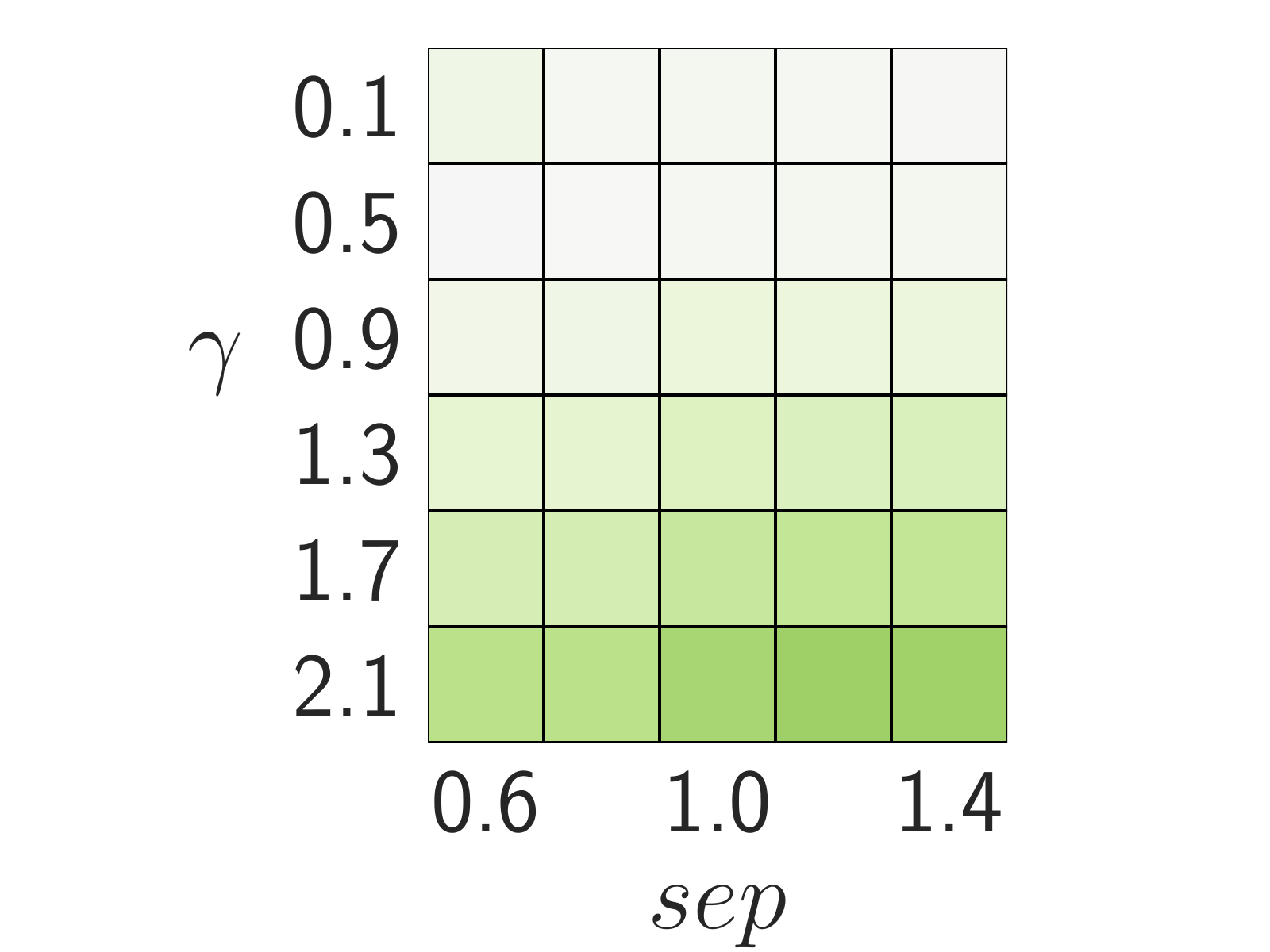} &
		\includegraphics[width=1\linewidth]{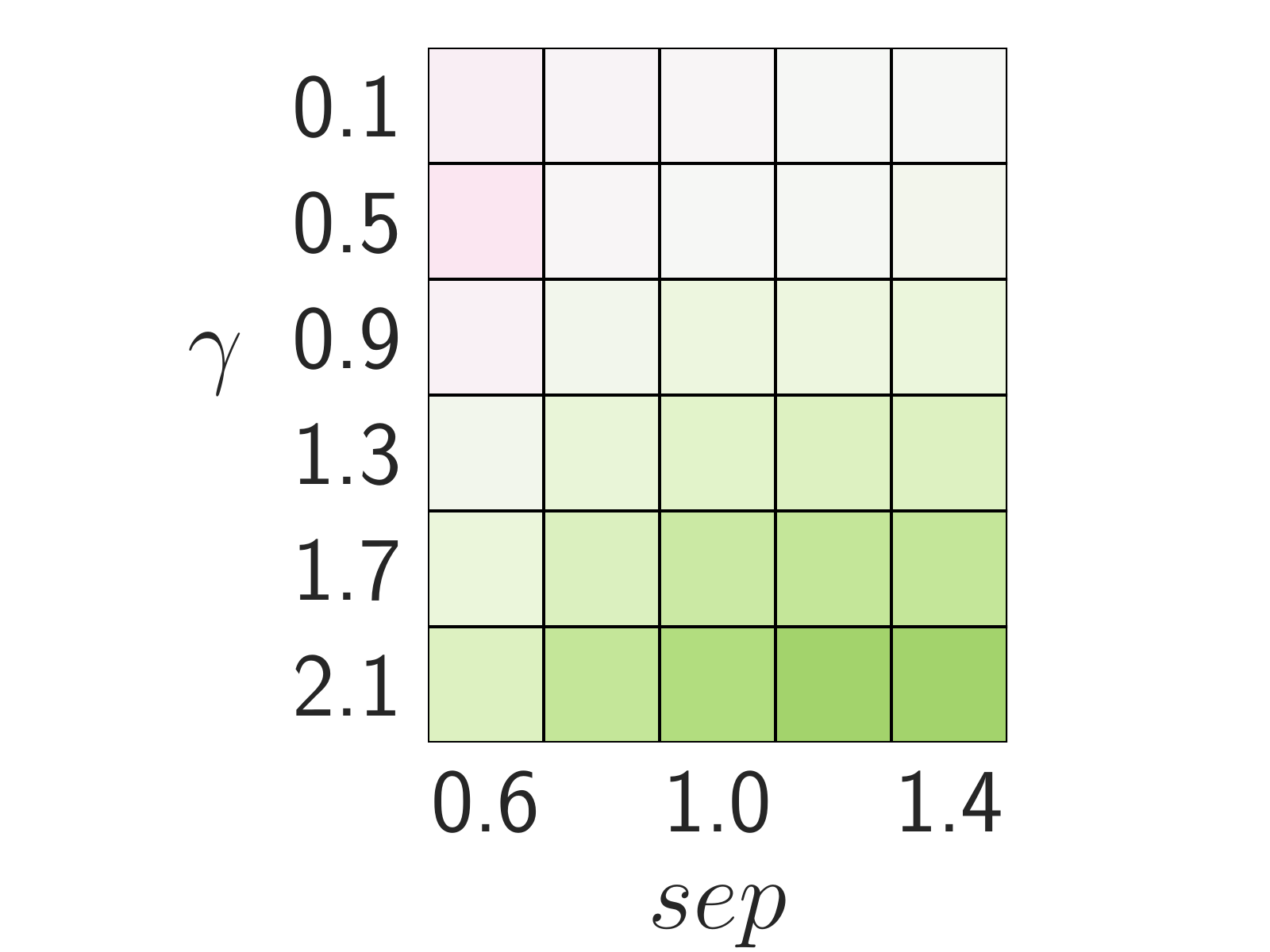} & \\
		8 & \includegraphics[width=1\linewidth]{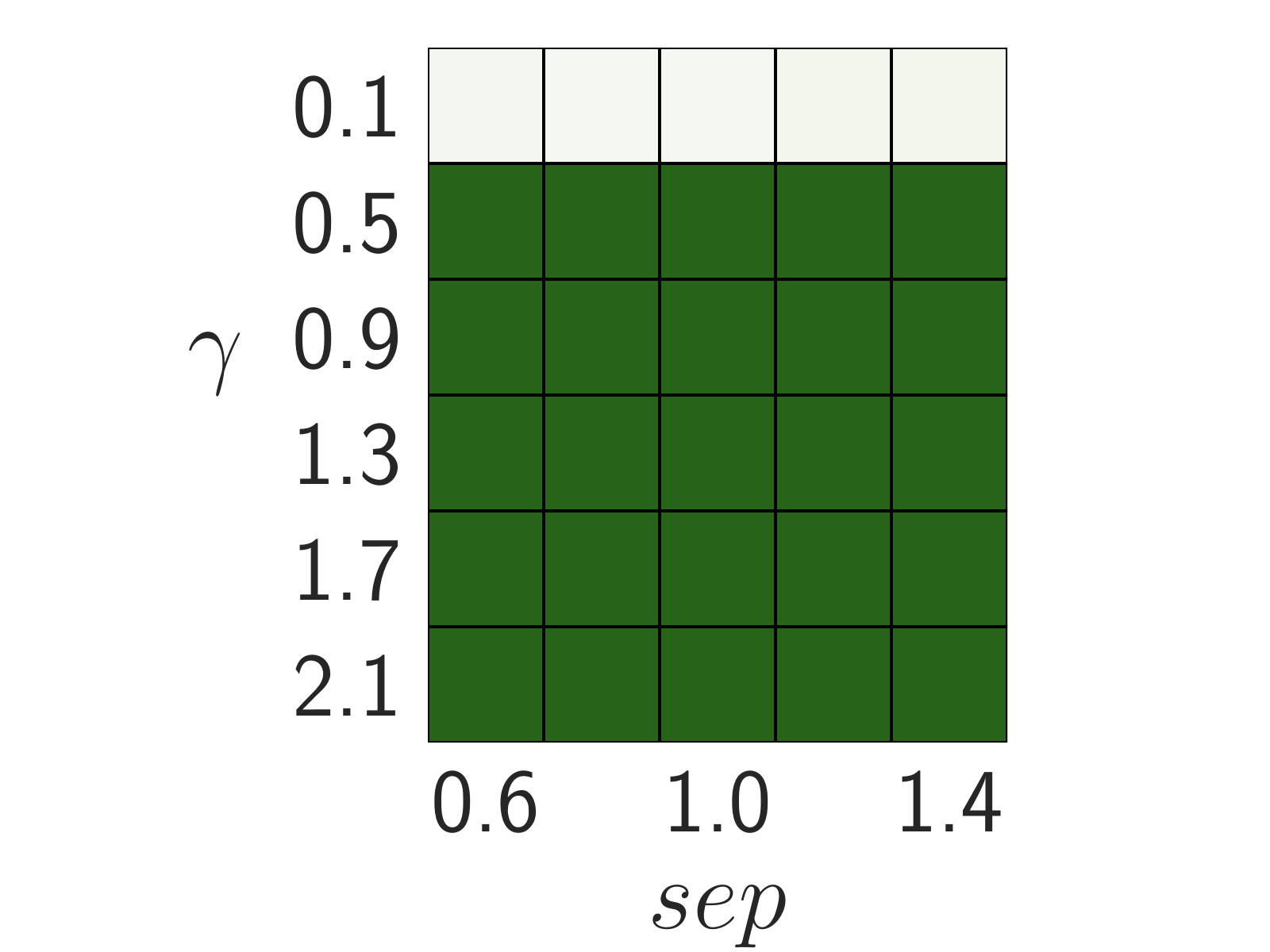} &
		\includegraphics[width=1\linewidth]{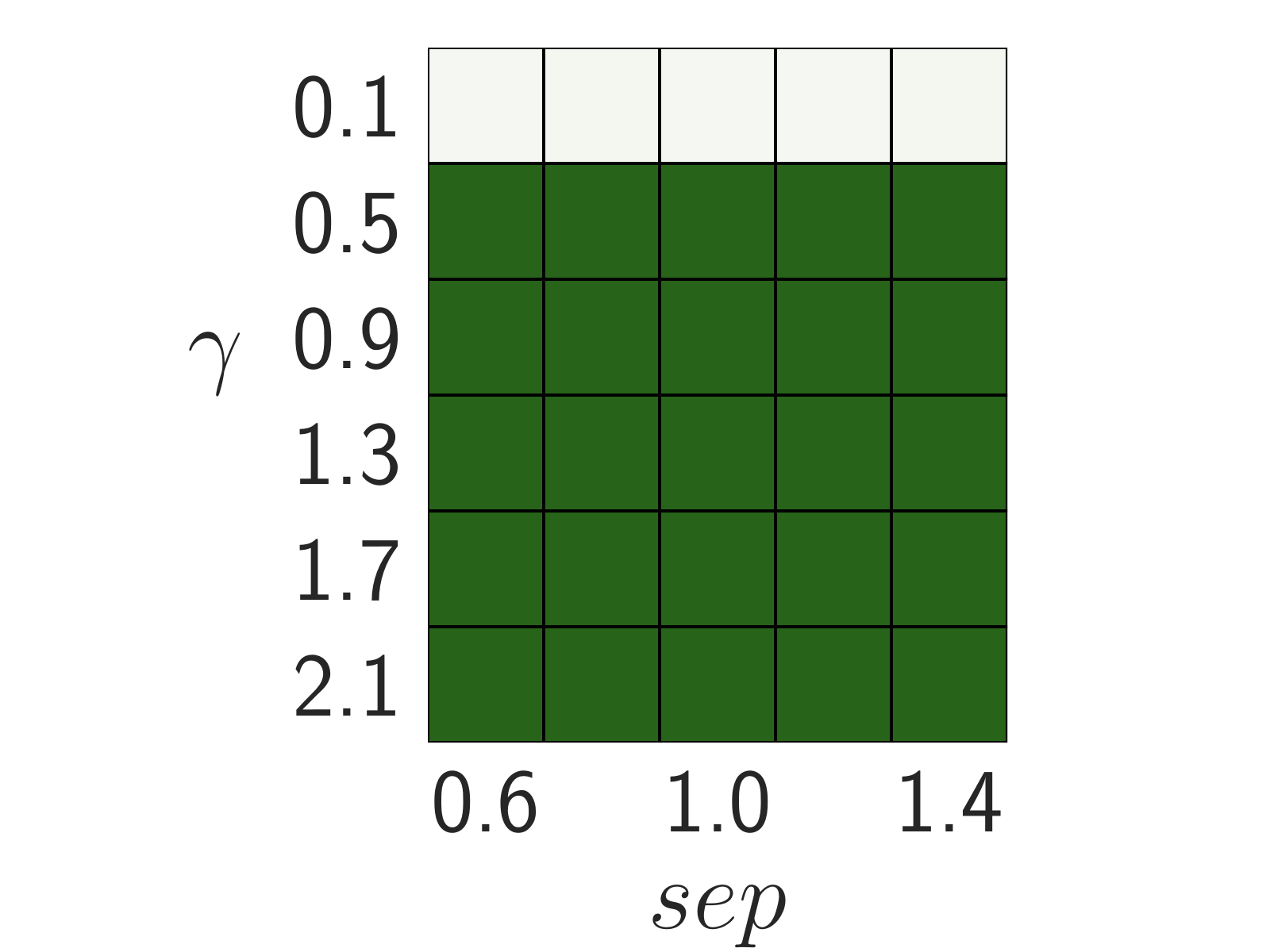} &
		\includegraphics[width=1\linewidth]{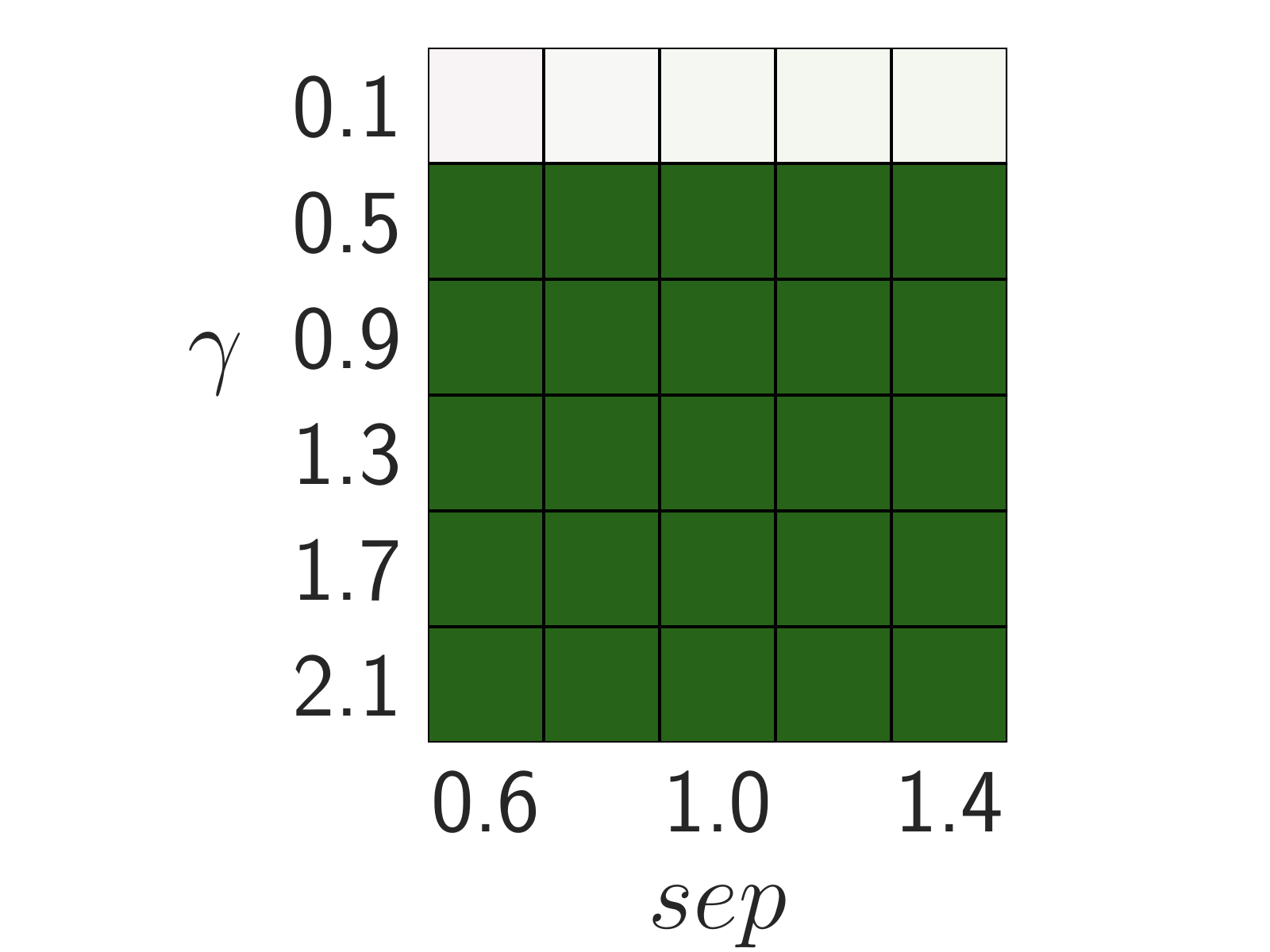} & \\
	\end{tabular}
\caption{$F1_{diff}$ heatmaps of \textit{OKSVM} and \textit{SVM}. The average performance of 100 runs on the test set for each configuration is represented varying $\gamma$ and $sep$. Higher is better.}
\label{fig:heatmap_Fm}
\end{figure}

\begin{figure}[t]
\centering
	\begin{tabular}{D|CCCc}
	    $dim$ & $C=0.5$ & $C=1.0$ & $C=1.5$ & \\
	    \midrule
		2 & \includegraphics[width=1\linewidth]{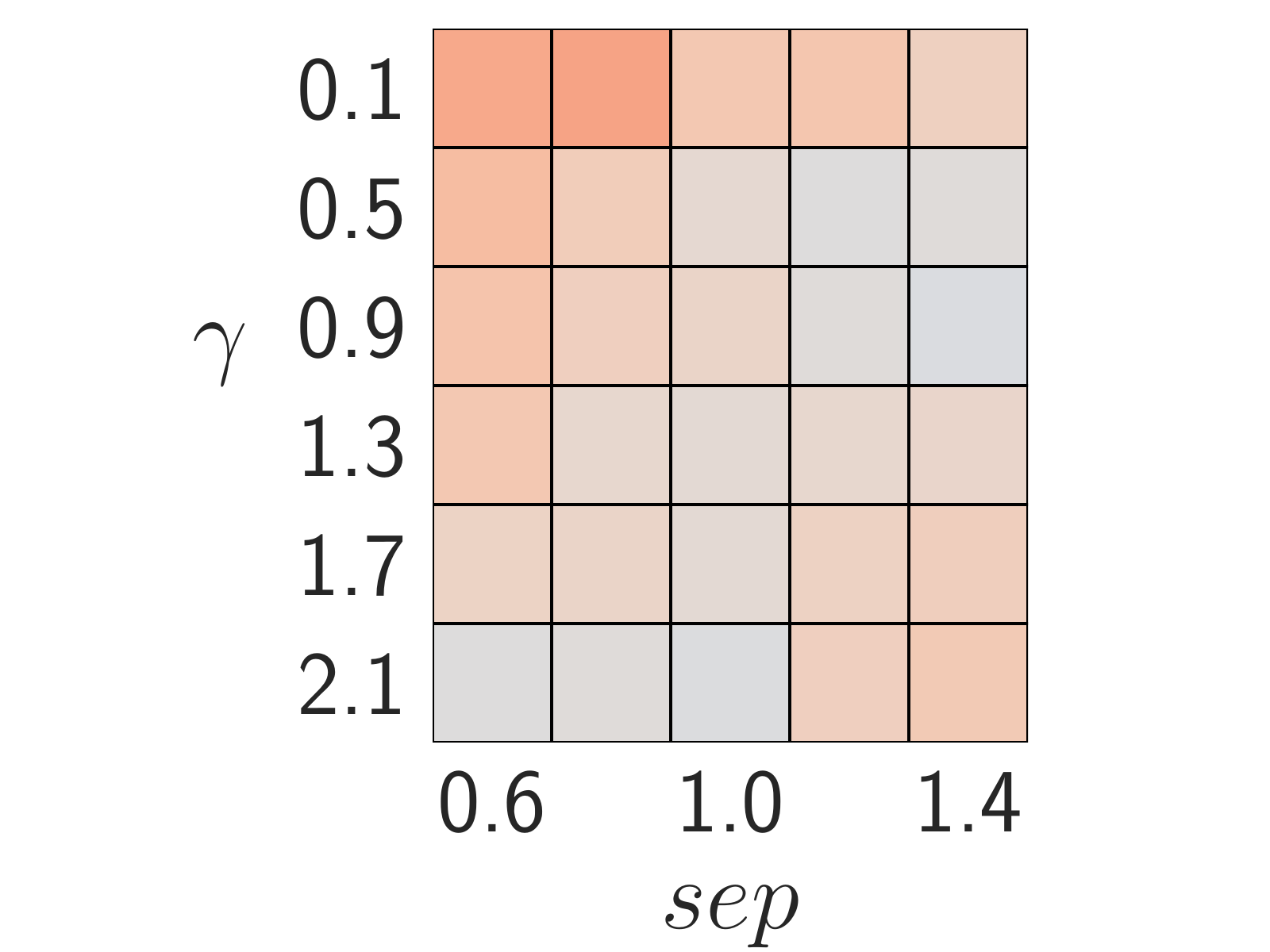} &
		\includegraphics[width=1\linewidth]{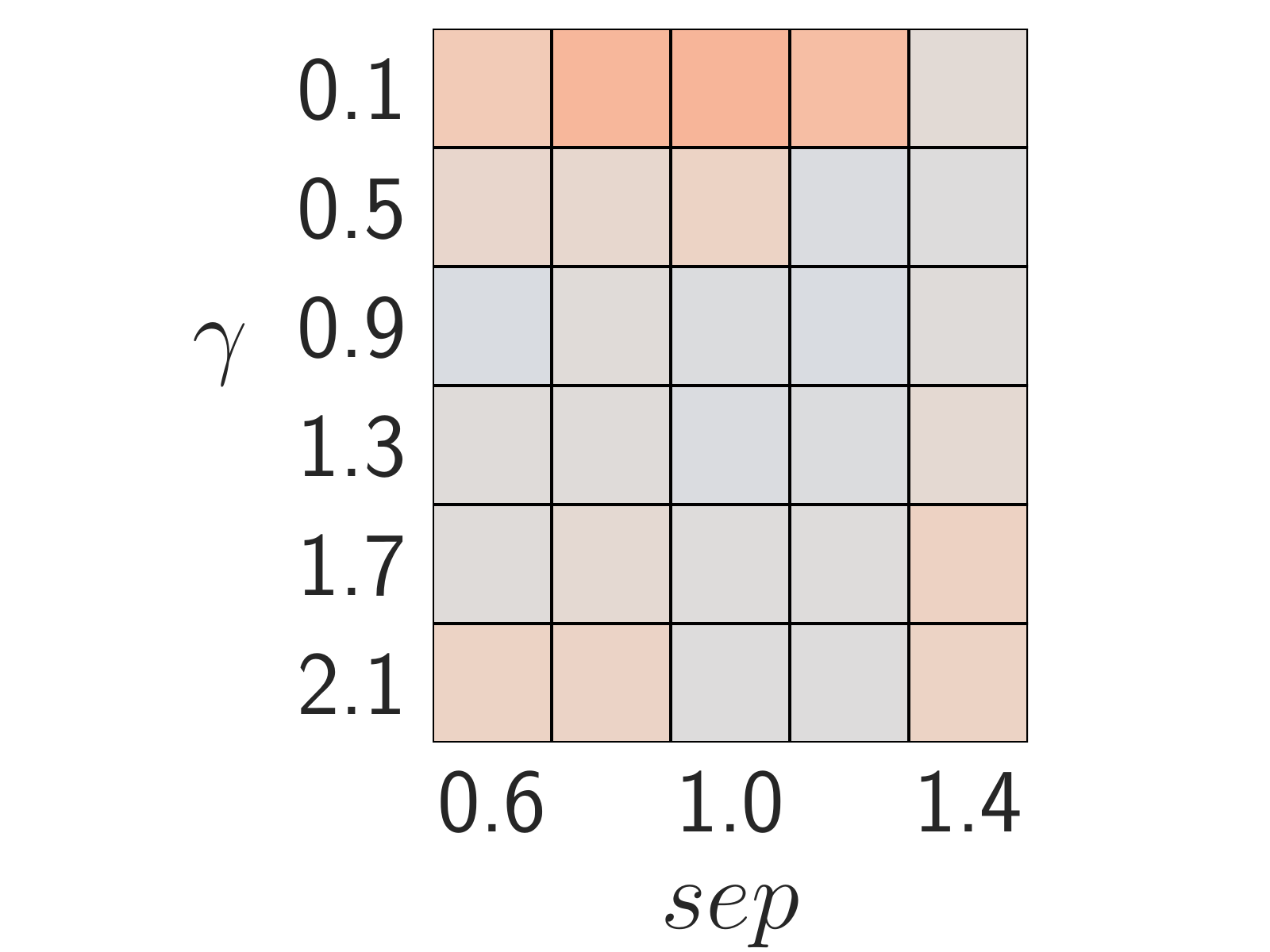} &
		\includegraphics[width=1\linewidth]{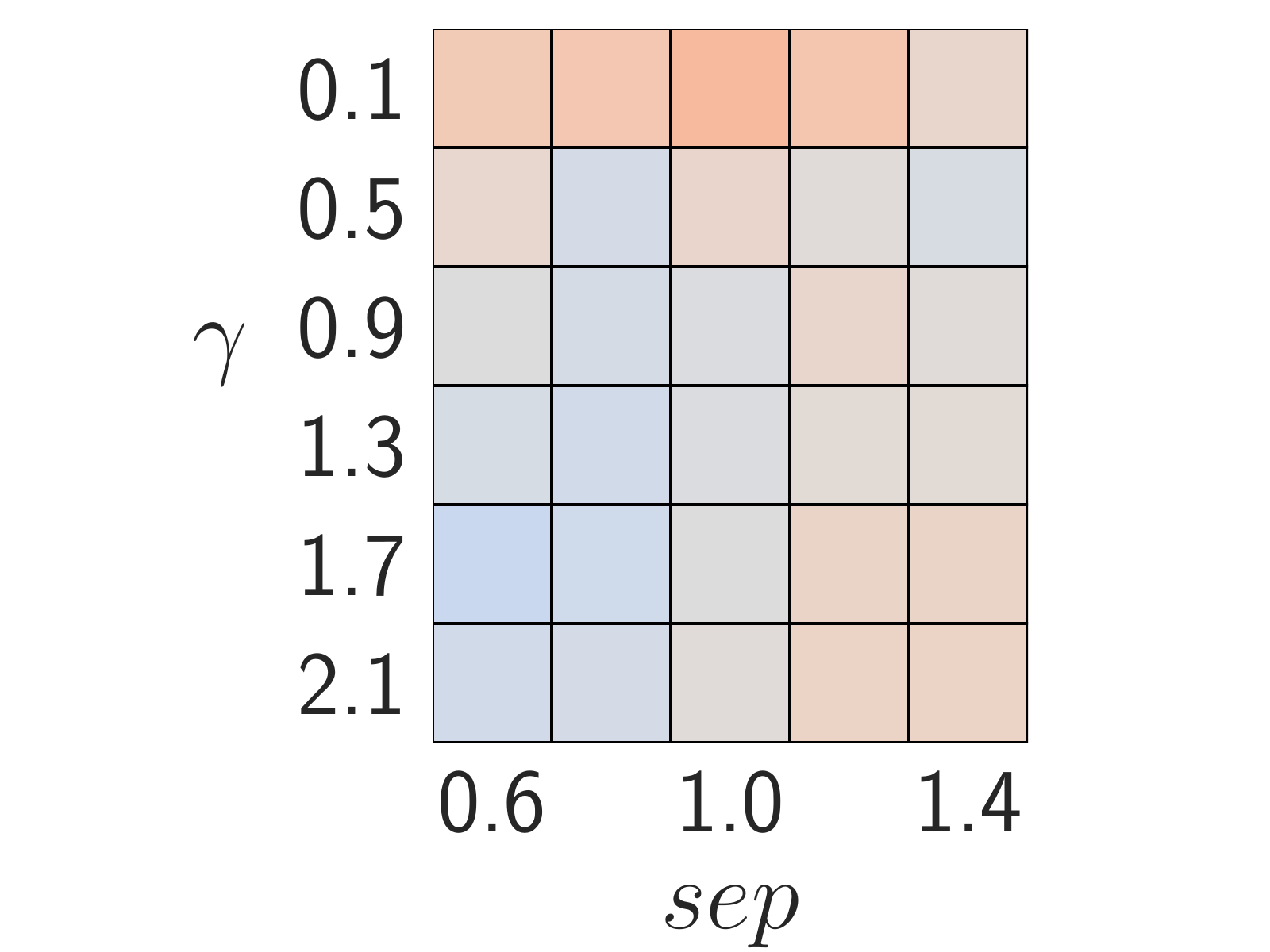} & \multirow{3}{*}[2.75em]{\includegraphics[width=0.07\linewidth]{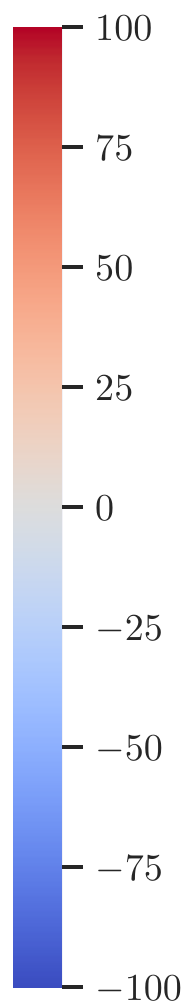}}\\
		4 & \includegraphics[width=1\linewidth]{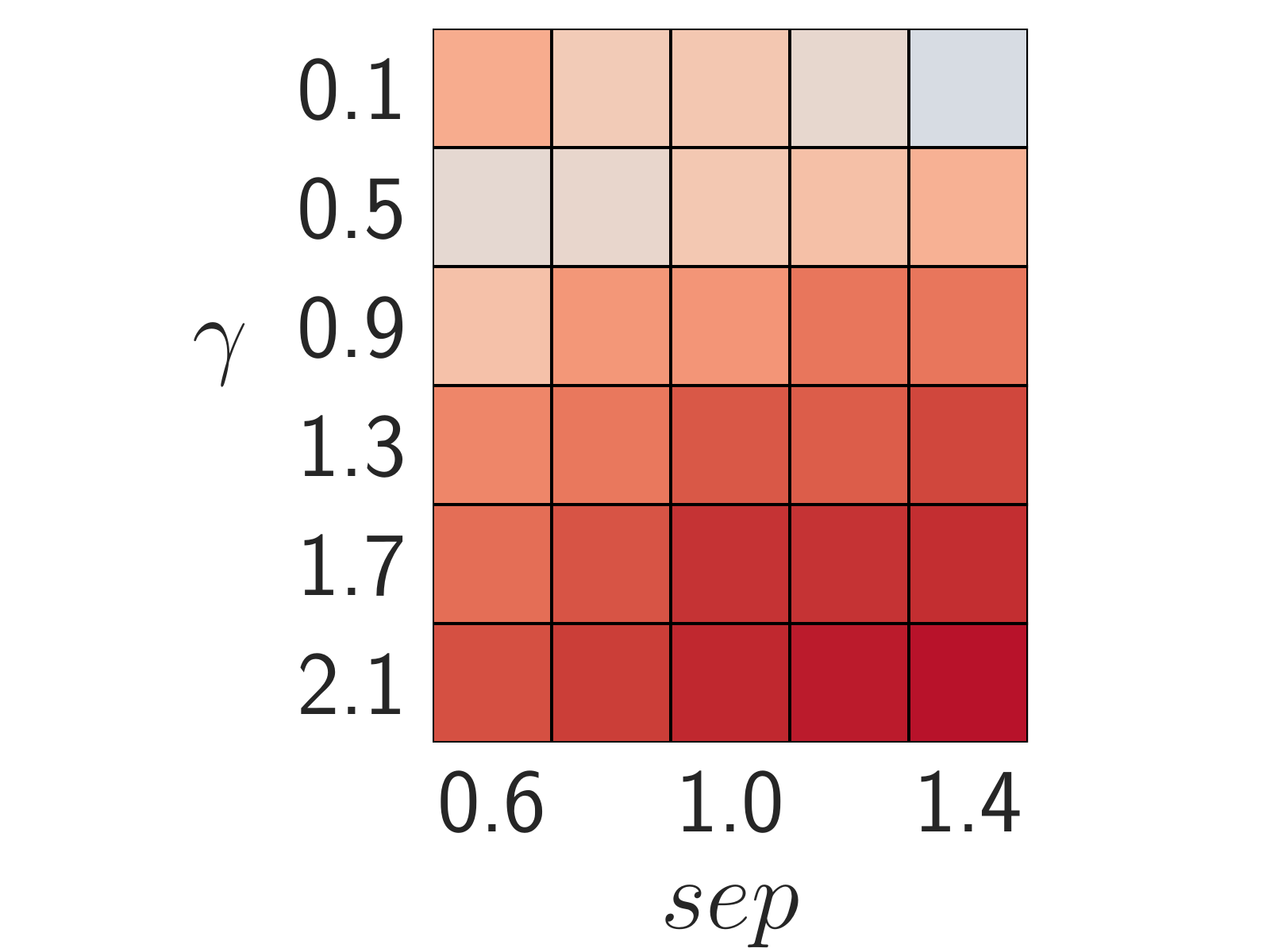} &
		\includegraphics[width=1\linewidth]{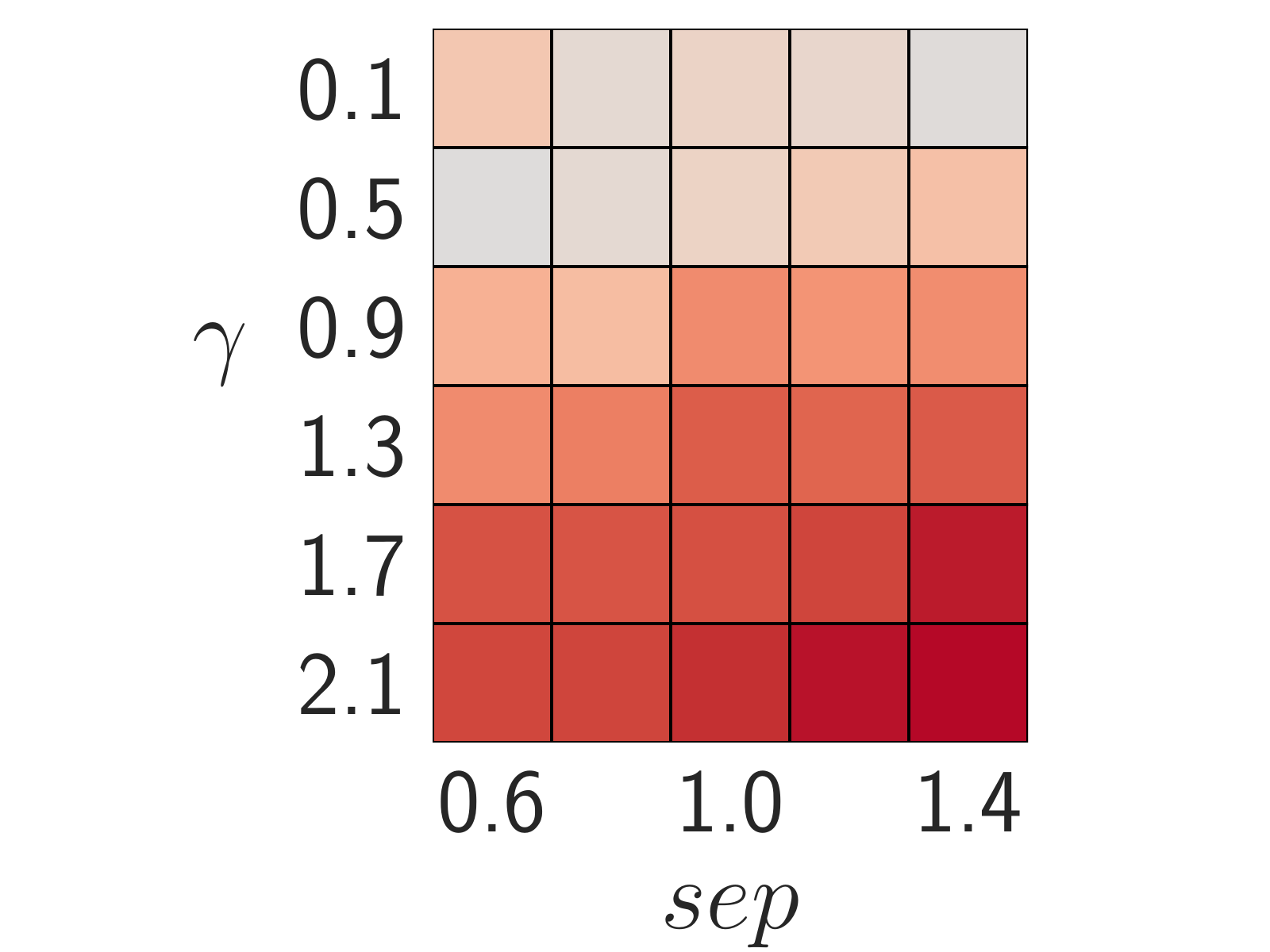} &
		\includegraphics[width=1\linewidth]{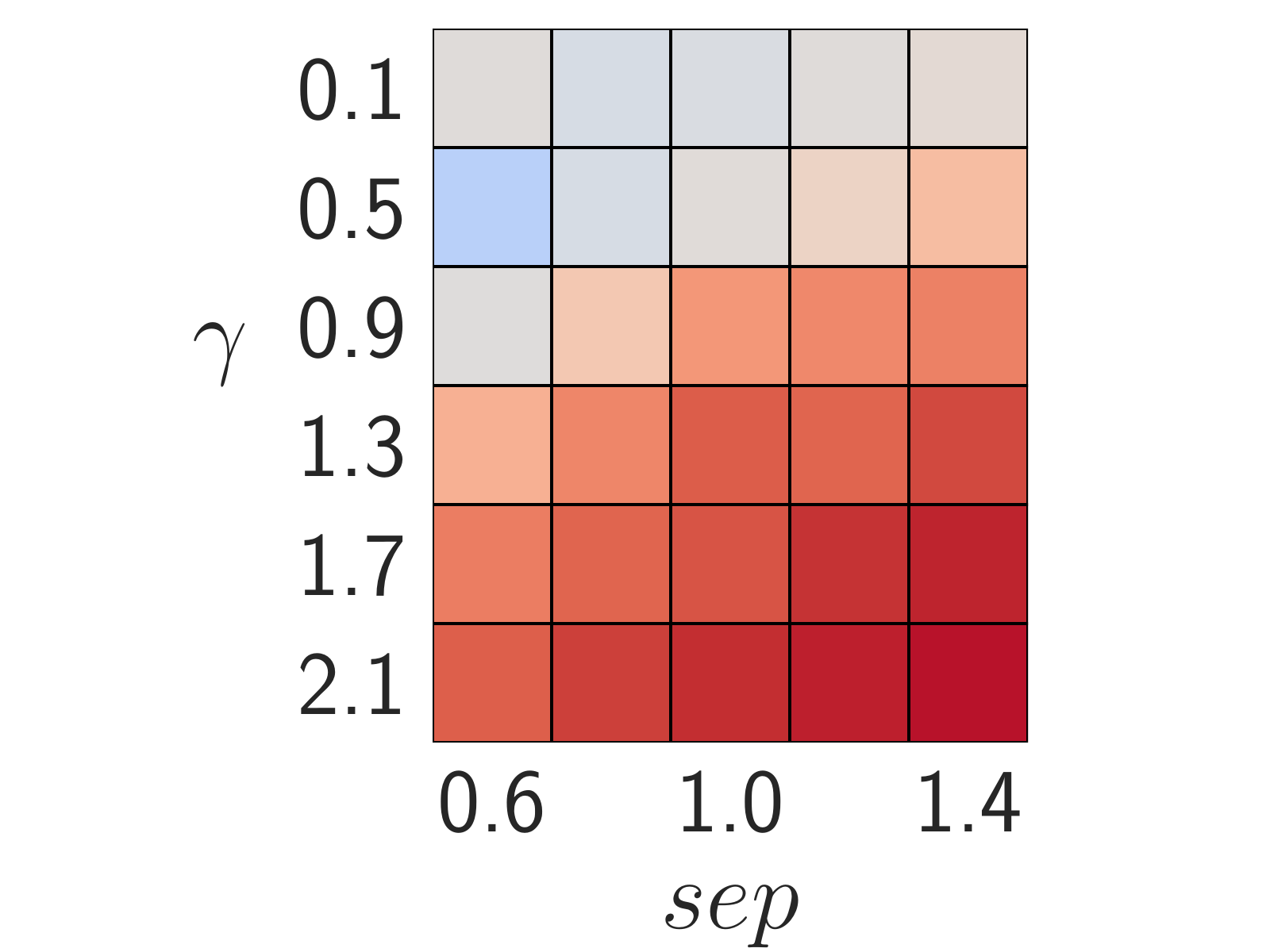} & \\
		8 & \includegraphics[width=1\linewidth]{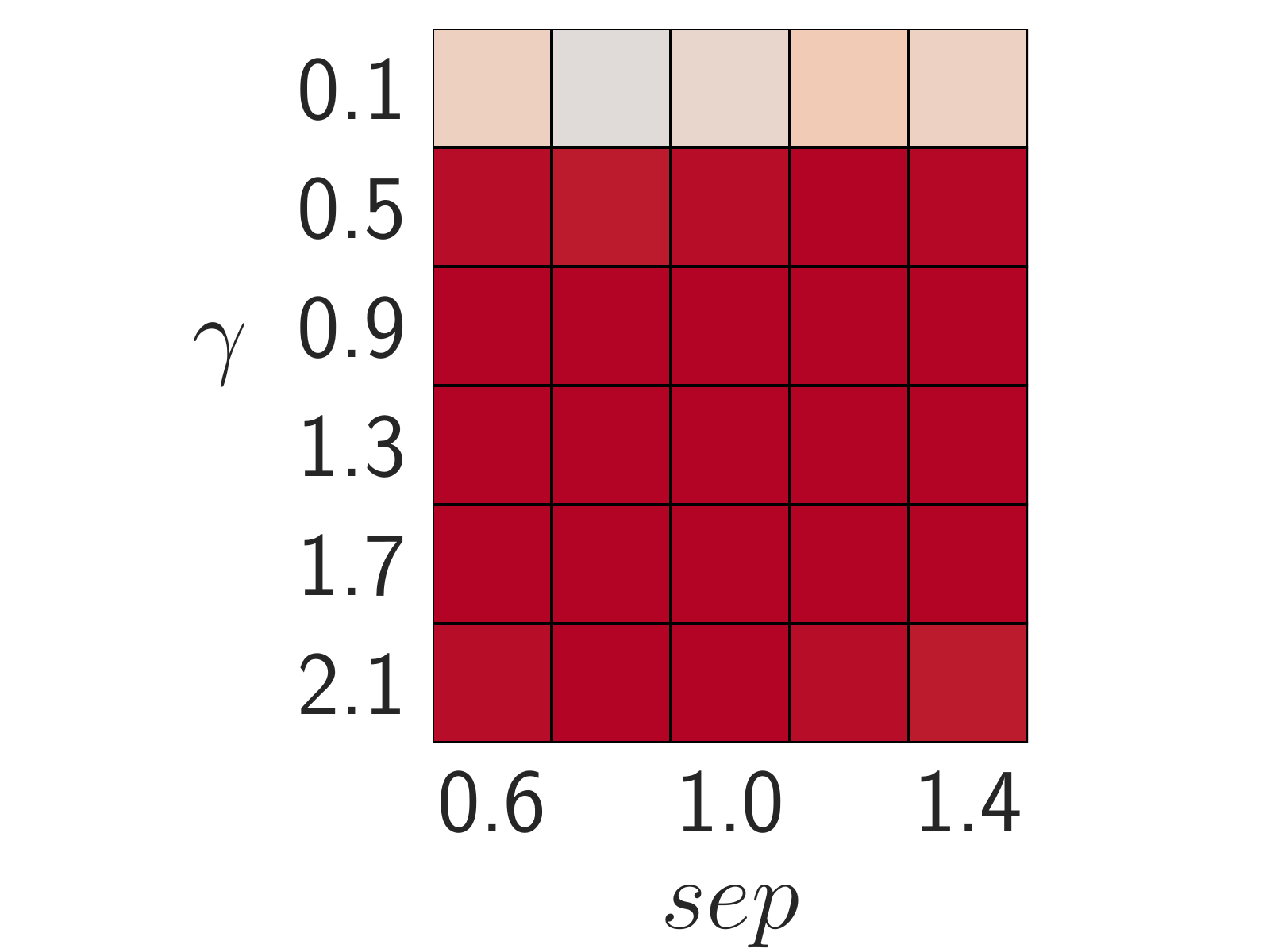} &
		\includegraphics[width=1\linewidth]{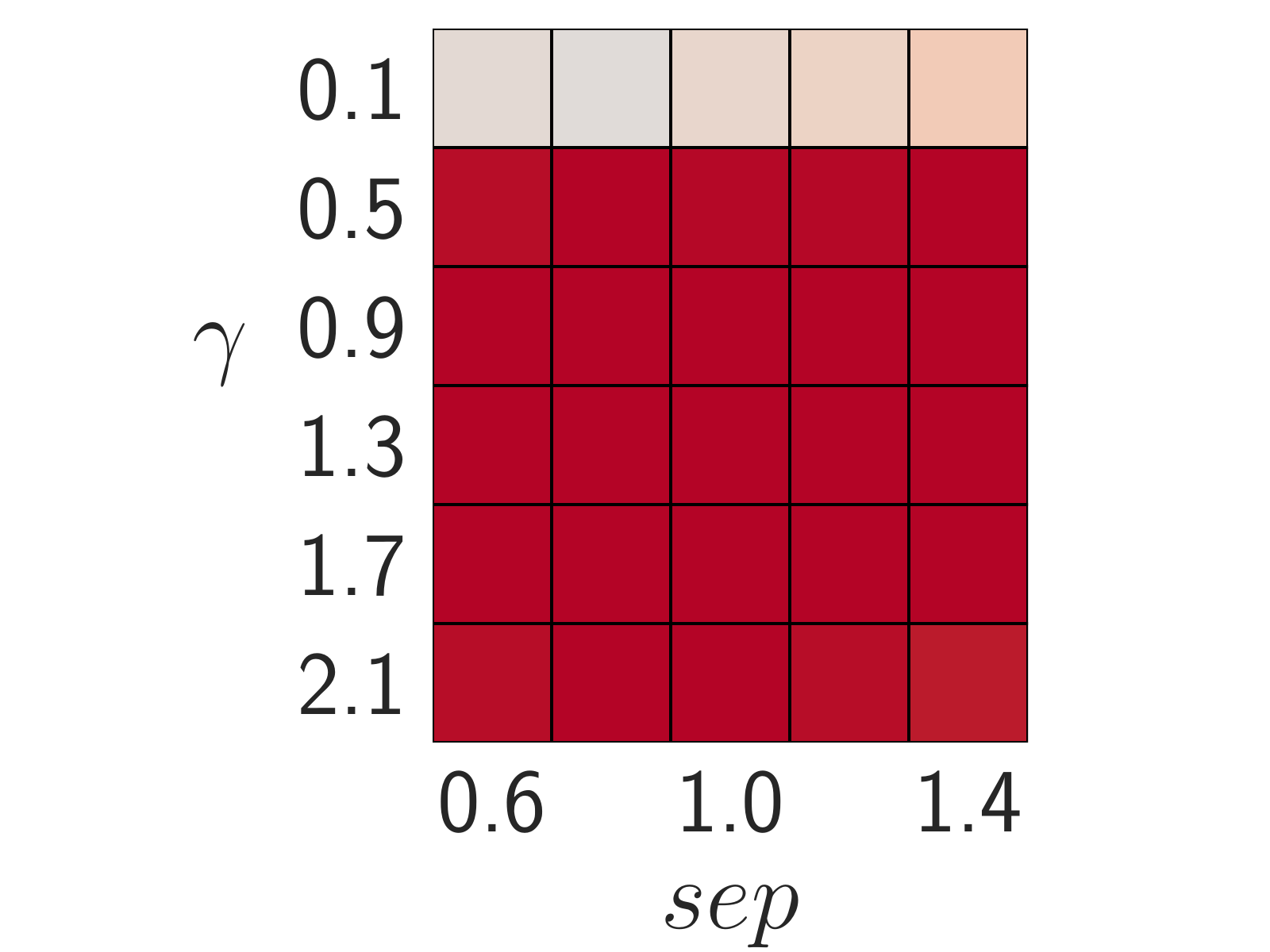} &
		\includegraphics[width=1\linewidth]{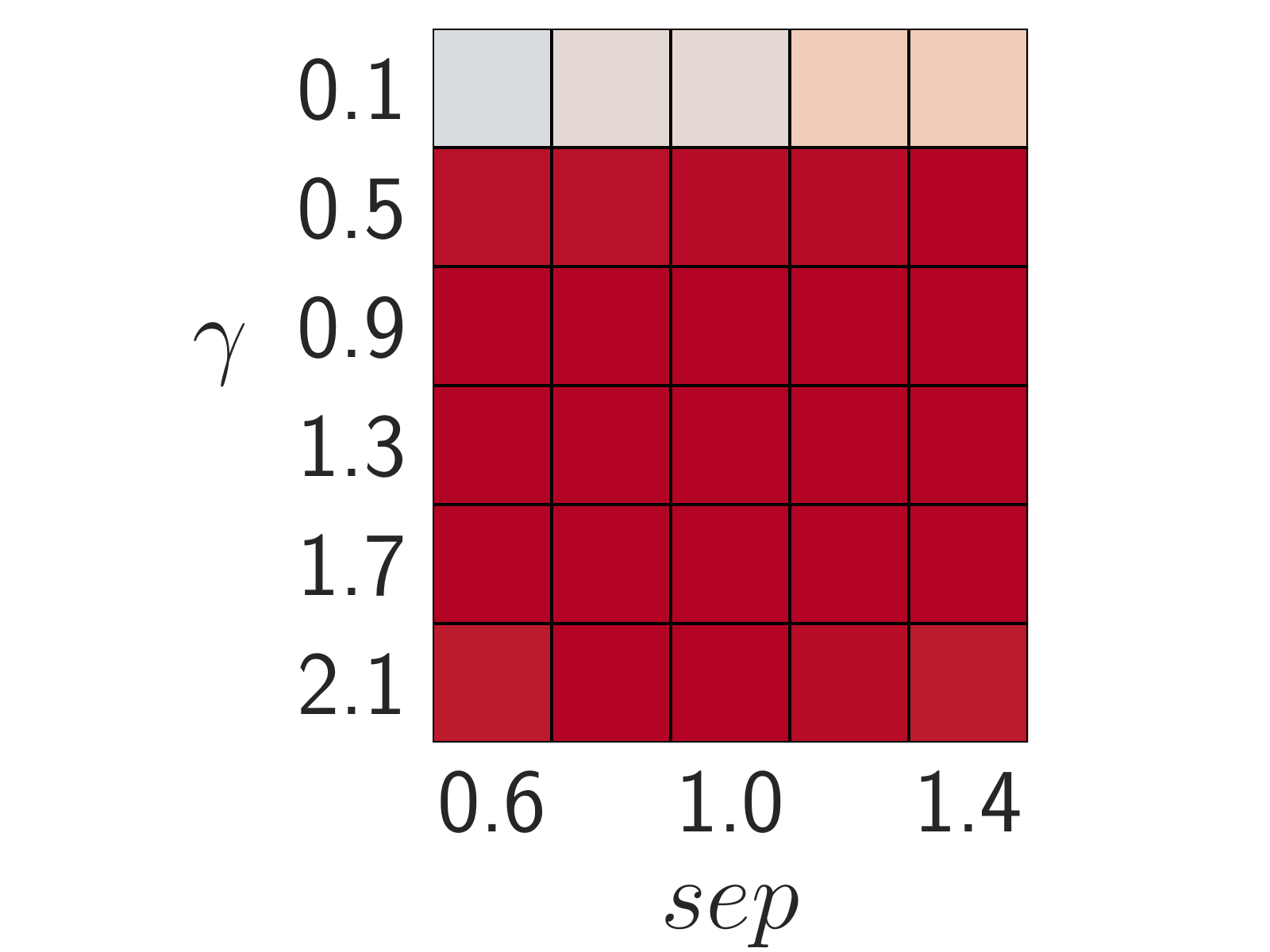} & \\
	\end{tabular}
\caption{Wins-Losses ratio heatmaps of \textit{OKSVM} and \textit{SVM}. The average performance of 100 runs on the test set for each configuration is represented varying $\gamma$ and $sep$. Higher is better.}
\label{fig:heatmap_Wins_Losses}
\end{figure}

A deeper analysis with more detail of this comparison is shown in \figurename\,\ref{fig:heatmap_Fm}, where the performance improvement ($F1_\text{diff}$) of \textit{OKSVM} against \textit{SVM} is reported. As can be observed, \textit{OKSVM} usually obtains better performance, although \textit{SVM} yields better performance for high values of $C$ and low values of $dim$. Moreover, the wins-losses ratio ($wlr$) comparison is reported in \figurename\,\ref{fig:heatmap_Wins_Losses}. It is interesting to observe how \textit{OKSVM} has a higher wins-losses ratio than \textit{SVM} while \textit{OKSVM} exhibits a lower $F1$ performance in some of the tested configurations, particularly those with a lower dimension and a higher $C$. This behavior has been analyzed into more detail for the configuration $dim=2$, $C=1.5$, $\gamma = 0.1$ and $sep = 0.6$ (top right image, top left square in \figurename s\,\ref{fig:heatmap_Fm} y \ref{fig:heatmap_Wins_Losses}). 
By studying the results of this configuration among the 100 runs, the difference between both methods is $-0.0072$ on average ($F1_\text{diff}=-0.72$) where the maximum difference in favor of \textit{OKSVM} is $0.1389$ and the maximum difference in favor of \textit{SVM} is $0.5369$. \textit{OKSVM} won 56 and lost 36 times (8 draws) against \textit{SVM}, so that, $wlr=20$. Therefore, it can be established that the proposed method \textit{OKSVM} usually outperforms \textit{SVM}; however, in the cases when it does not achieve a good performance it is highly surpassed by the traditional \textit{SVM}, probably caused by over-fitting.

\subsubsection{Optimized hyperparameters comparison}

The second experiment compares the performances of both methods when the data is taken into account to tune the hyperparameters $C$ and $\gamma$ before testing the SVMs. This process consists of a grid search by using a validation splits of the data to find the optimal parameters. Therefore, ten tuning runs for each configuration of $C$ and $\gamma$ are executed for \textit{SVM} method, and the parameters with best mean $F1$ are selected. For \textit{OKSVM}, only $C$ is tuned since $\gamma$ is optimized during the training. 

\begin{figure}[t]
\centering
\includegraphics[width=0.32\linewidth]{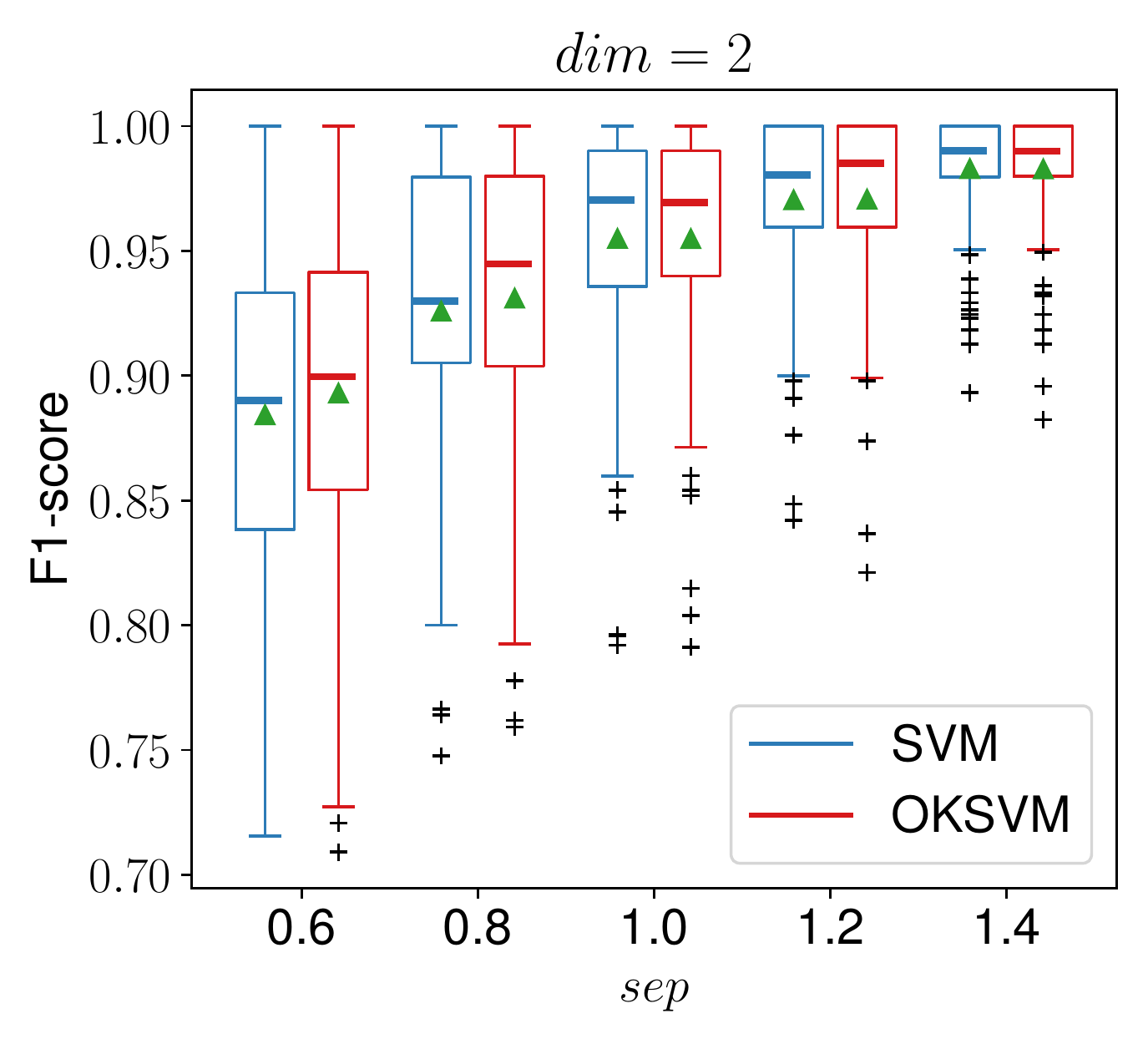}
\includegraphics[width=0.32\linewidth]{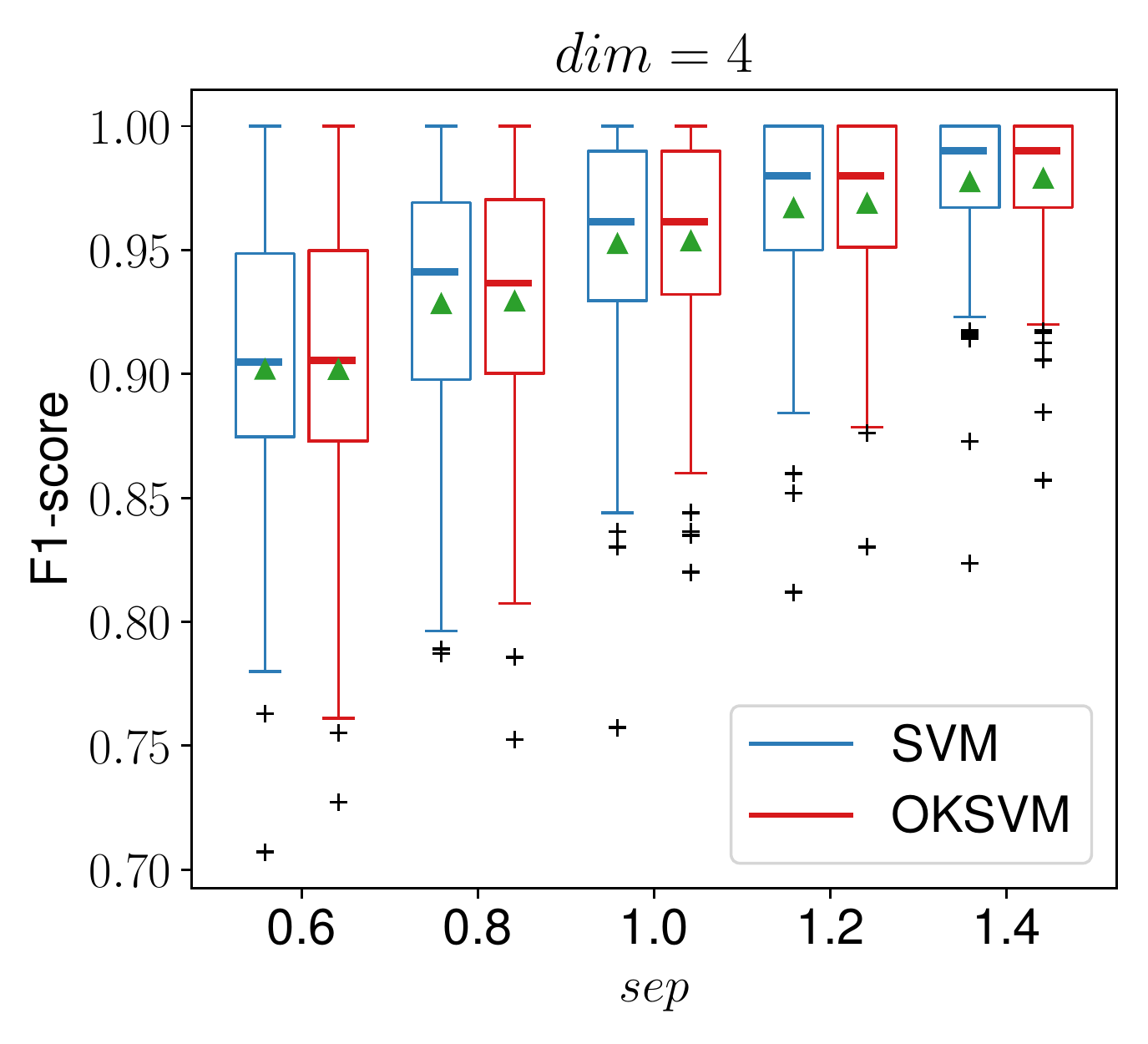}
\includegraphics[width=0.32\linewidth]{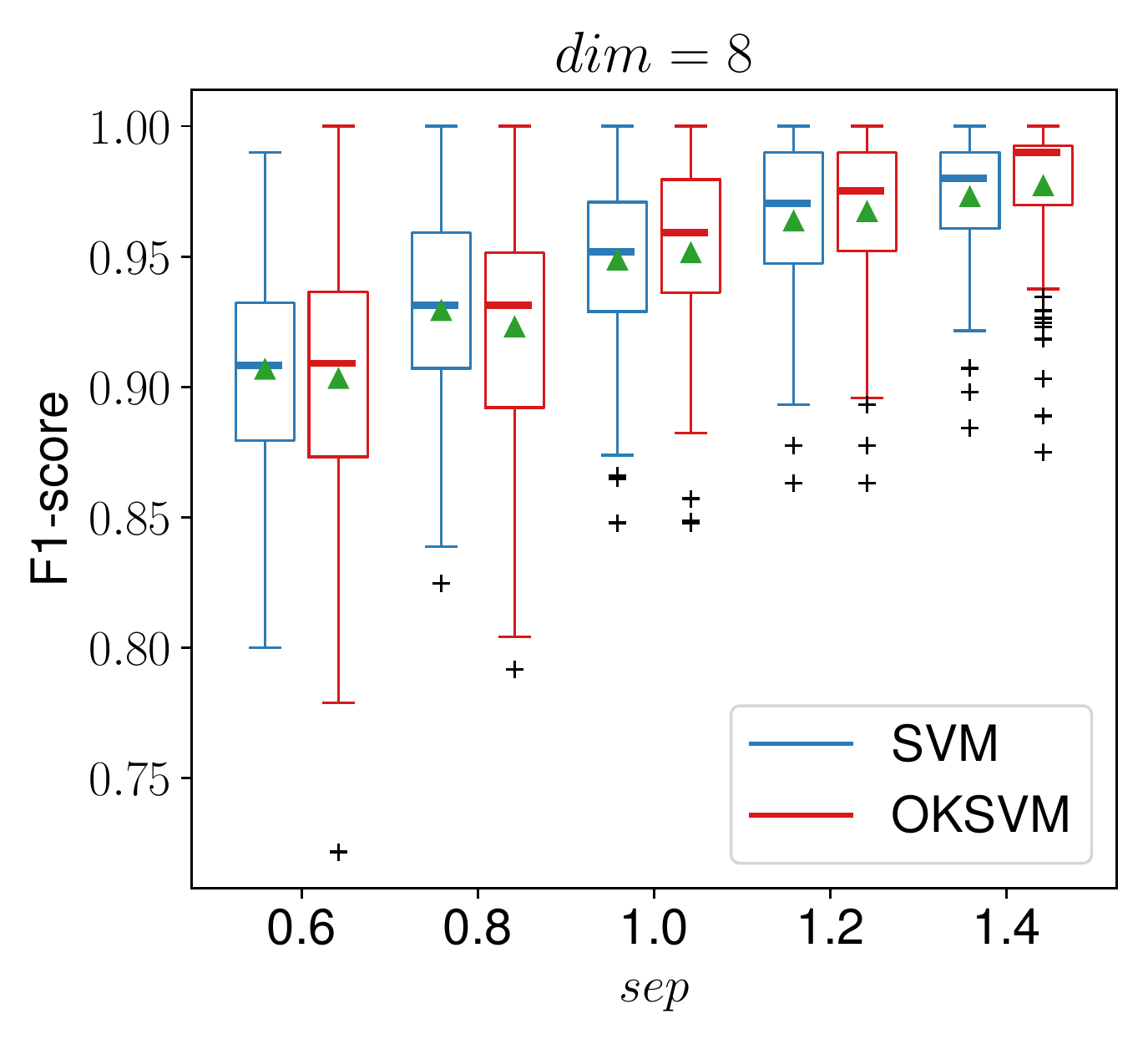} \\
\caption{$F1$ boxplots of \textit{OKSVM} and \textit{SVM} among 100 runs. Results with dimensions 2, 4, and 8 are depicted varying the class separation $sep$. Median (line) and mean (triangle) values are represented.}
\label{fig:Boxplots_Fm_AutoC}
\end{figure}

\figurename\,\ref{fig:Boxplots_Fm_AutoC} exhibits boxplots with the $F1$ performance of both methods among 100 runs. The results are related to those presented in previous subsection: generally, \textit{OKSVM} outperforms \textit{SVM}, while the higher the separation $sep$ of the clusters of the dataset, the easier the classification. Mean and median \textit{OKSVM} values are higher in most cases, while the number of outlying runs is similar. Thus, \textit{OKSVM} has a good performance even if the hyperparameters are previously tuned.

\subsection{Real datasets}\label{sec:Real_datasets}

Several real datasets of the the UCI repository \cite{asuncion2007uci} have been selected to carry out experiments to compare \textit{OKSVM} and \textit{SVM} methods, which offer a wide range of complexity and size: Wisconsin Breast Cancer (\textit{breast-cancer-wisconsin}), Heart Disease (\textit{processed.cleveland}), Wisconsin Diagnostic Breast Cancer (\textit{wdbc}), Haberman's Survival (\textit{haberman}), Iris (\textit{iris}), Wine Quality (\textit{winequality-red}) and Banknote Authentication (\textit{banknote}). 
All of them are oriented for binary classification with the exception of \textit{iris}, where the classification between Virginica and Versicolor is done, and \textit{winequality-red}, where the label class ranges between 0 and 10, and we binarize it as $\leq5$ and $>5$.
Datasets were randomly split into training (80\%) and testing sets (20\%) in a stratified mode, while the tuned configurations of the parameters $C$ and $\gamma$ for both methods in these experiments were $C=\{0.1, 0.4, 0.7, 1.0, 1.3, 1.6, 1.9\}$ and $\gamma=\{0.005, 0.01, 0.05, 0.1, 0.5, 1.0, 1.5\}$.

\begin{figure}[t]
\centering
	\begin{tabular}{cc}
		\includegraphics[width=0.45\linewidth]{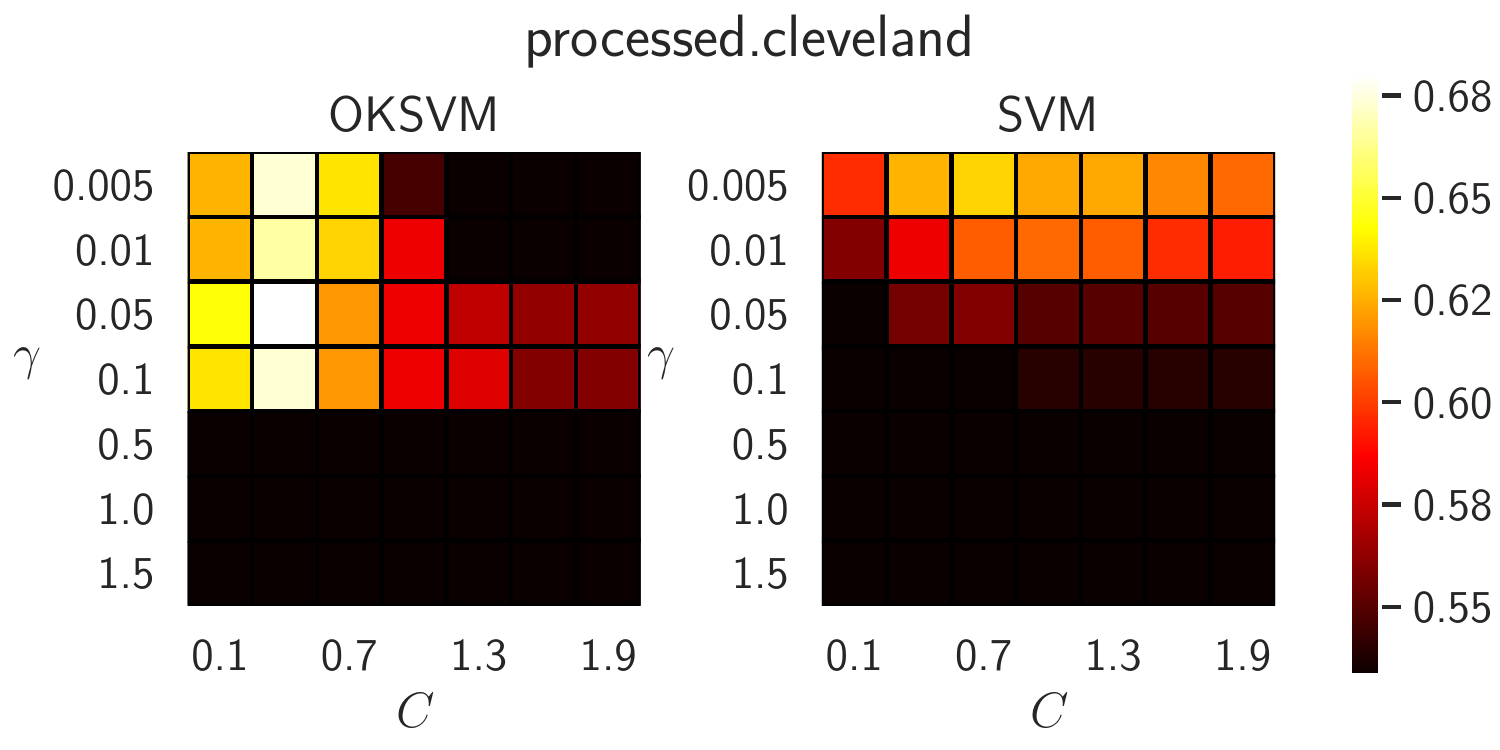} &
		\includegraphics[width=0.45\linewidth]{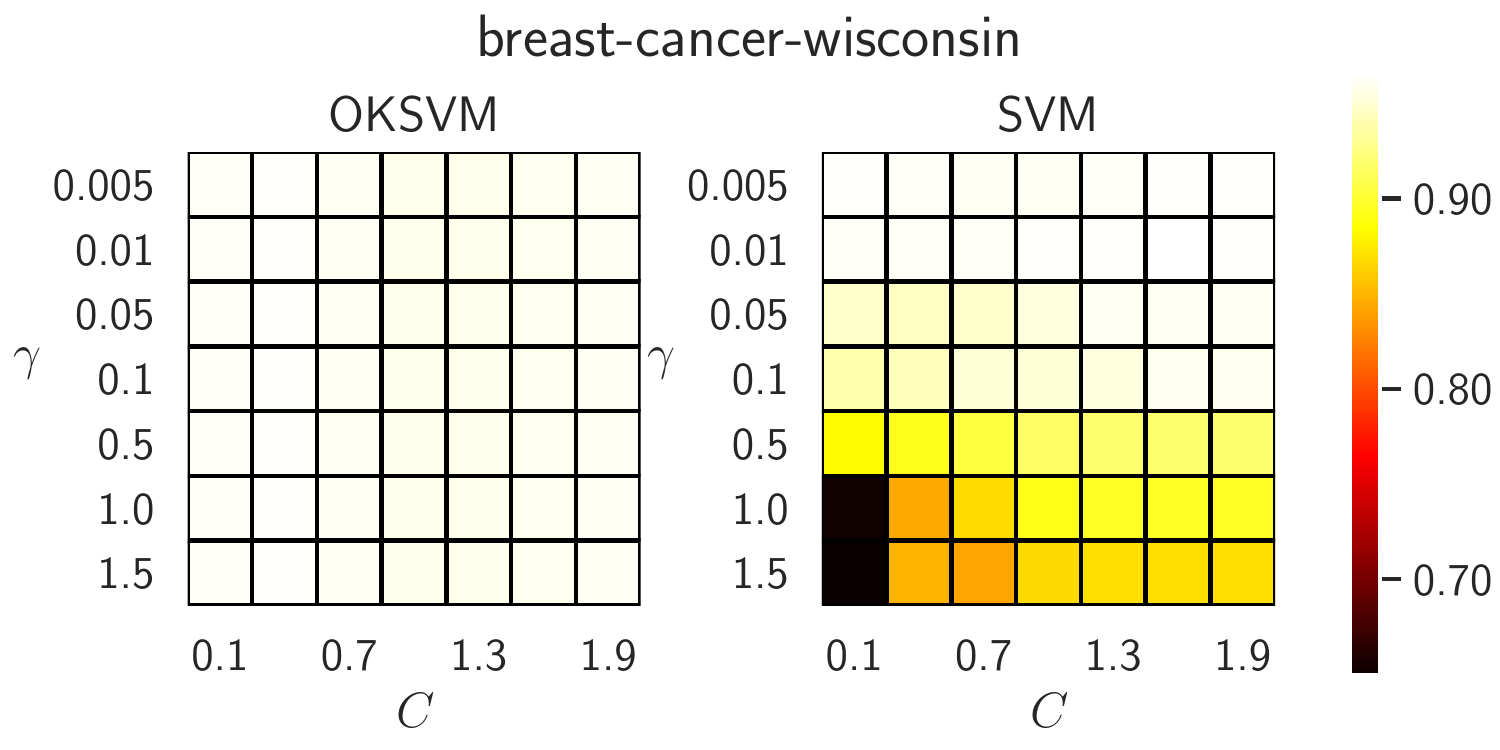}\\
		\includegraphics[width=0.45\linewidth]{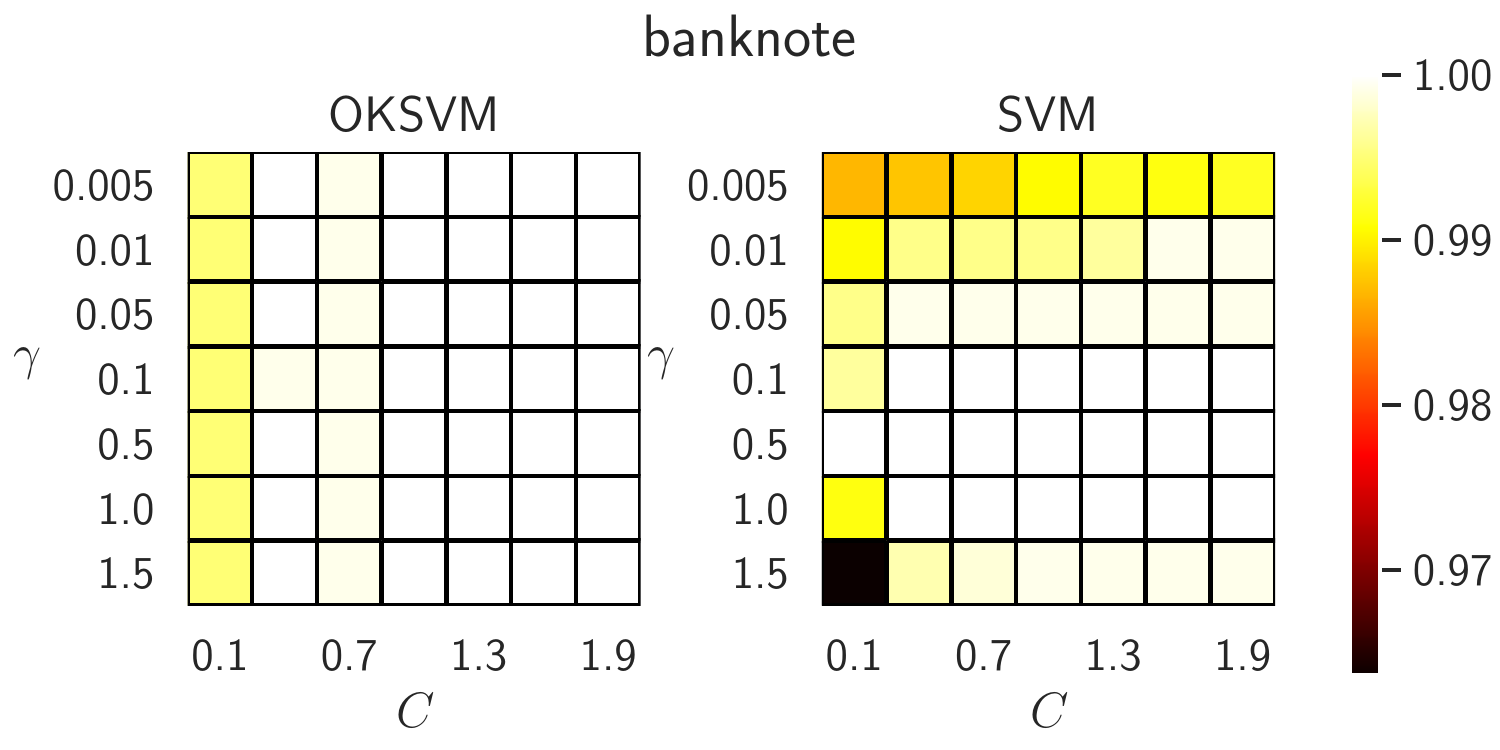} &
		\includegraphics[width=0.45\linewidth]{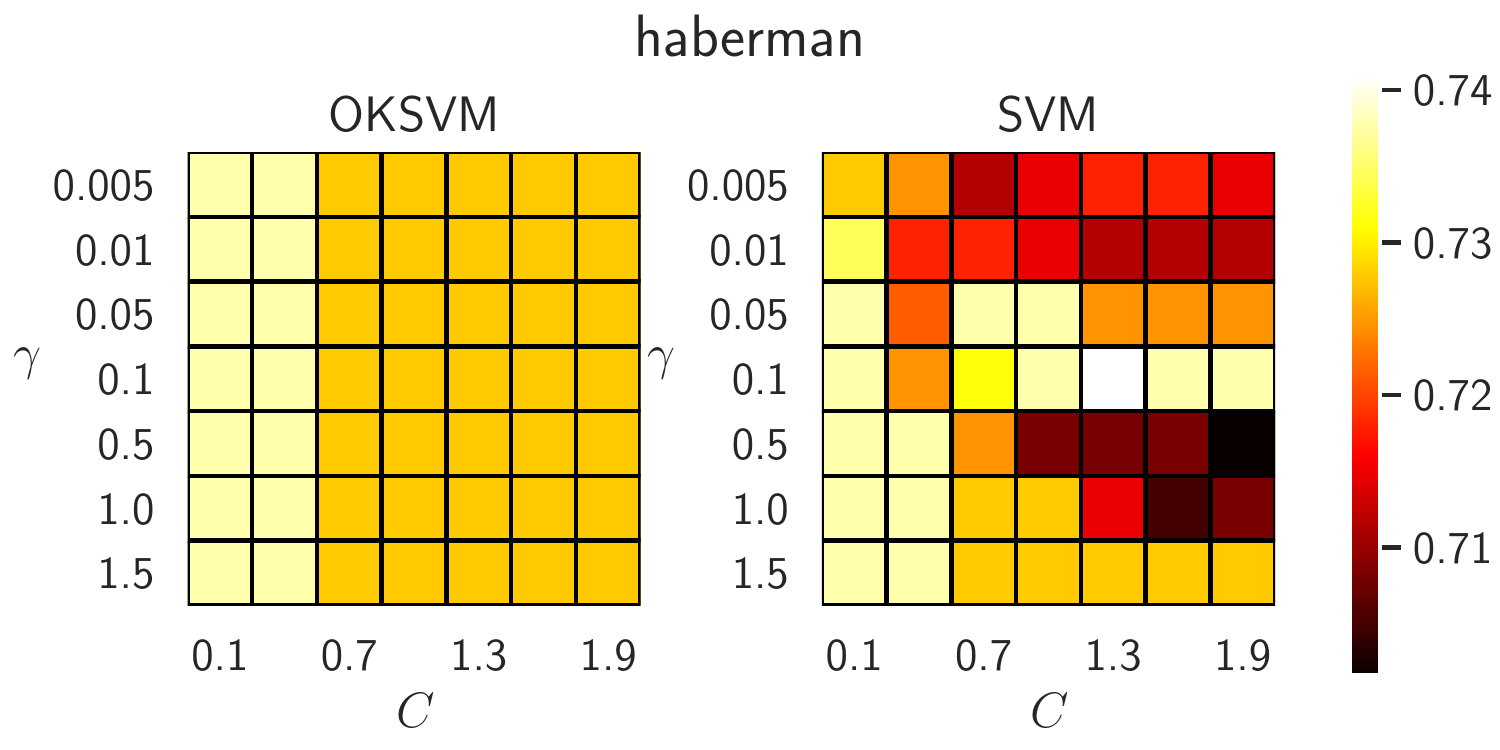} 
	\end{tabular}
\caption{$F1$ heatmaps of \textit{OKSVM} and \textit{SVM} with real datasets. The average performance of 100 runs on the test set for each configuration is represented varying $\gamma$ and $C$. Lighter tones are better.}
\label{fig:heatmap_Fm_comparison_realData}
\end{figure}

Following experiments from synthetic data subsection, first both methods were tested by fixing the hyperparameters. Some results are shown in Figure \ref{fig:heatmap_Fm_comparison_realData}. Generally, \textit{OKSVM} does not highly depend on $\gamma$ and outperforms \textit{SVM} and achieves the best results for each value of $C$. Moreover, for most datasets, the maximum performance obtained by \textit{OKSVM} is higher than those yielded by \textit{SVM}.

\setlength\tabcolsep{6pt}
\begin{table}[!htbp]
    \caption{Performance comparisons of \textit{OKSVM} and \textit{SVM} with real datasets. The class distribution and the average measures of 5-fold cross validation are shown. Best results are marked in \textbf{bold}.}
    \label{tab:real_data_cv}
    \centering
    \scalebox{0.83}{
    \begin{tabular}{l|cc|ccccc|ccccc}
        \toprule
        Method & \multicolumn{2}{c|}{Classes} & \multicolumn{5}{c}{\textit{SVM}} & \multicolumn{5}{c}{\textit{OKSVM}}\\
        \midrule
        Dataset & $N_-$ & $N_+$ & \textit{Acc} & \textit{RC} & \textit{PR} & \textit{F1} & \textit{AUC} & \textit{Acc} & \textit{RC} & \textit{PR} & \textit{F1} & \textit{AUC}\\
        \midrule
        \textit{cleveland} & 160 & 137 & 0.623 & 0.579 & 0.603 & 0.590 & 0.673 & \textbf{0.687} & \textbf{0.643} & \textbf{0.675} & \textbf{0.656} & \textbf{0.736}\\
        \textit{wdbc} & 357 & 212 & 0.895 & \textbf{0.924} & 0.819 & 0.867 & 0.940 & \textbf{0.928} & 0.900 & \textbf{0.908} & \textbf{0.902} & \textbf{0.969}\\
        \textit{breast-cancer} & 444 & 239 & \textbf{0.965} & \textbf{0.975} & \textbf{0.929} & \textbf{0.951} & \textbf{0.993} & 0.962 & \textbf{0.975} & 0.922 & 0.947 & 0.988\\
        \textit{iris} & 50 & 50 & 0.900 & 0.920 & 0.894 & 0.903 & 0.968 & \textbf{0.930} & \textbf{0.940} & \textbf{0.927} & \textbf{0.931} & \textbf{0.986}\\
        \textit{banknote} & 761 & 610 & 0.992 & \textbf{1.000} & 0.982 & 0.991 & \textbf{1.000} & \textbf{1.000} & \textbf{1.000} & \textbf{1.000} & \textbf{1.000} & \textbf{1.000}\\
        \textit{haberman} & 81 & 224 & 0.725 & 0.916 & \textbf{0.760} & 0.831 & \textbf{0.624} & \textbf{0.728} & \textbf{0.982} & 0.737 & \textbf{0.842} & 0.598\\
        \textit{winequality-red} & 744 & 855 & 0.643 & \textbf{0.995} & 0.600 & 0.749 & \textbf{0.788} & \textbf{0.654} & 0.970 & \textbf{0.611} & \textbf{0.750} & 0.785\\
        \bottomrule
    \end{tabular}
    }
\end{table}

In the second experiment, an optimization of the hyperparameters was done with a validation set (25\% of the training set) and \tablename\,\ref{tab:real_data_cv} sums up the average results of a 5-fold cross-validation. The breast-cancer dataset shows that traditional SVM works better if the hyperparameter are tuned, although the heatmap of \figurename\,\ref{fig:heatmap_Fm_comparison_realData} reveals that \textit{OKSVM} yields great outputs whatever the $\gamma$ and $C$ values are. \textit{OKSVM} outperforms \textit{SVM} for the other datasets in most of the measures.

\section{Conclusions} \label{sec:Conclusions}

In this work, a new methodology to learn the spread hyperparameter of the RBF kernel of a Support Vector Machine has been proposed. The method minimizes an upper bound of the misclassification probability by applying gradient descent on the hyperparameter. Steps of gradient descent are interleaved with steps of optimization of the dual form parameters of the SVM to ensure the joint optimization of the adjustable parameters of the SVM. Specific provisions are made to avoid pitfalls in the optimization of the hyperparameter. Experimental results have been reported with synthetic and real datasets. Our approach has shown superior performance when compared with the standard SVM. Our method finds an appropriate value for the hyperparameter irrespective of the initial value. Therefore, it is suitable for reliable automated tuning of SVM classifiers.

\section*{Broader Impact}

This work is expected to enhance the performance of classification systems based on Support Vector Machines. All organizations that employ this kind of machine learning techniques might benefit from it. There are no people who are likely to be put at disadvantage from this research. Failure in classifications systems might cause severe consequences depending on the application, although this is true for any classification scheme, not only ours. At this point in our research, we have not detected that our method leverages any bias in the data.

\small
\bibliographystyle{unsrt}

\end{document}


\maketitle

\section*{Supplementary material} \label{sec:Supplementary}

The main paper presents part of the experiments, focusing on dimensions 2, 4, and 8. Therefore, this document collects the remaining results for the different synthetic and real datasets that have been used to establish a fair comparison against the competitor method.

In addition, the source code is provided as supplementary material to keep a high level of transparency in our work. This code will be published in GitHub in case of acceptance. 

\subsection*{Synthetic datasets}

First of all, recall that the tuned configurations for the number of features are $dim=\{2, 3, 4, 5, 6, 7, 8\}$. Experiments for each tuned configuration have been repeated 100 times ($R=100$) with randomly generated datasets, and the average performance has been considered to establish the efficiency comparisons. Each dataset was divided into training and testing sets, with 50\% of the samples each.

Here, results for $dim=\{3, 5, 6, 7\}$ are presented and discussed. Again, each type of experiment is separated into different sections for the sake of clarity.

\newpage

\subsubsection*{Fixed hyperparameters comparison}

These experiments compare the performance of \textit{SVM} and \textit{OKSVM} methods when they have the same tuned hyperparameter configuration. The set of tested parameters are the same as the ones mentioned in the paper.

\figurename\,\ref{fig:heatmap_Fm_comparison} depicts the outputs for dimensions 3 and 5. The $F1$-score of each method for each value of $C$, $\gamma$, and the class separation $sep$ is presented. As it can be observed, the tendency is similar to the results presented for dimensions 2 and 4. The higher the dimension, the more effective is our proposal. Note that for $dim=5$, the differences are remarkable since the performance obtained by \textit{OKSVM} method usually is equal or better than the best performance yielded by the traditional SVM for a fixed value of $sep$ (view the output by columns). That is due to the precise optimization process of \textit{OKSVM} that allows us to find the best $\gamma$ value that minimized the classification problem. 

\setlength\tabcolsep{1.5pt} 
\newcolumntype{C}{ >{\centering\arraybackslash} m{3.4cm} }
\newcolumntype{D}{ >{\centering\arraybackslash} m{1.5cm} }

\begin{figure}
\centering
	\begin{tabular}{c|D|CCCc}
	    $dim$ & Method & $C=0.5$ & $C=1.0$ & $C=1.5$ \\
	    \midrule
		\multirow{2}{*}[-3em]{3} & OKSVM & \includegraphics[width=1\linewidth]{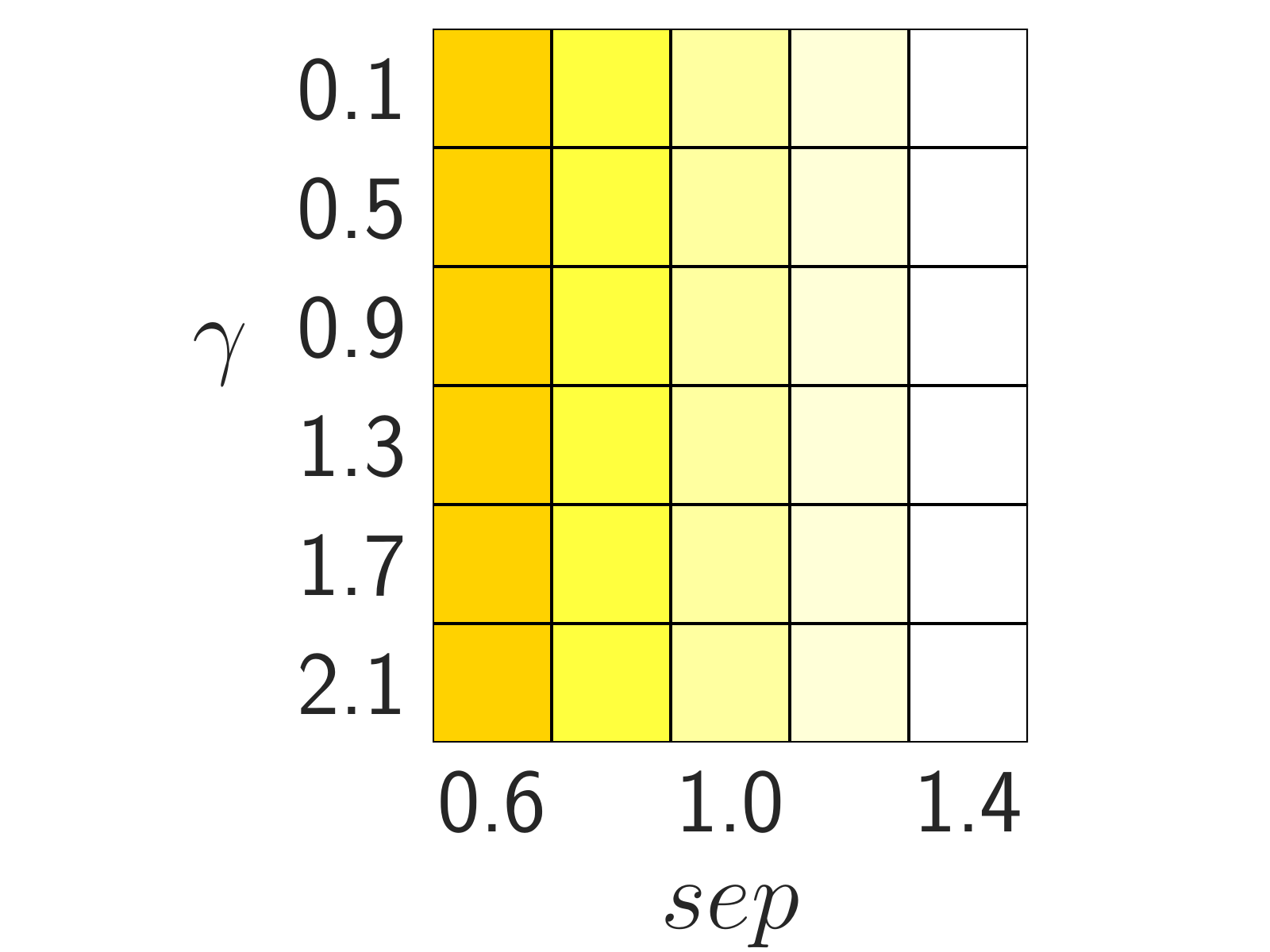} &
		\includegraphics[width=1\linewidth]{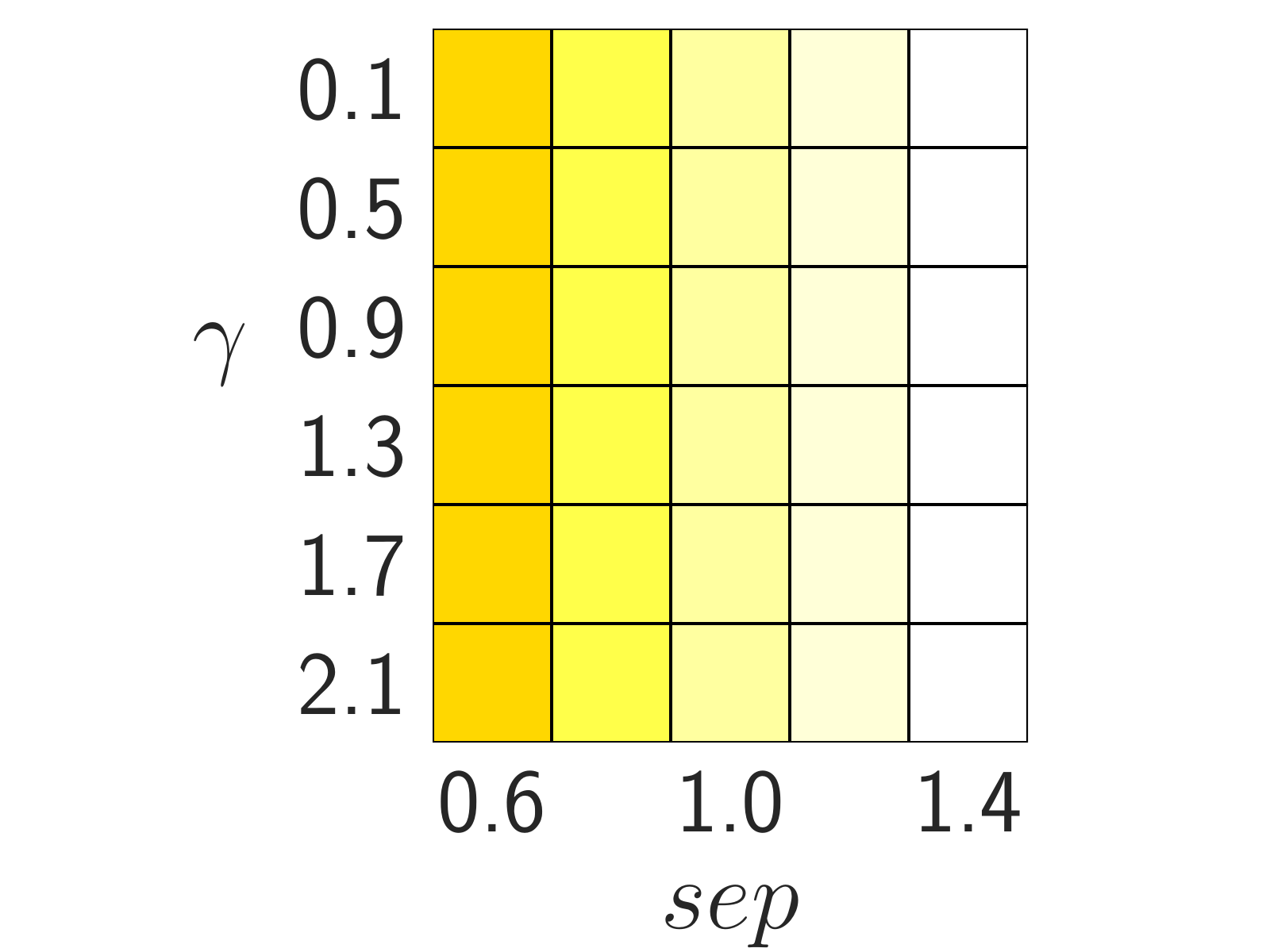} &
		\includegraphics[width=1\linewidth]{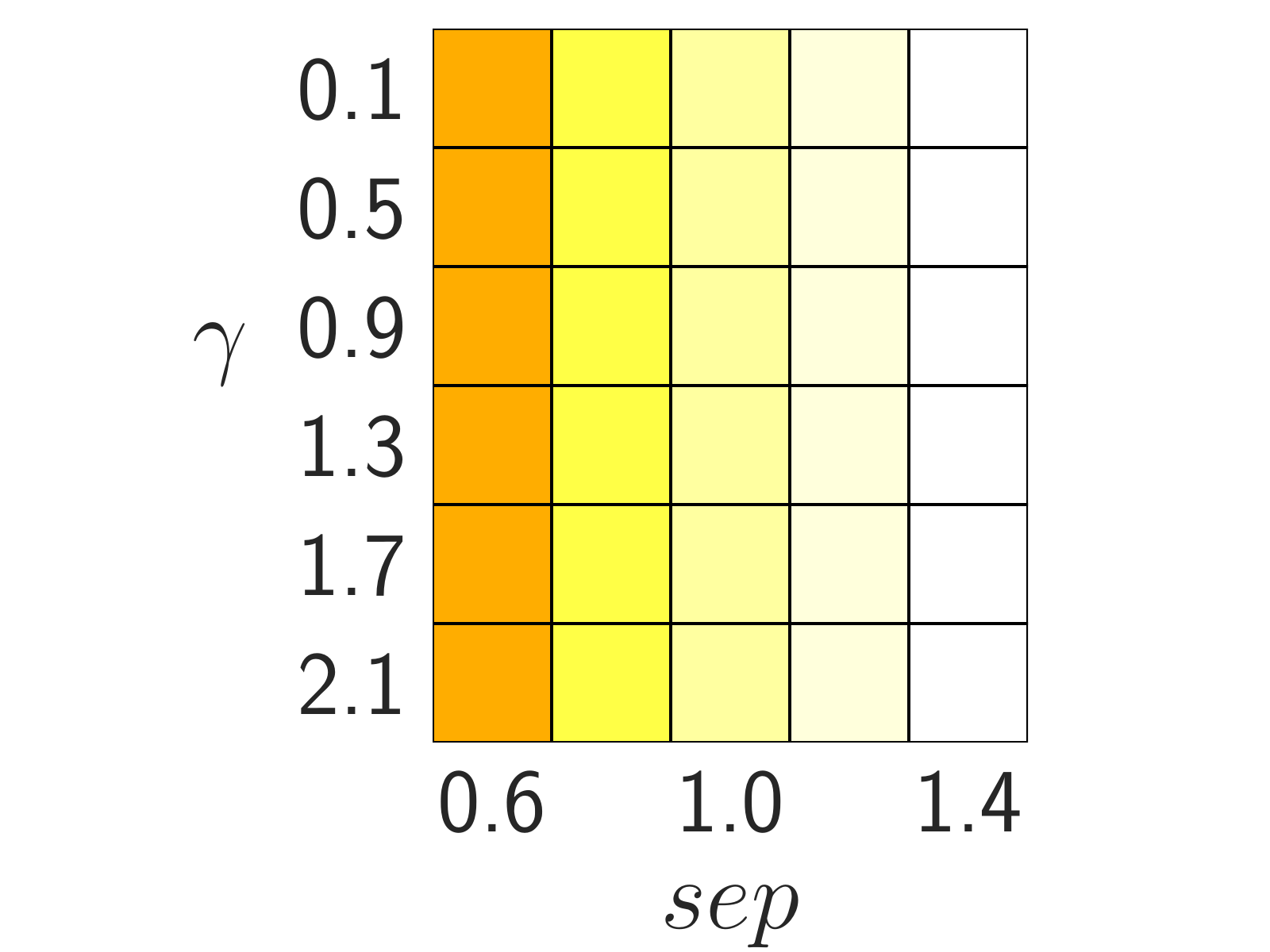} & \multirow{4}{*}[-1em]{\includegraphics[width=0.075\linewidth]{images/legend_FmHot_vertical.pdf}}\\
		& SVM  & \includegraphics[width=1\linewidth]{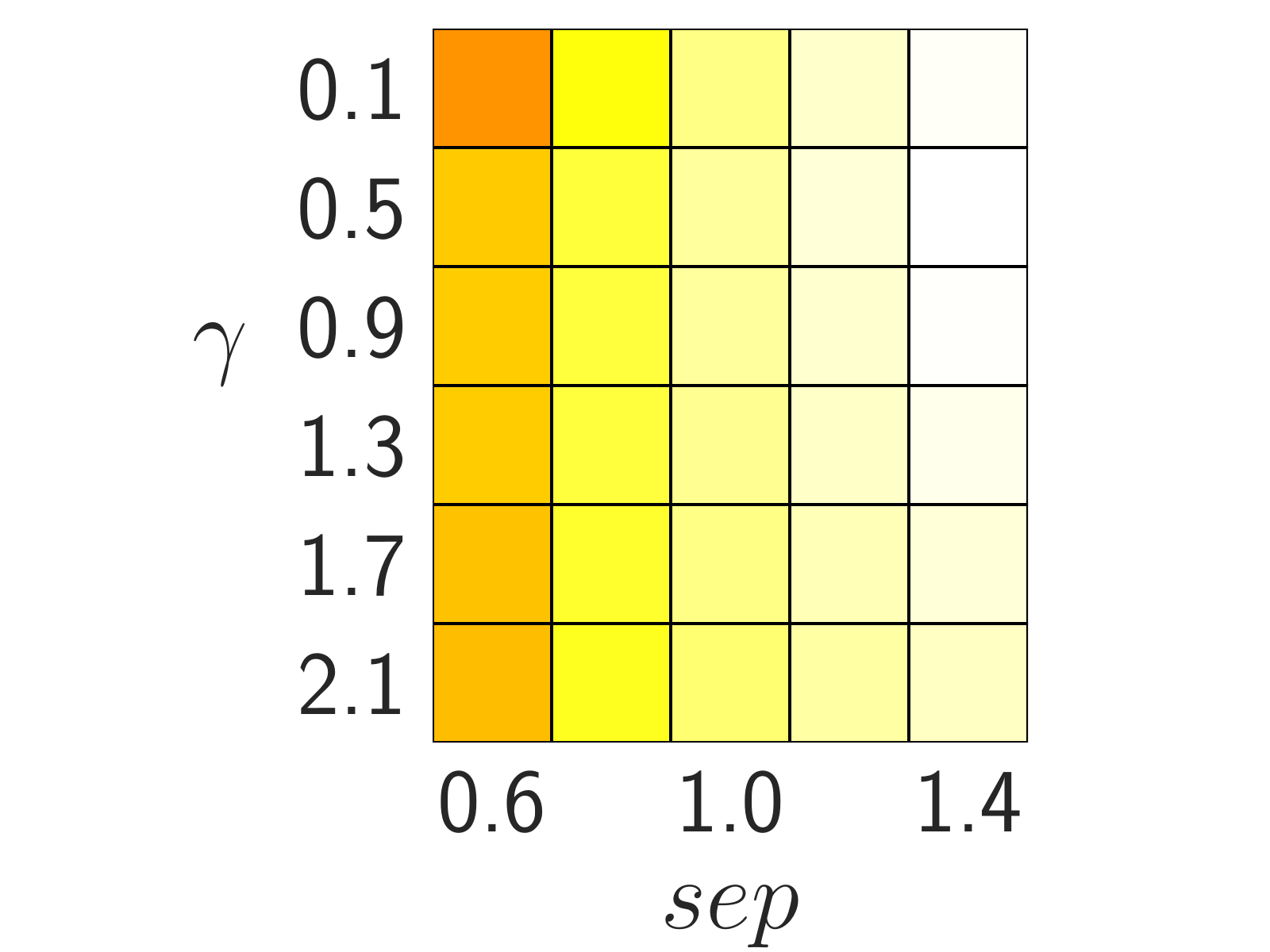} &
		\includegraphics[width=1\linewidth]{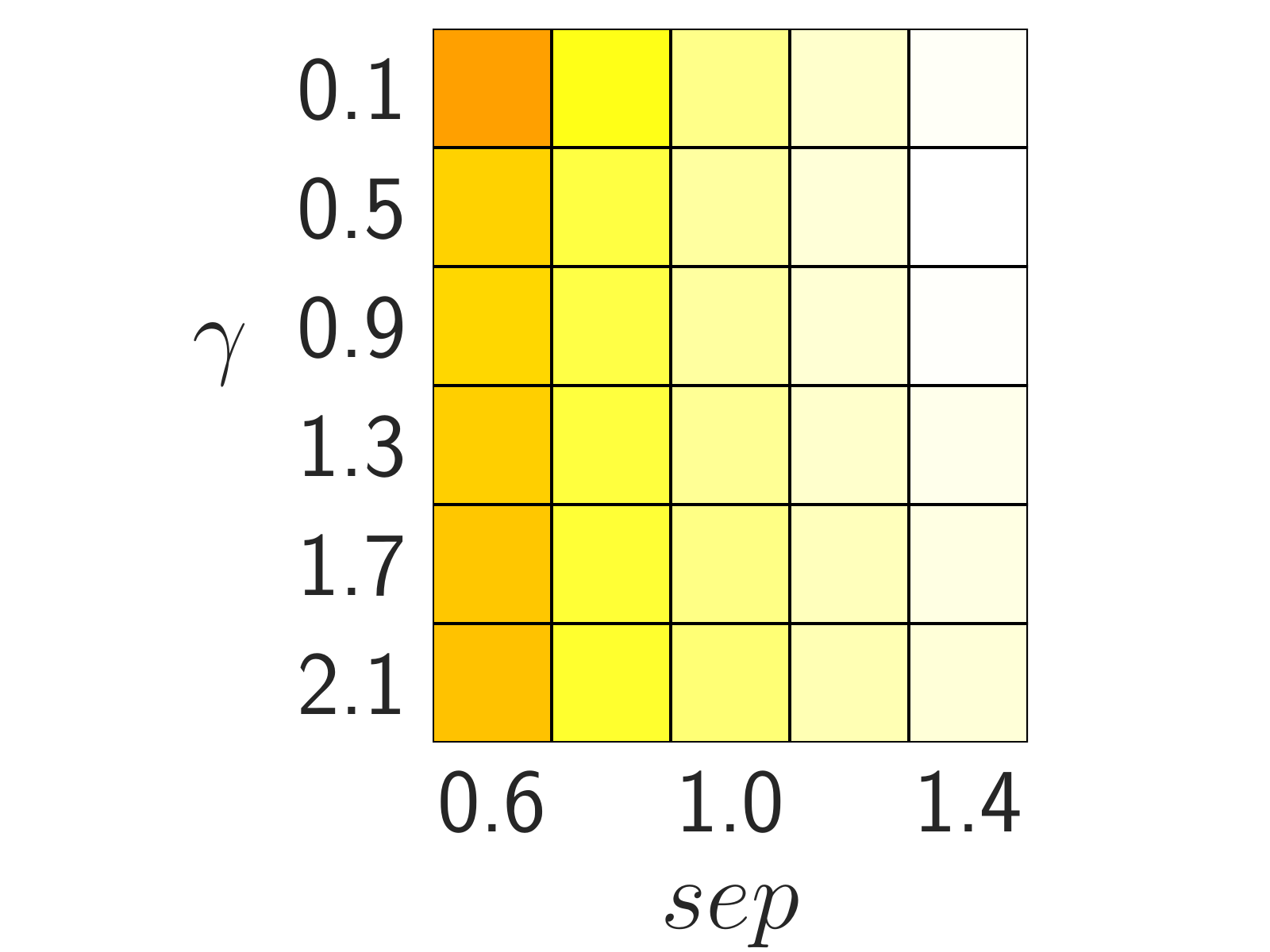} &
		\includegraphics[width=1\linewidth]{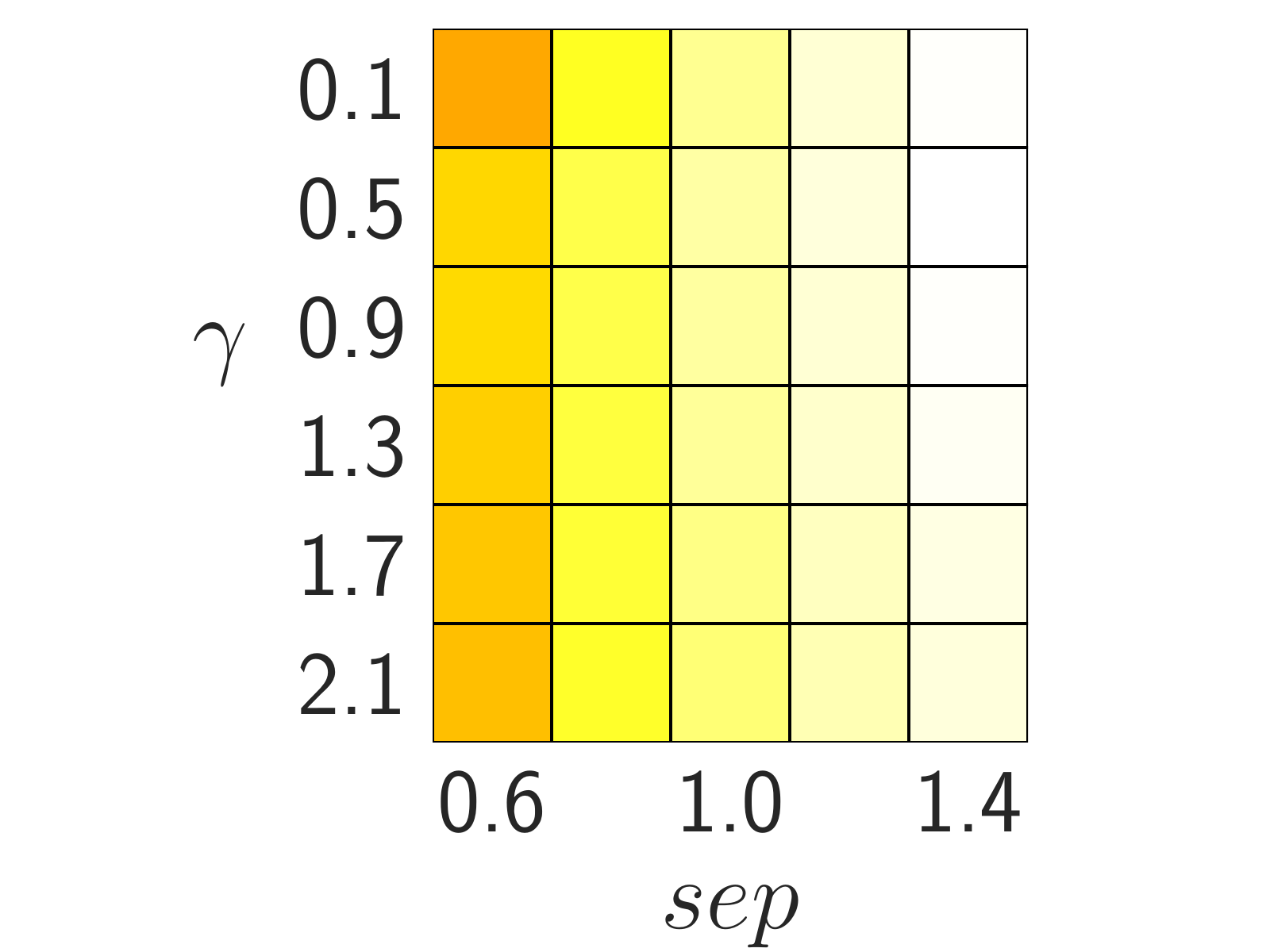} & \\
		\cmidrule{1-5}
		\multirow{2}{*}[-3em]{5} & OKSVM & \includegraphics[width=1\linewidth]{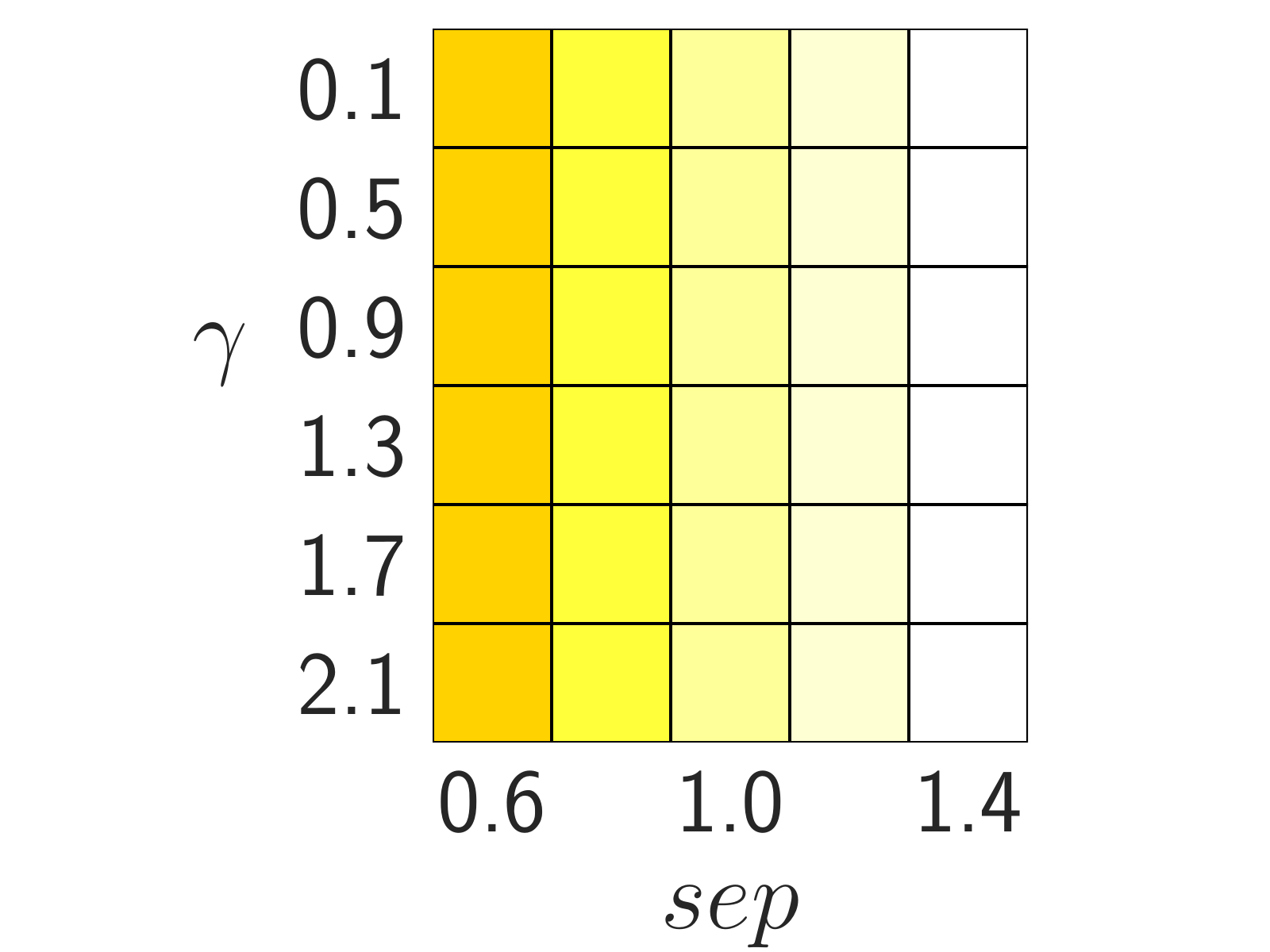} &
		\includegraphics[width=1\linewidth]{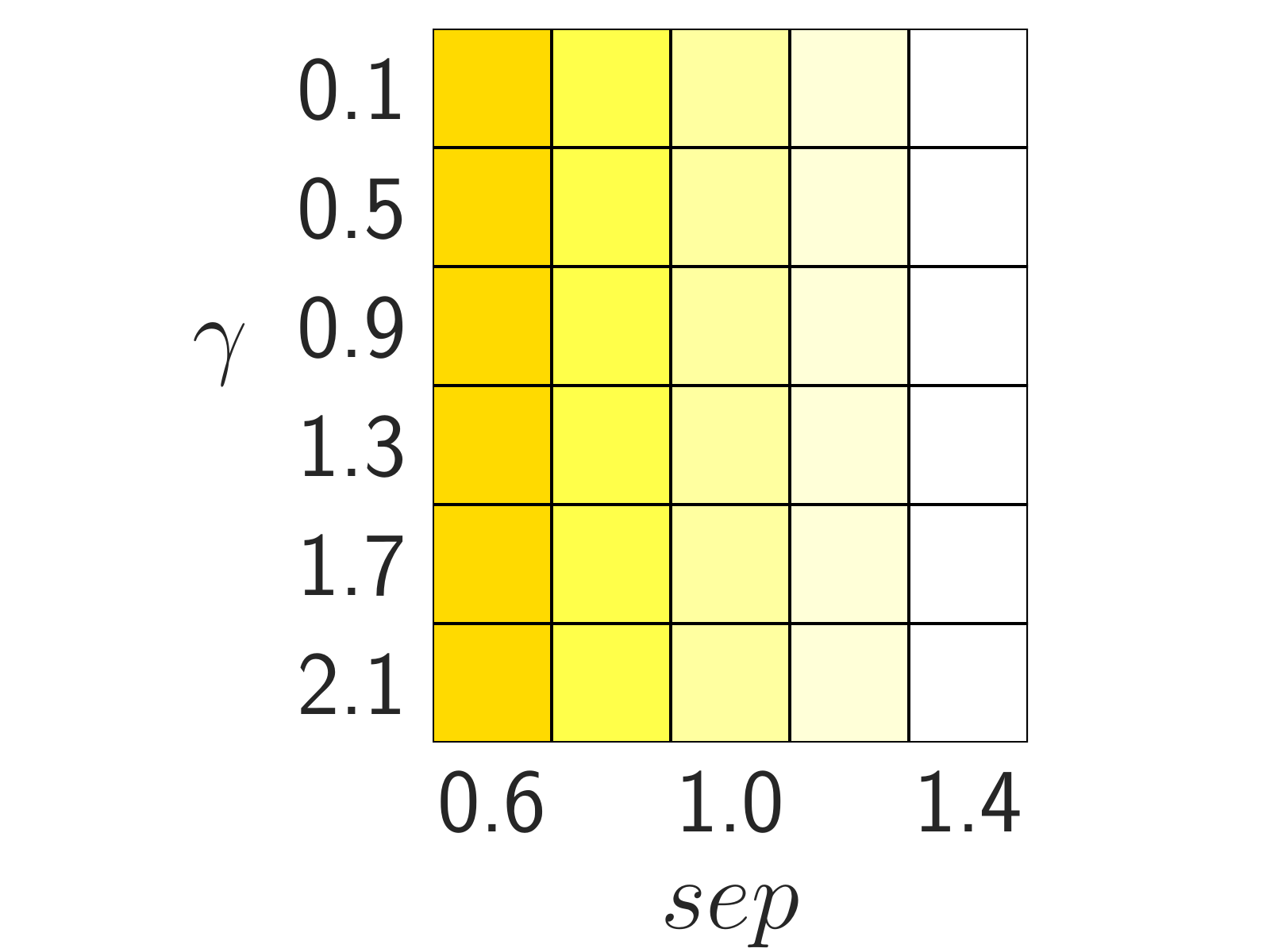} &
		\includegraphics[width=1\linewidth]{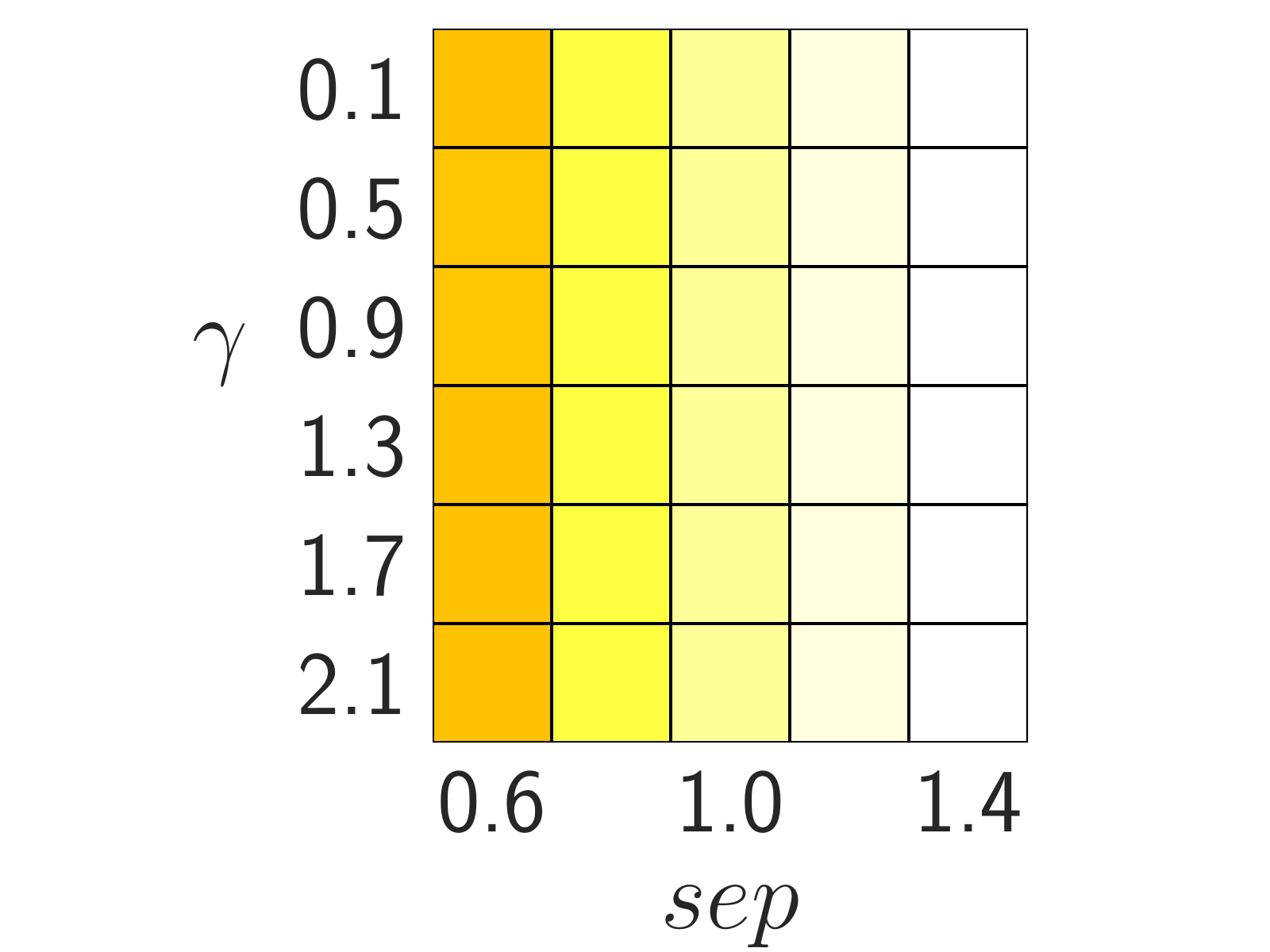} & \\
		 & SVM  & \includegraphics[width=1\linewidth]{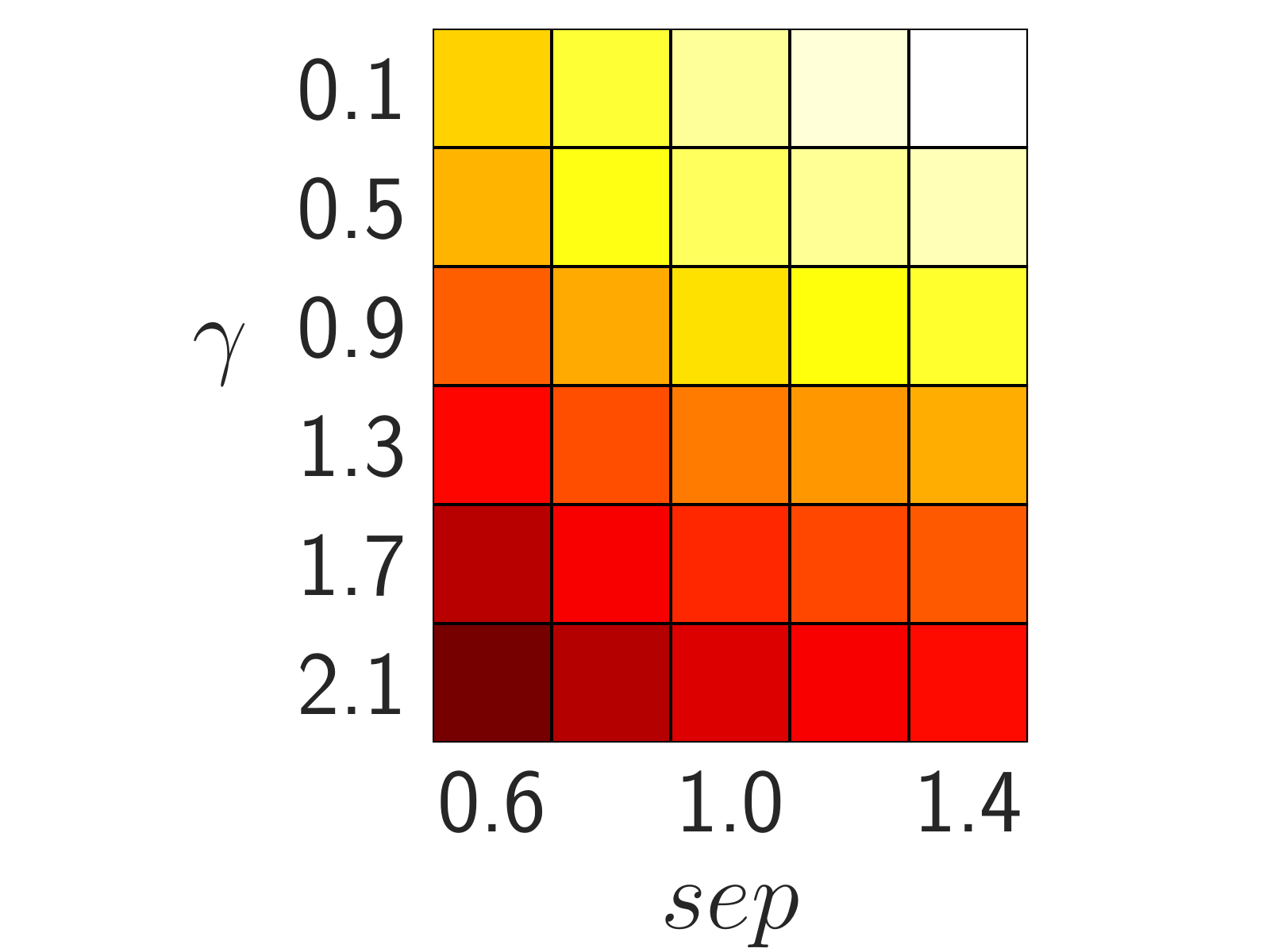} &
		\includegraphics[width=1\linewidth]{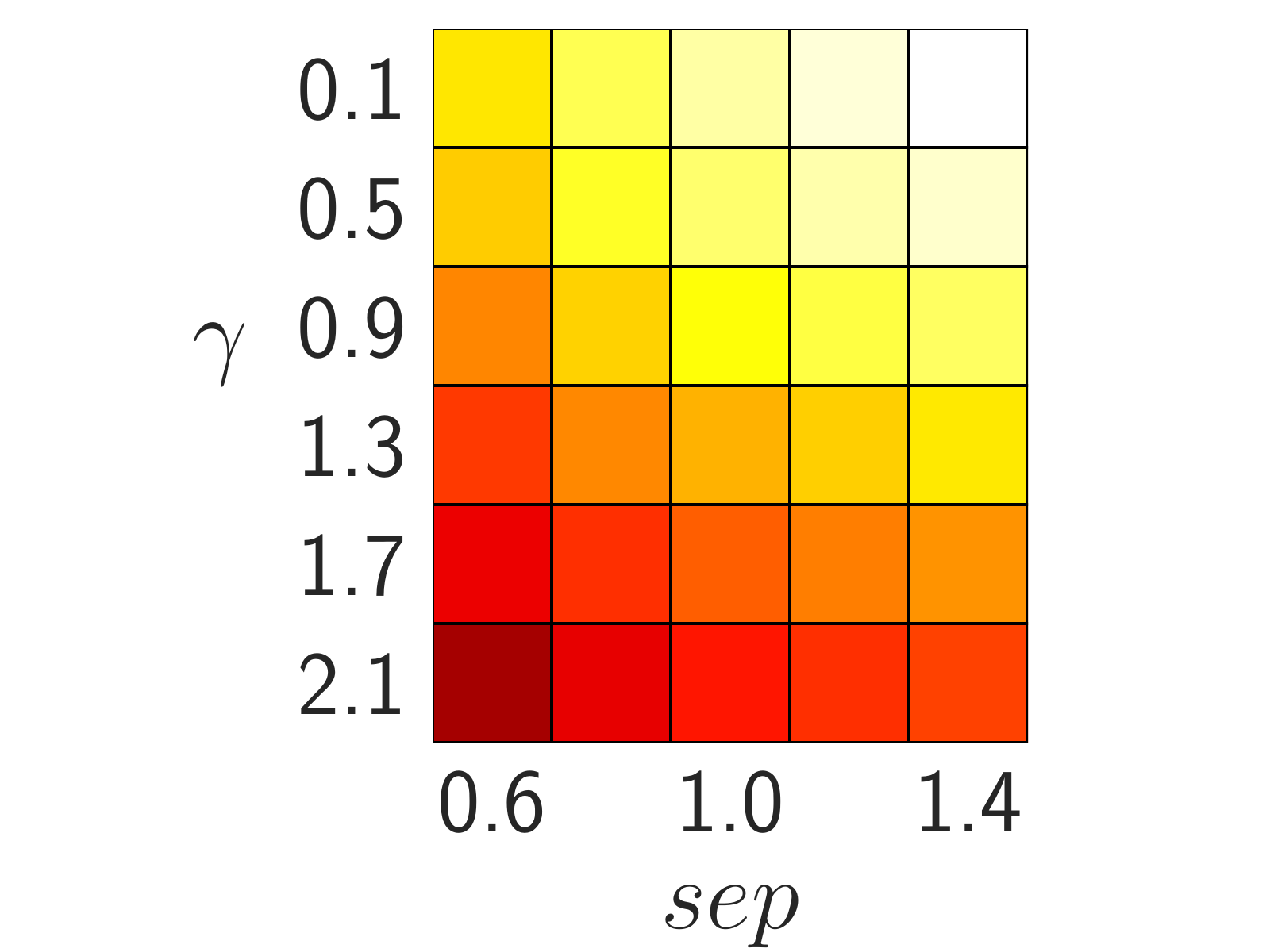} &
		\includegraphics[width=1\linewidth]{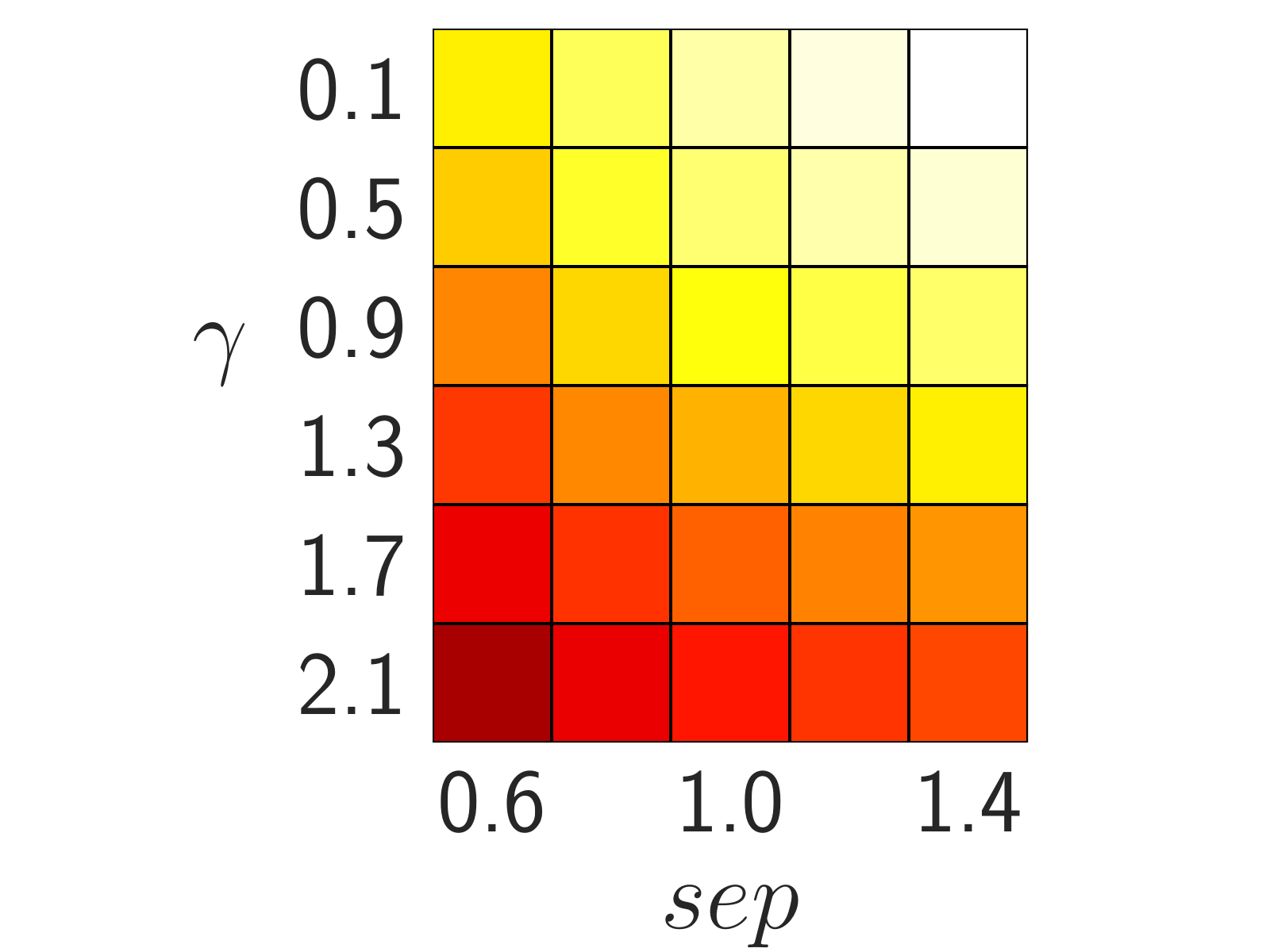} & 
	\end{tabular}
\caption{Performance ($F1$-score) of \textit{OKSVM} and \textit{SVM}. Heatmaps are organized in the figure according to their tuned configuration where each row indicated the number of dimensions $dim$ of the datasets and the method, while columns indicated the parameter $C$. Each heatmap exhibits the average performance of the tuned configuration where each row of the heatmap represents the parameter $gamma$ and columns represent the separation $sep$ of the clusters of the datasets. Lighter is better.}
\label{fig:heatmap_Fm_comparison}
\end{figure}

\newpage

\begin{figure}
\centering
	\begin{tabular}{c|D|CCCc}
	    $dim$ & Method & $C=0.5$ & $C=1.0$ & $C=1.5$ \\
	    \midrule
		\multirow{2}{*}[-3em]{6} & OKSVM & \includegraphics[width=1\linewidth]{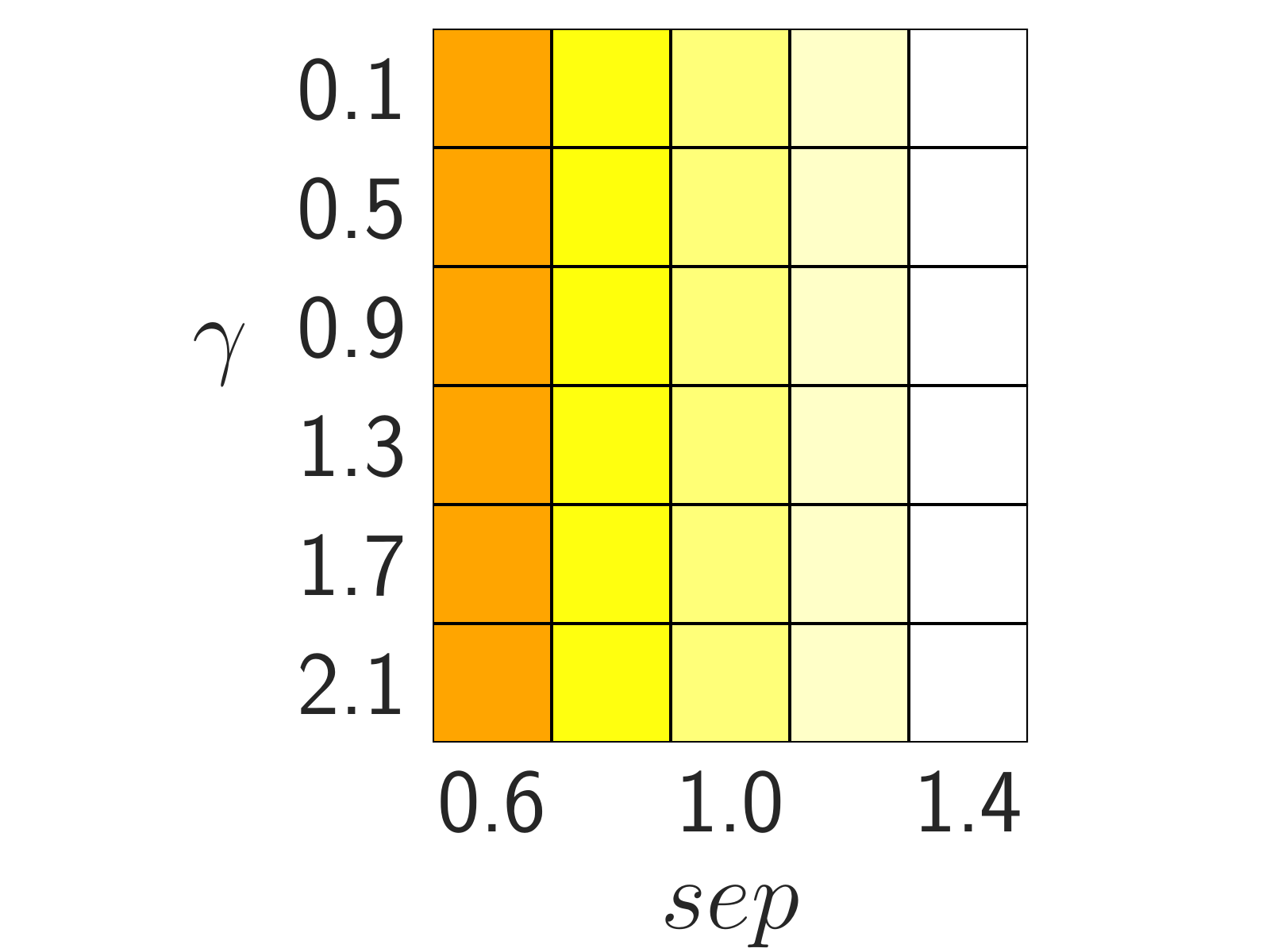} &
		\includegraphics[width=1\linewidth]{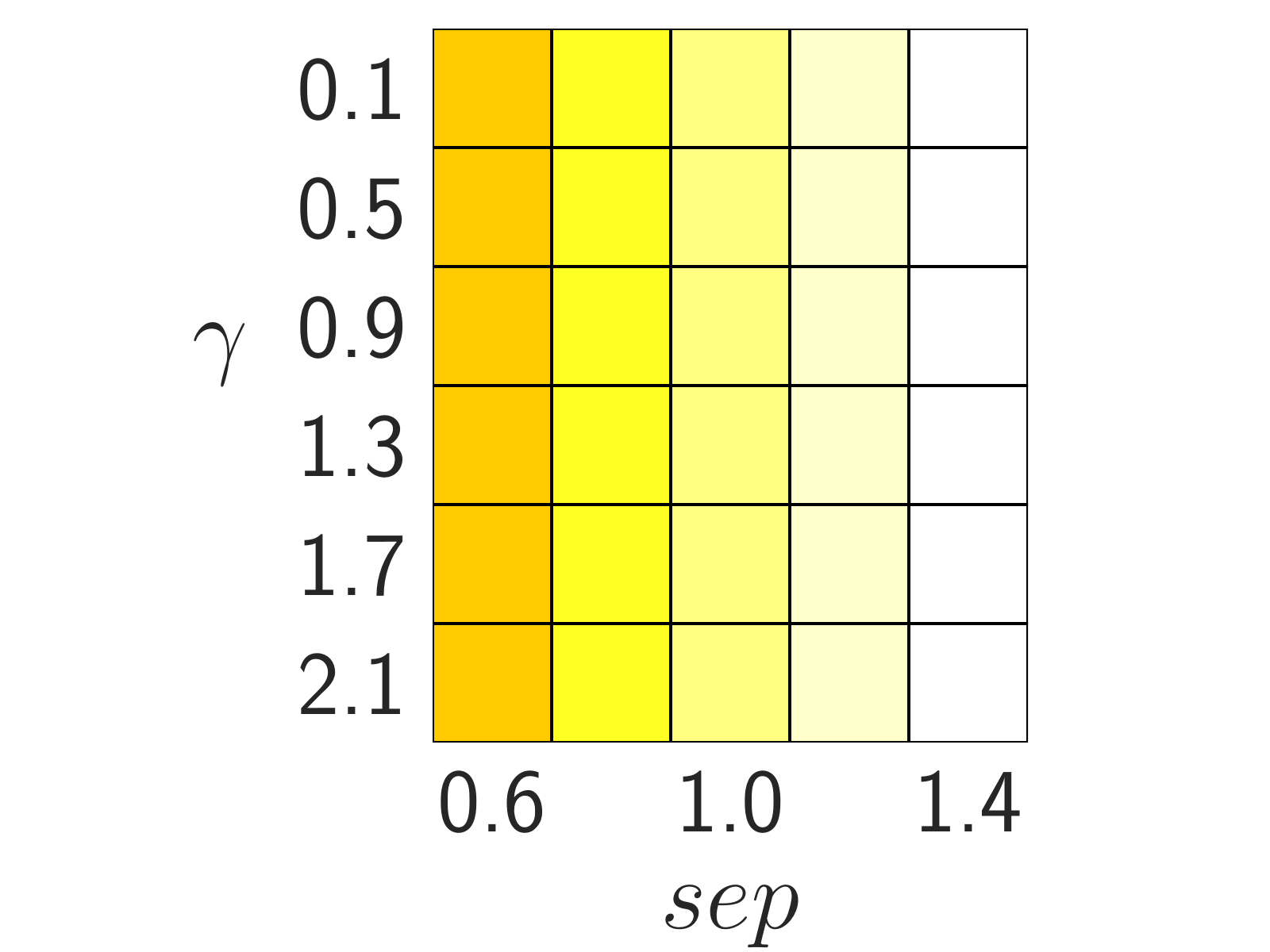} &
		\includegraphics[width=1\linewidth]{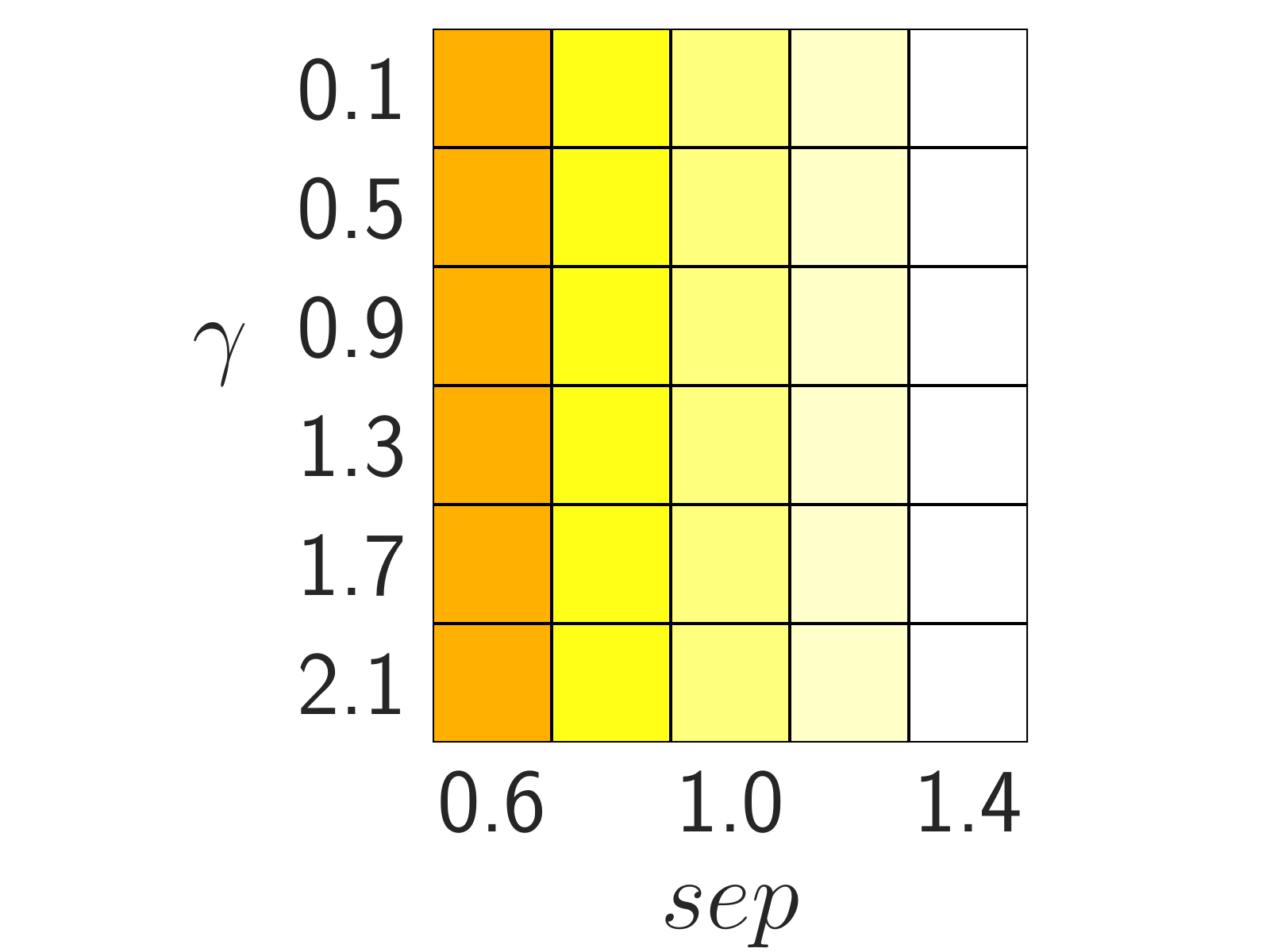} & \multirow{4}{*}[-1em]{\includegraphics[width=0.075\linewidth]{images/legend_FmHot_vertical.pdf}}\\
		& SVM  & \includegraphics[width=1\linewidth]{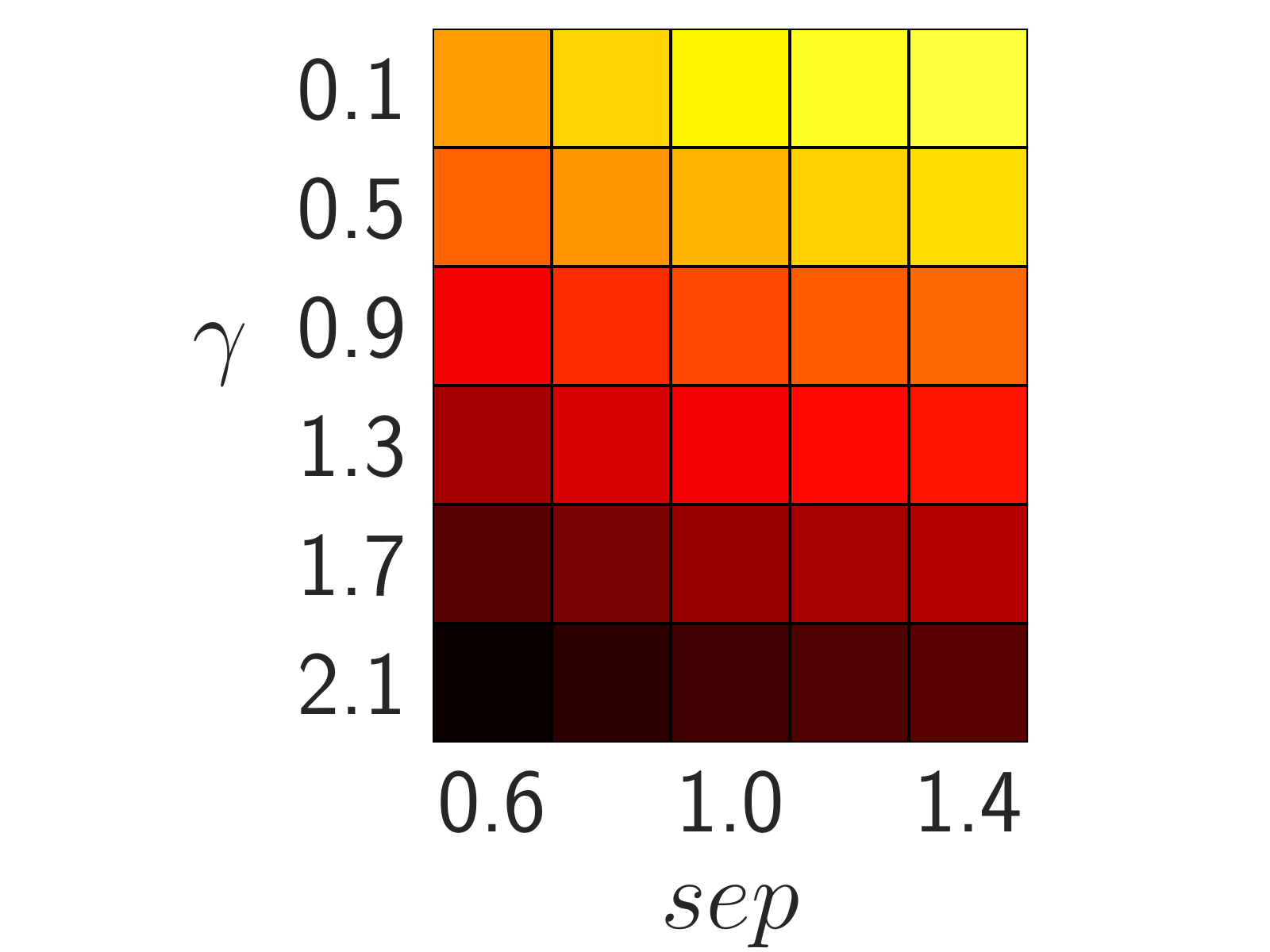} &
		\includegraphics[width=1\linewidth]{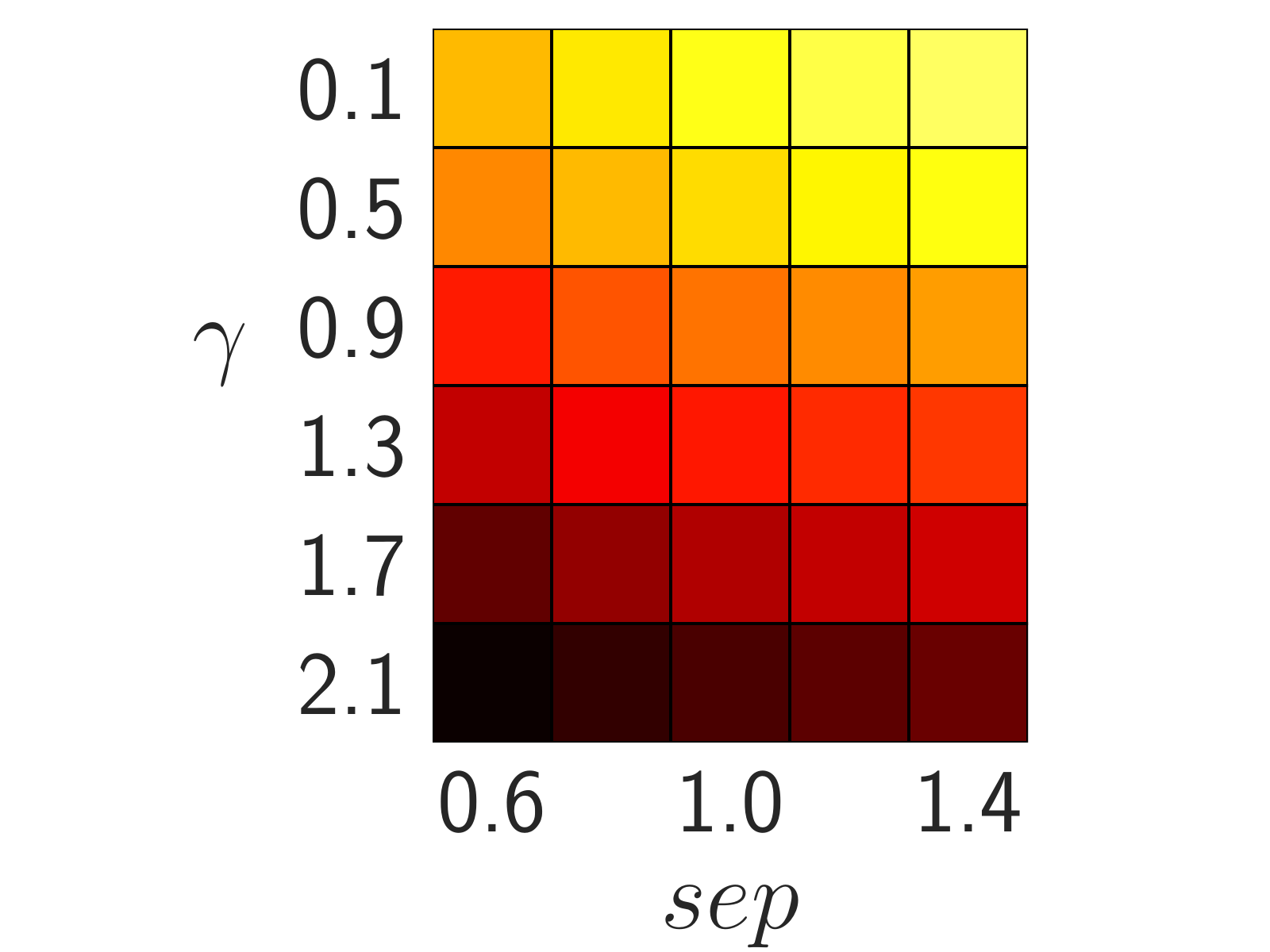} &
		\includegraphics[width=1\linewidth]{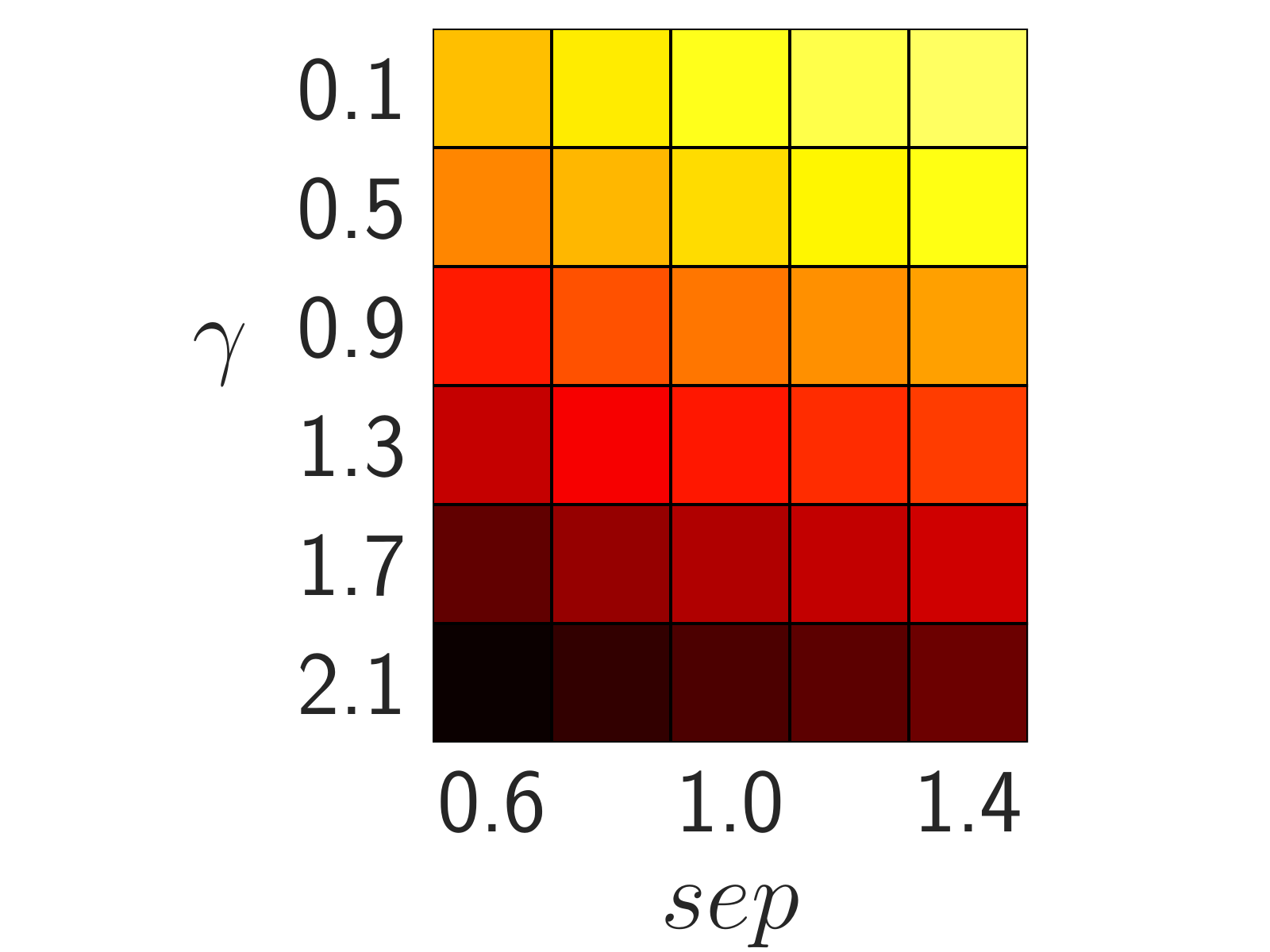} & \\
		\cmidrule{1-5}
		\multirow{2}{*}[-3em]{7} & OKSVM & \includegraphics[width=1\linewidth]{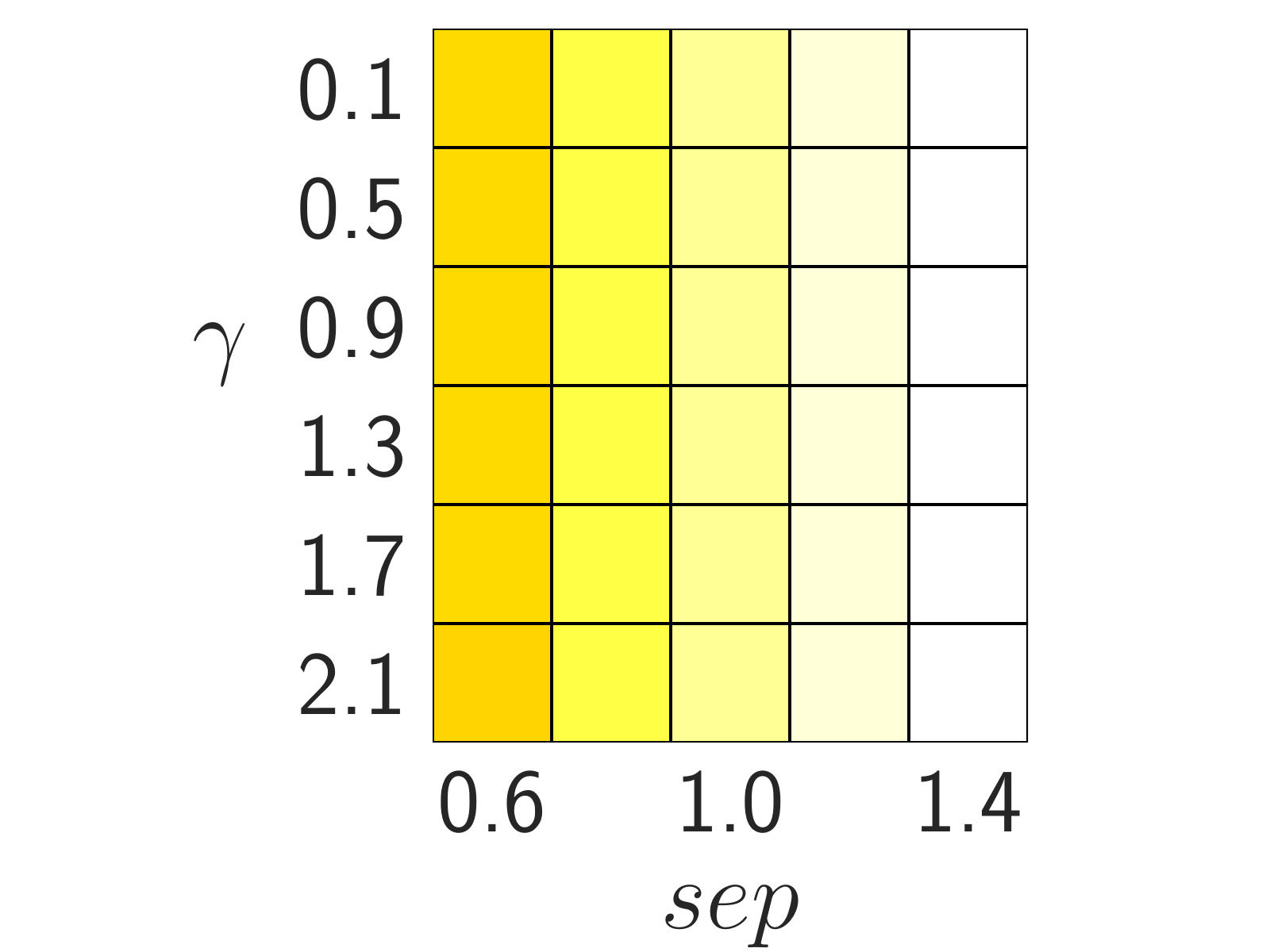} &
		\includegraphics[width=1\linewidth]{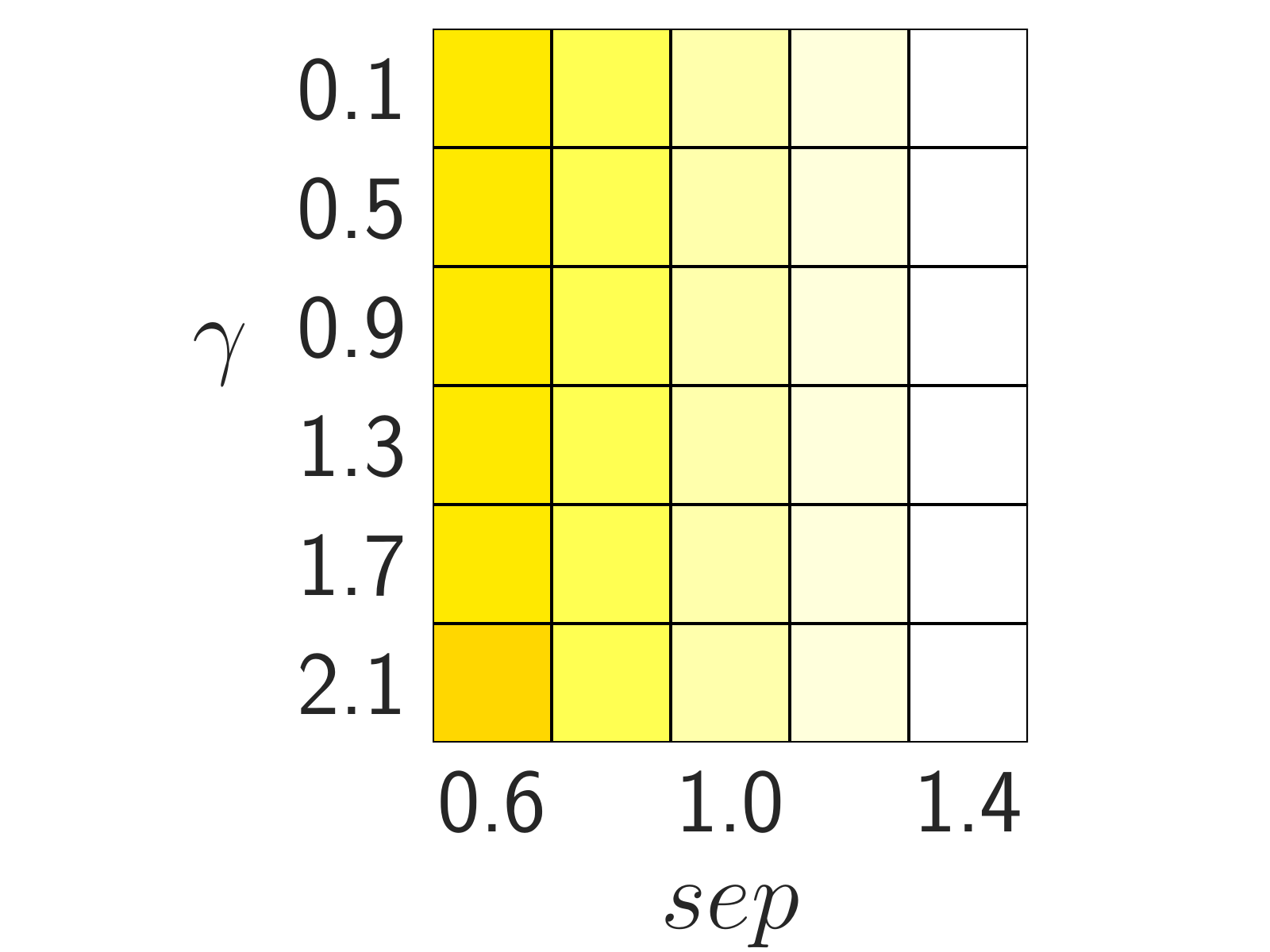} &
		\includegraphics[width=1\linewidth]{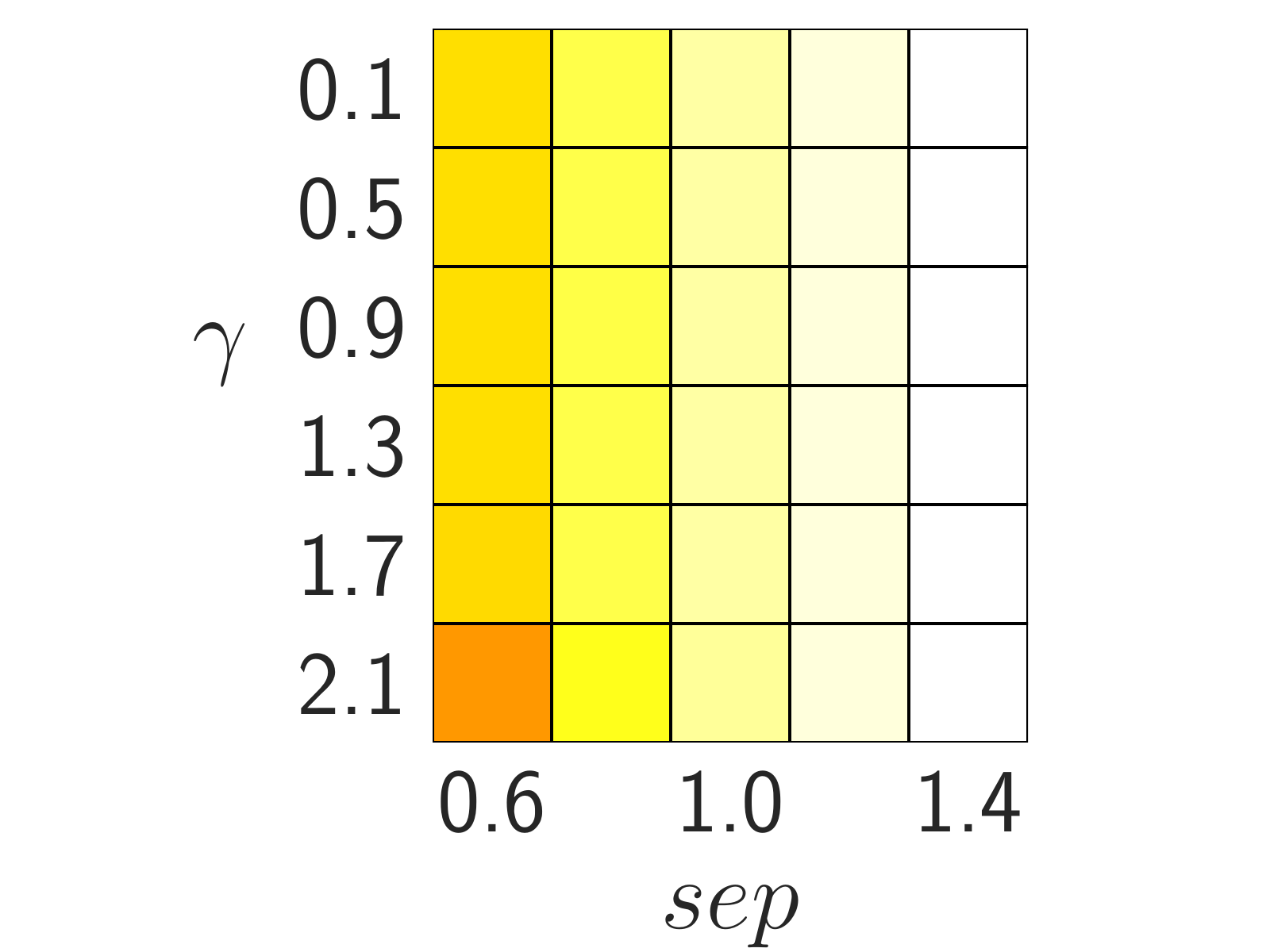} & \\
		 & SVM  & \includegraphics[width=1\linewidth]{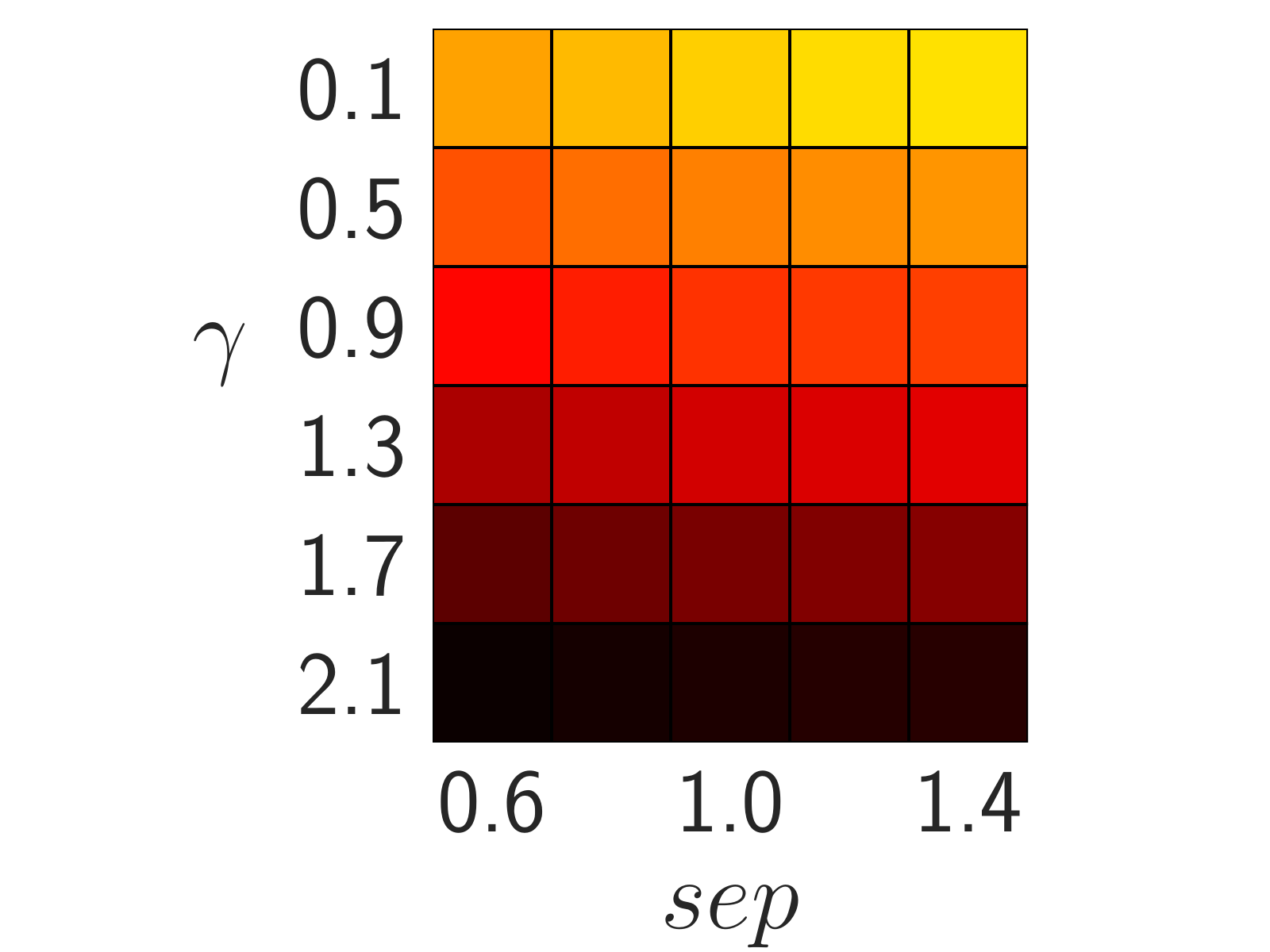} &
		\includegraphics[width=1\linewidth]{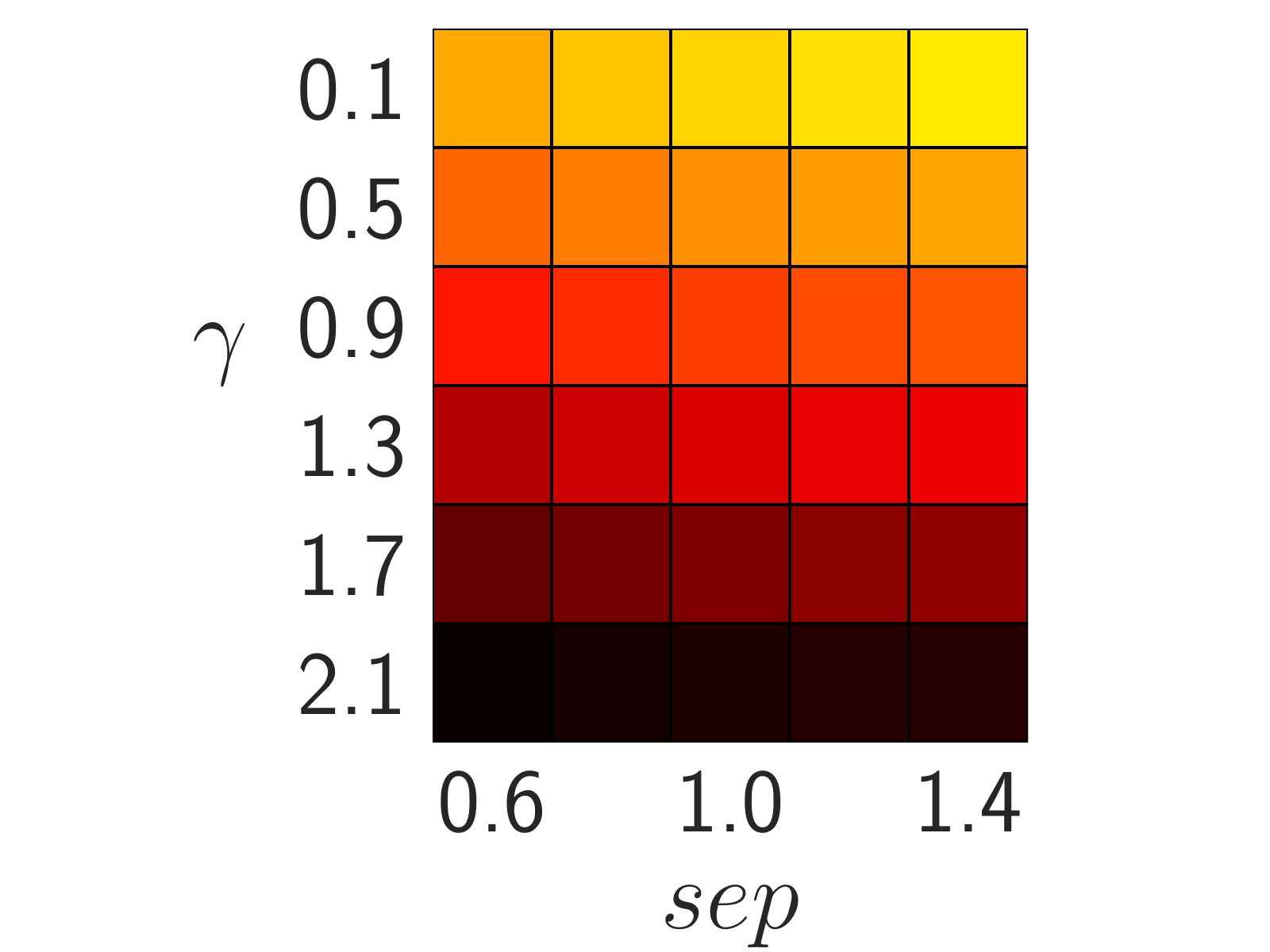} &
		\includegraphics[width=1\linewidth]{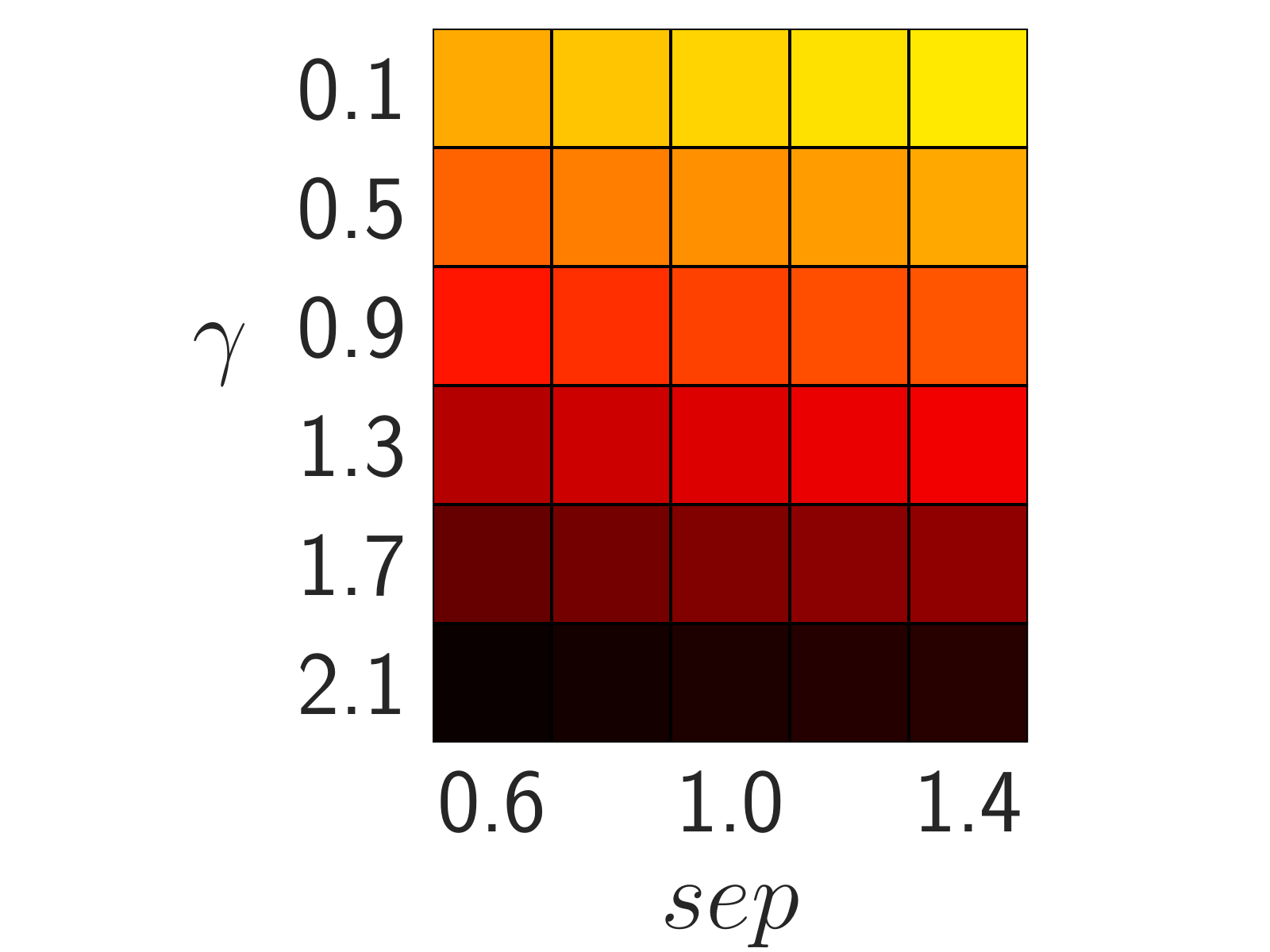} & 
	\end{tabular}
\caption{Performance ($F1$-score) of \textit{OKSVM} and \textit{SVM}. Heatmaps are organized in the figure according to their tuned configuration where each row indicated the number of dimensions $dim$ of the datasets and the method, while columns indicated the parameter $C$. Each heatmap exhibits the average performance of the tuned configuration where each row of the heatmap represents the parameter $gamma$ and columns represent the separation $sep$ of the clusters of the datasets. Higher is better.}
\label{fig:heatmap_Fm_comparison2}
\end{figure}

\figurename\,\ref{fig:heatmap_Fm_comparison2} depicts the outputs for dimensions 6 and 7. In this case, the differences are even higher because the classification problem increases in complexity. For any value of the tested parameters, \textit{OKSVM} outperforms \textit{SVM} method, generating the best measures when the class separation is higher (right part of the heatmap). The regularization term $C$ has not a great impact on the overall results, although the last column ($C=1.5$) shows that the performance is worse (darker color). This behavior agrees with the theoretical effect of this hyperparameter, that is, the higher values, the more misclassification.

\newpage

\begin{figure}
\centering
	\begin{tabular}{D|CCCc}
	    $dim$ & $C=0.5$ & $C=1.0$ & $C=1.5$ & \\
	    \midrule
		3 & \includegraphics[width=1\linewidth]{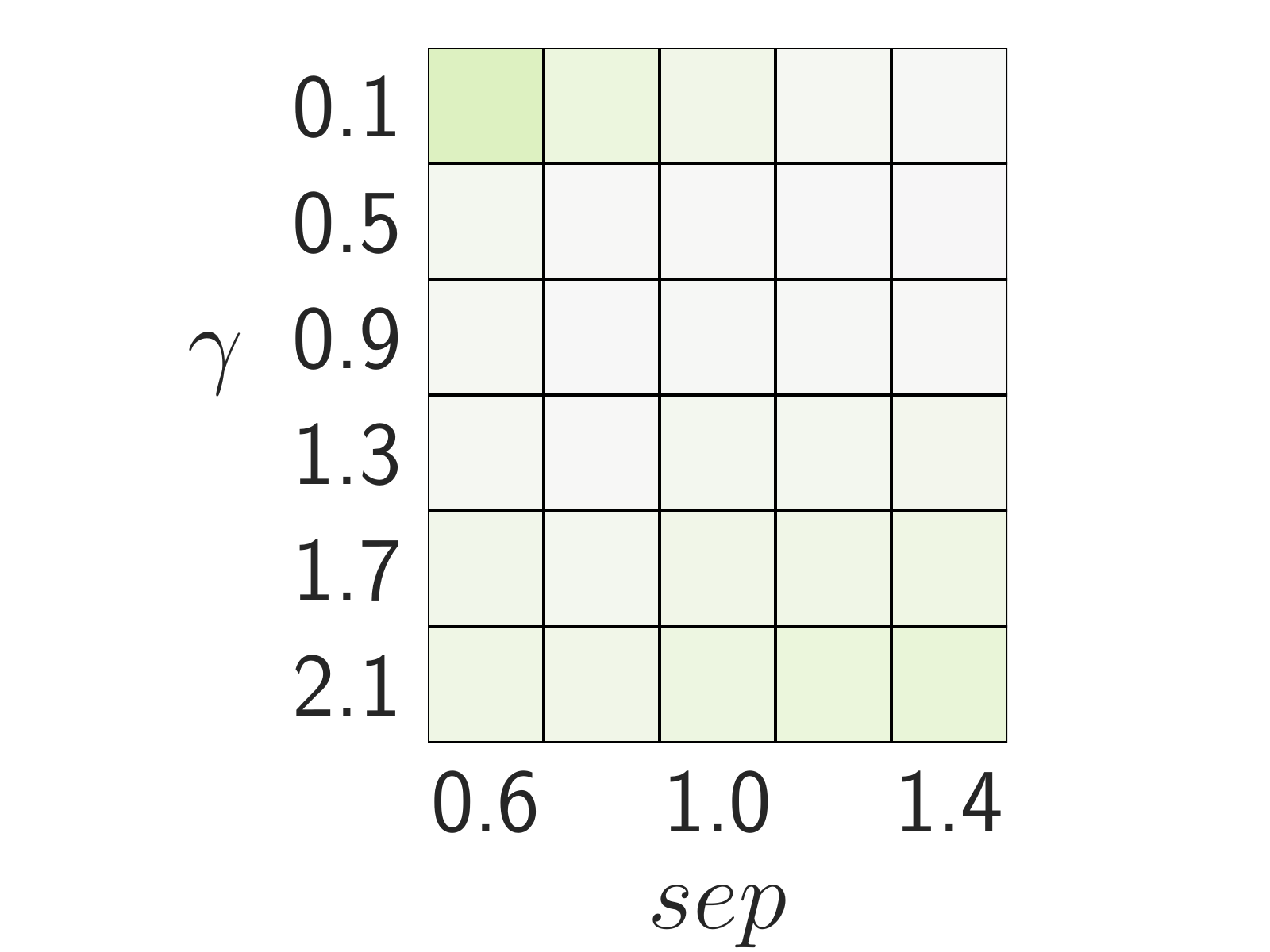} &
		\includegraphics[width=1\linewidth]{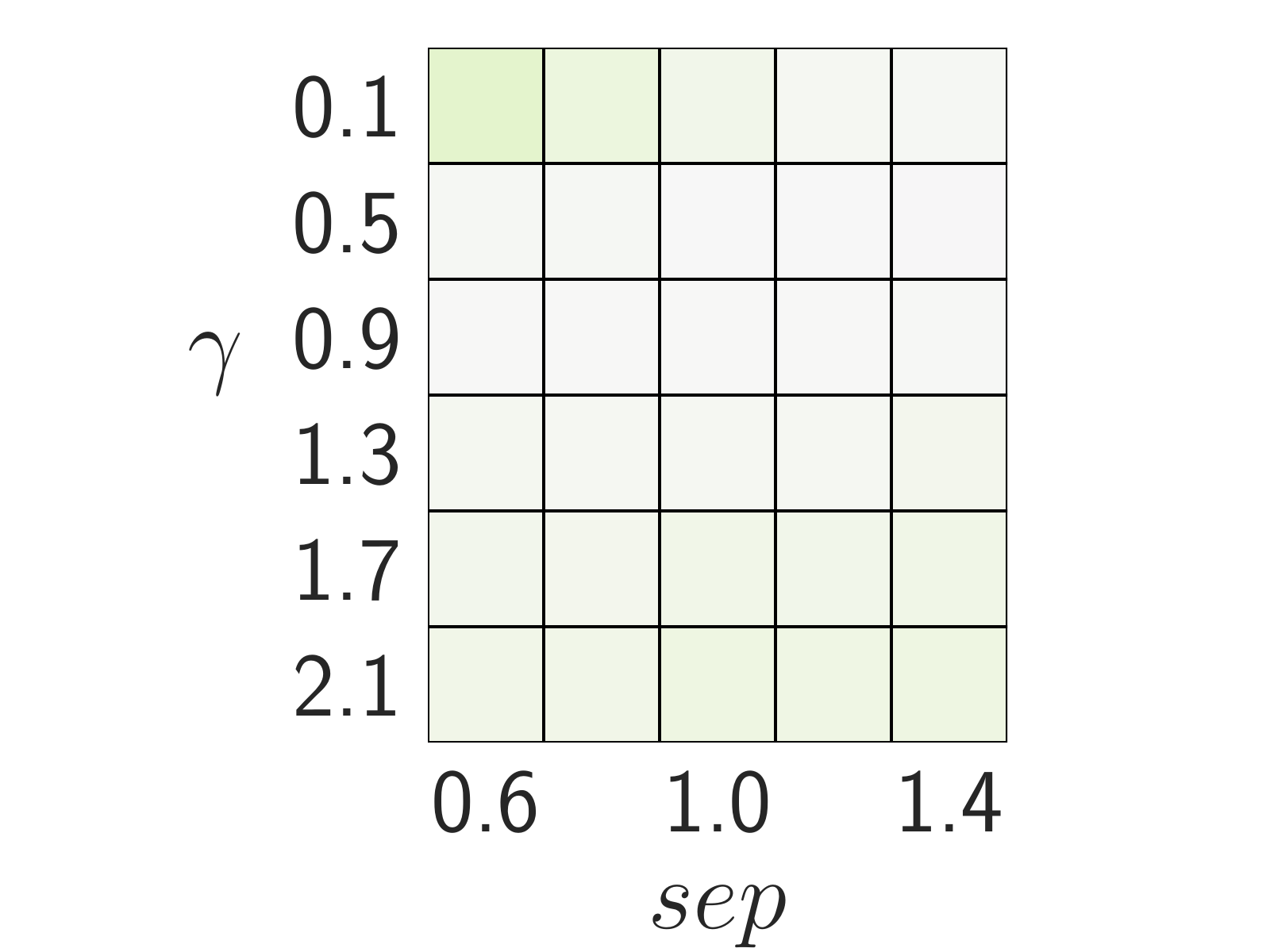} &
		\includegraphics[width=1\linewidth]{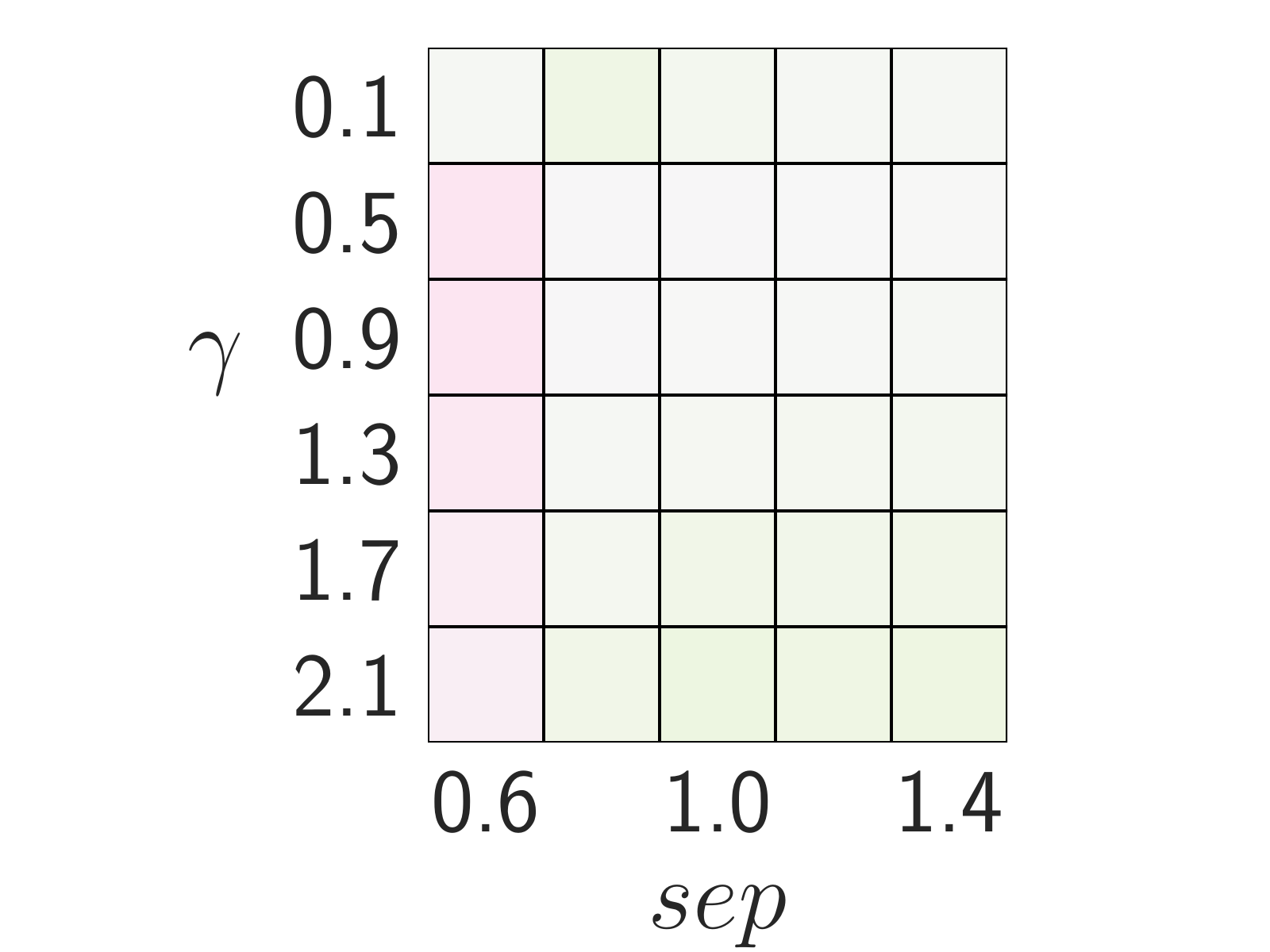} & \multirow{4}{*}[-1em]{\includegraphics[width=0.1\linewidth]{images/legend_Fm_vertical}}\\
		5 & \includegraphics[width=1\linewidth]{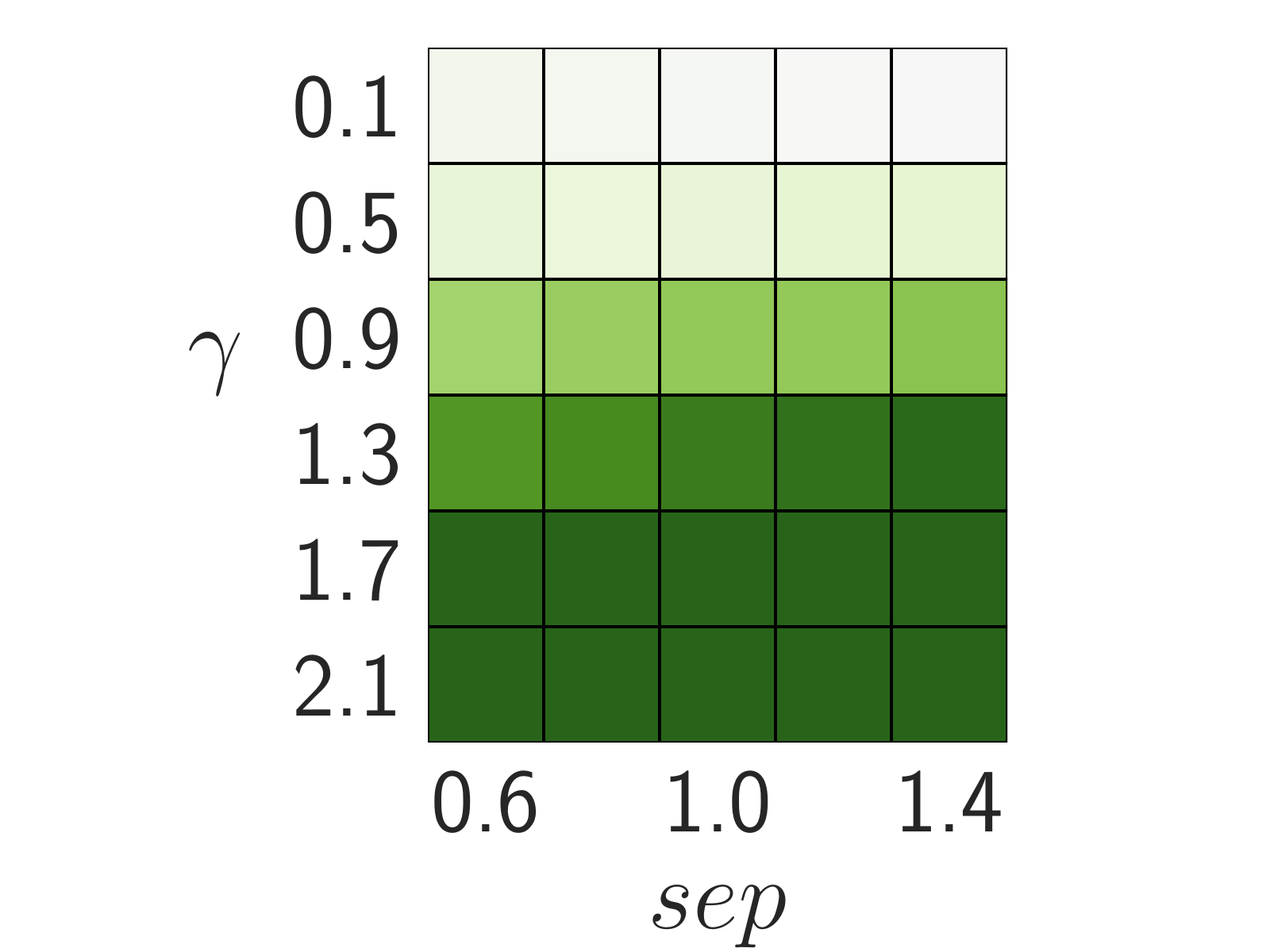} &
		\includegraphics[width=1\linewidth]{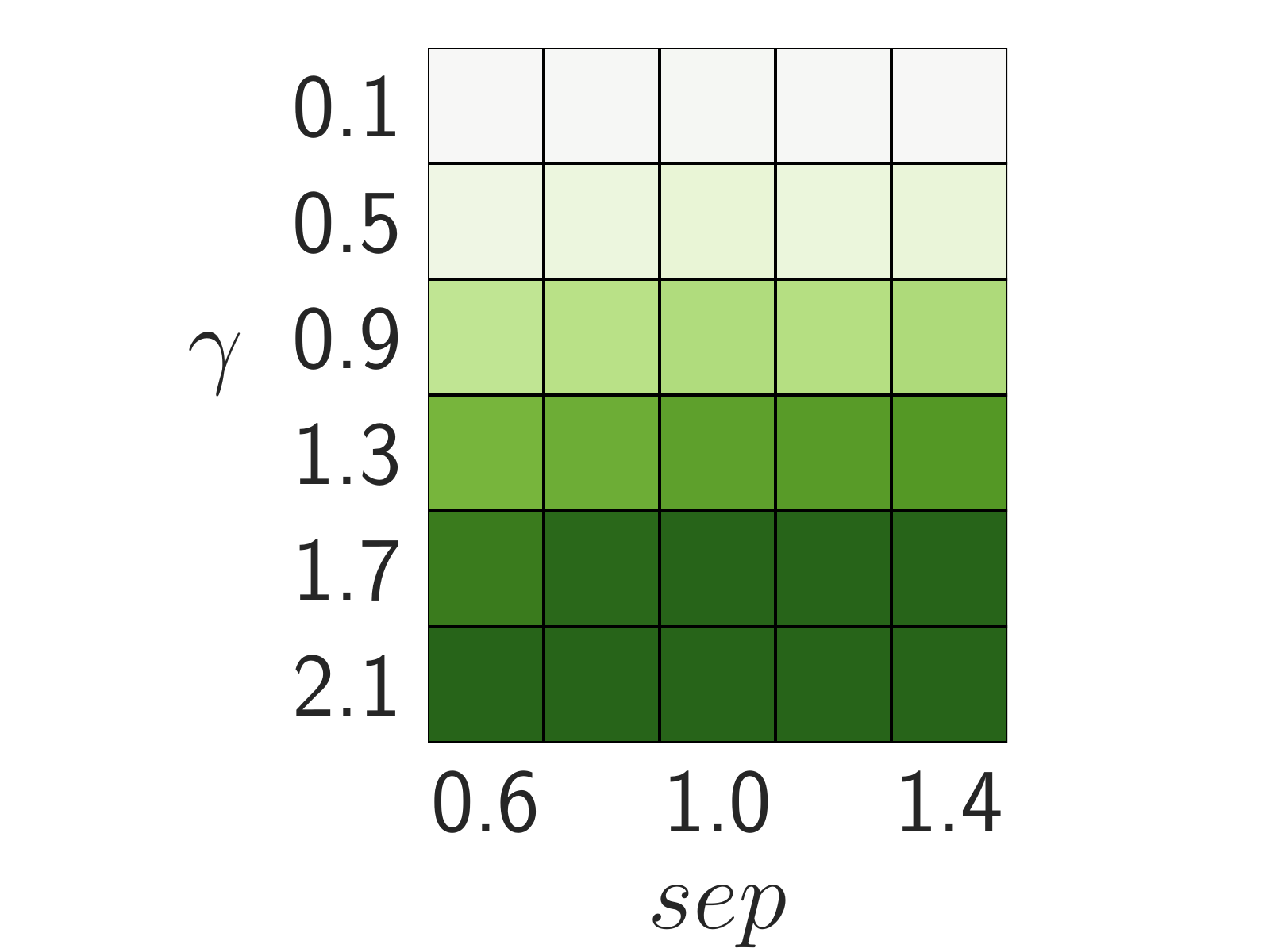} &
		\includegraphics[width=1\linewidth]{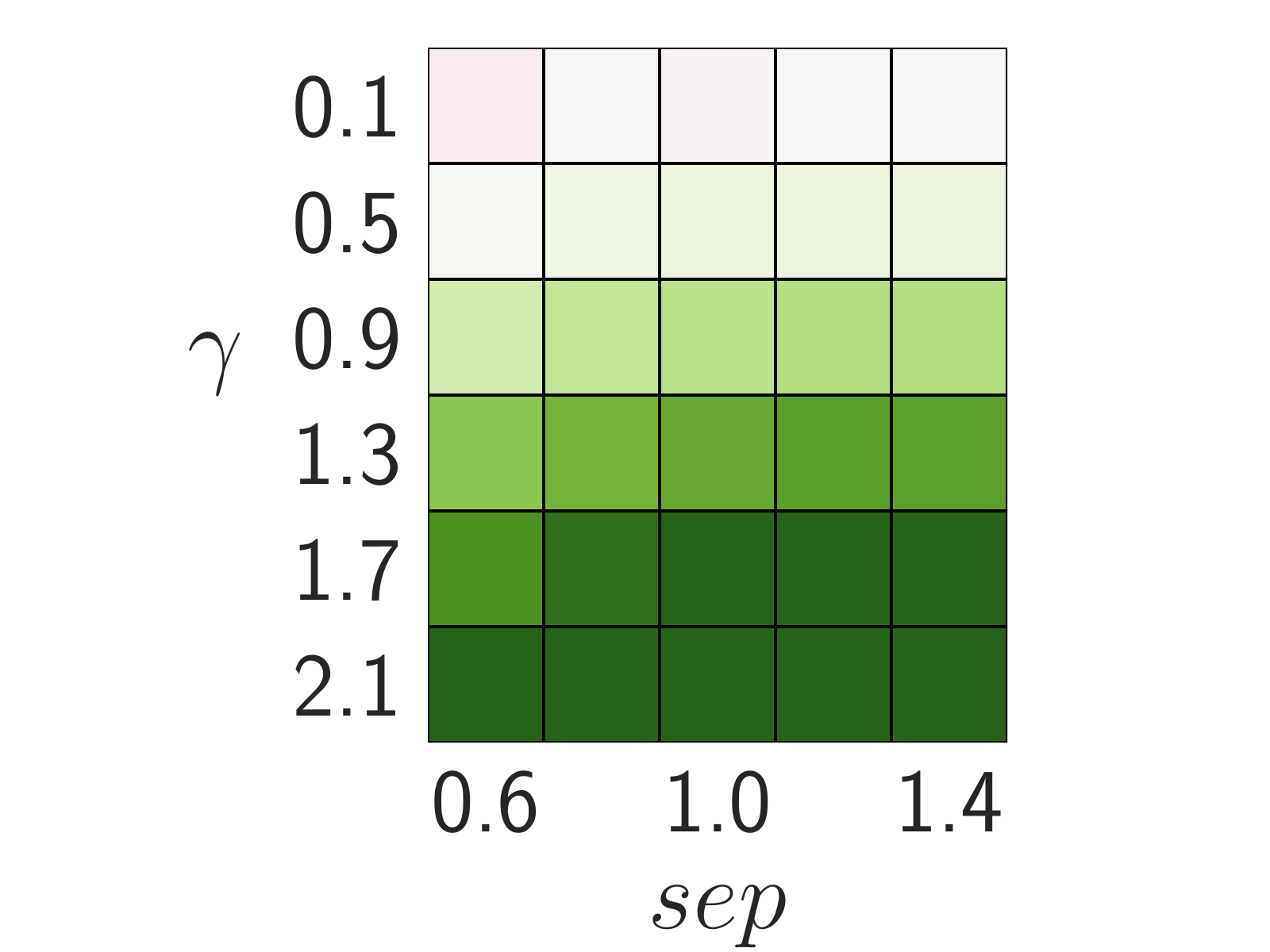} & \\
		6 & \includegraphics[width=1\linewidth]{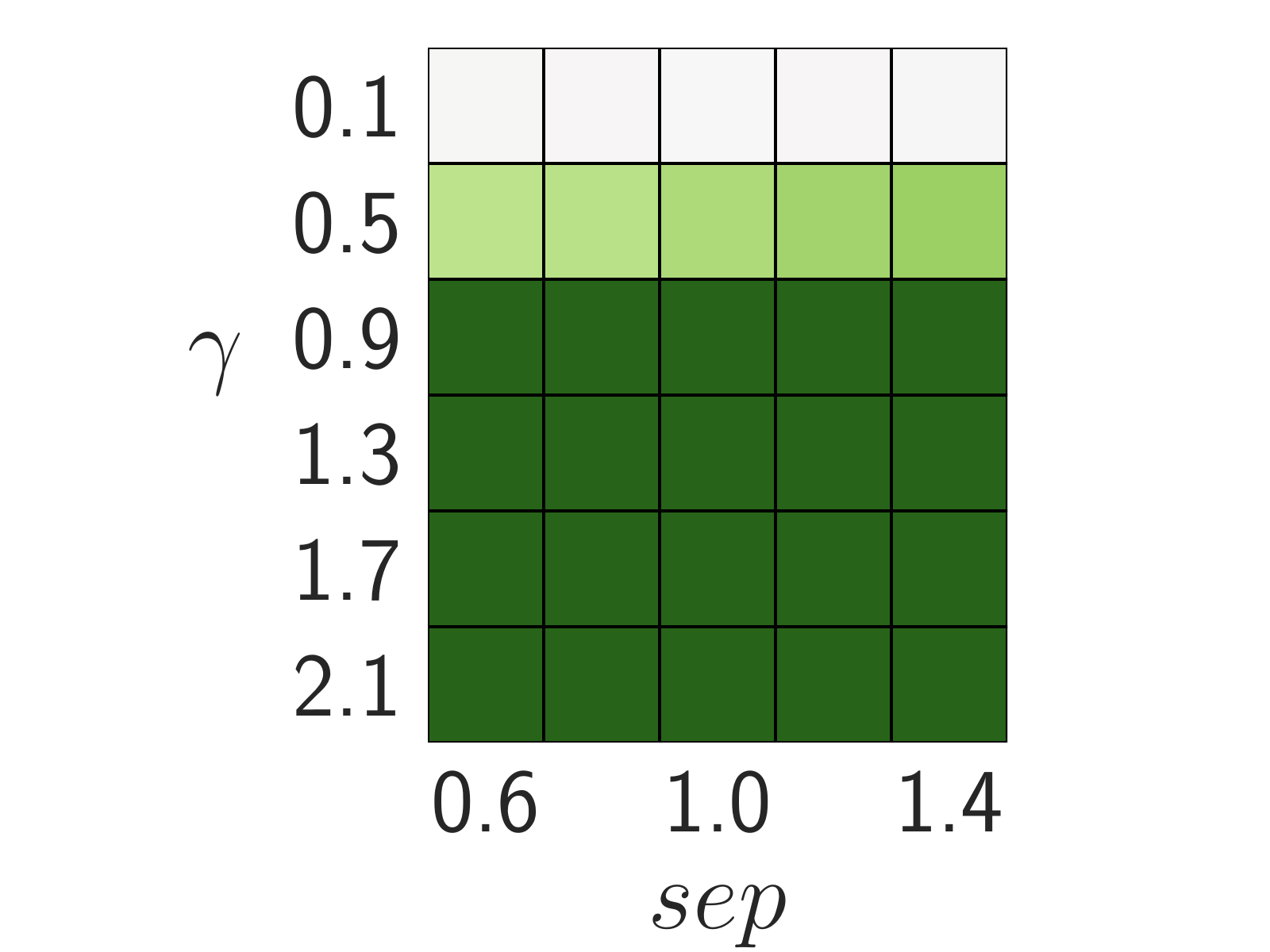} &
		\includegraphics[width=1\linewidth]{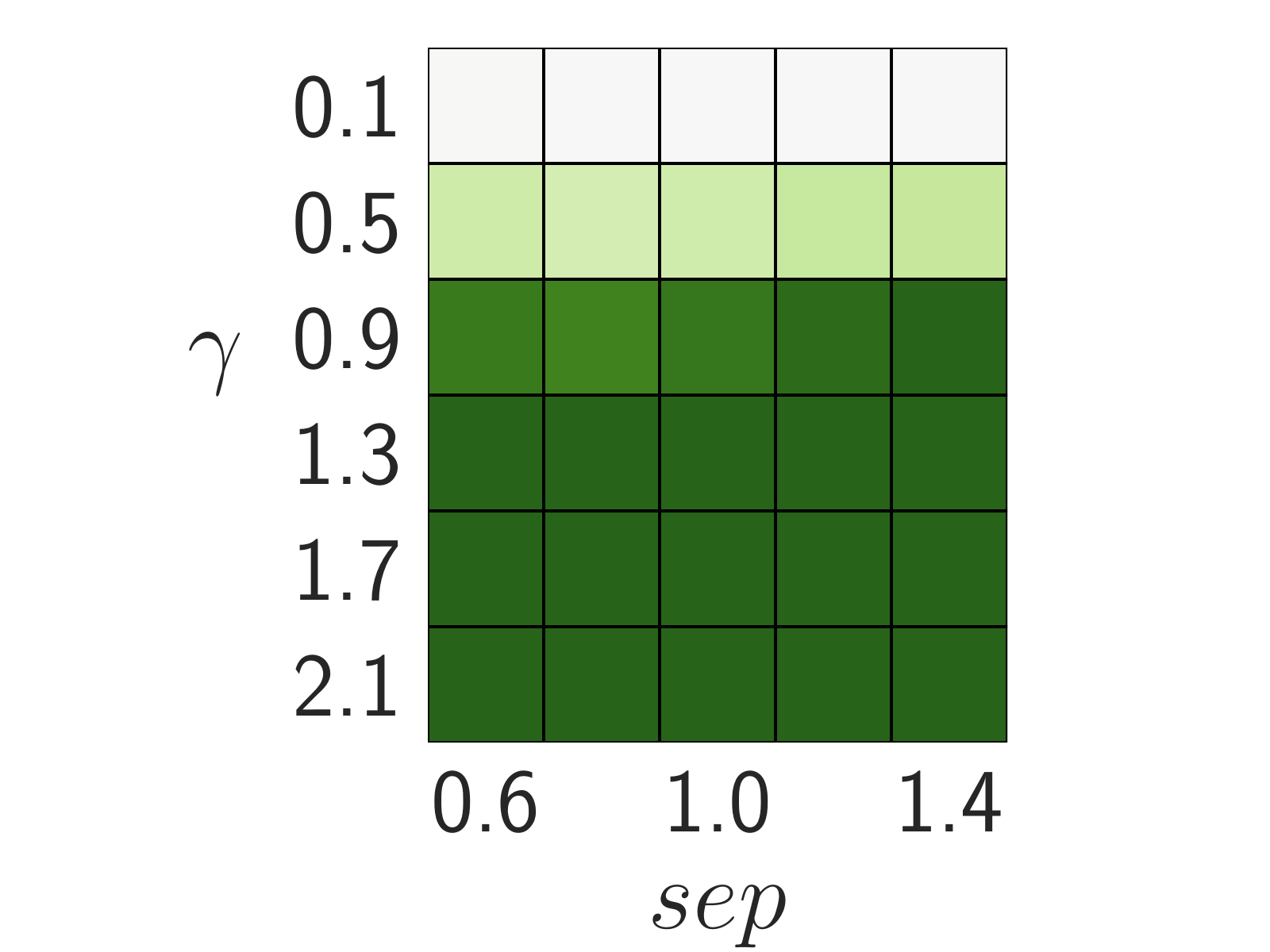} &
		\includegraphics[width=1\linewidth]{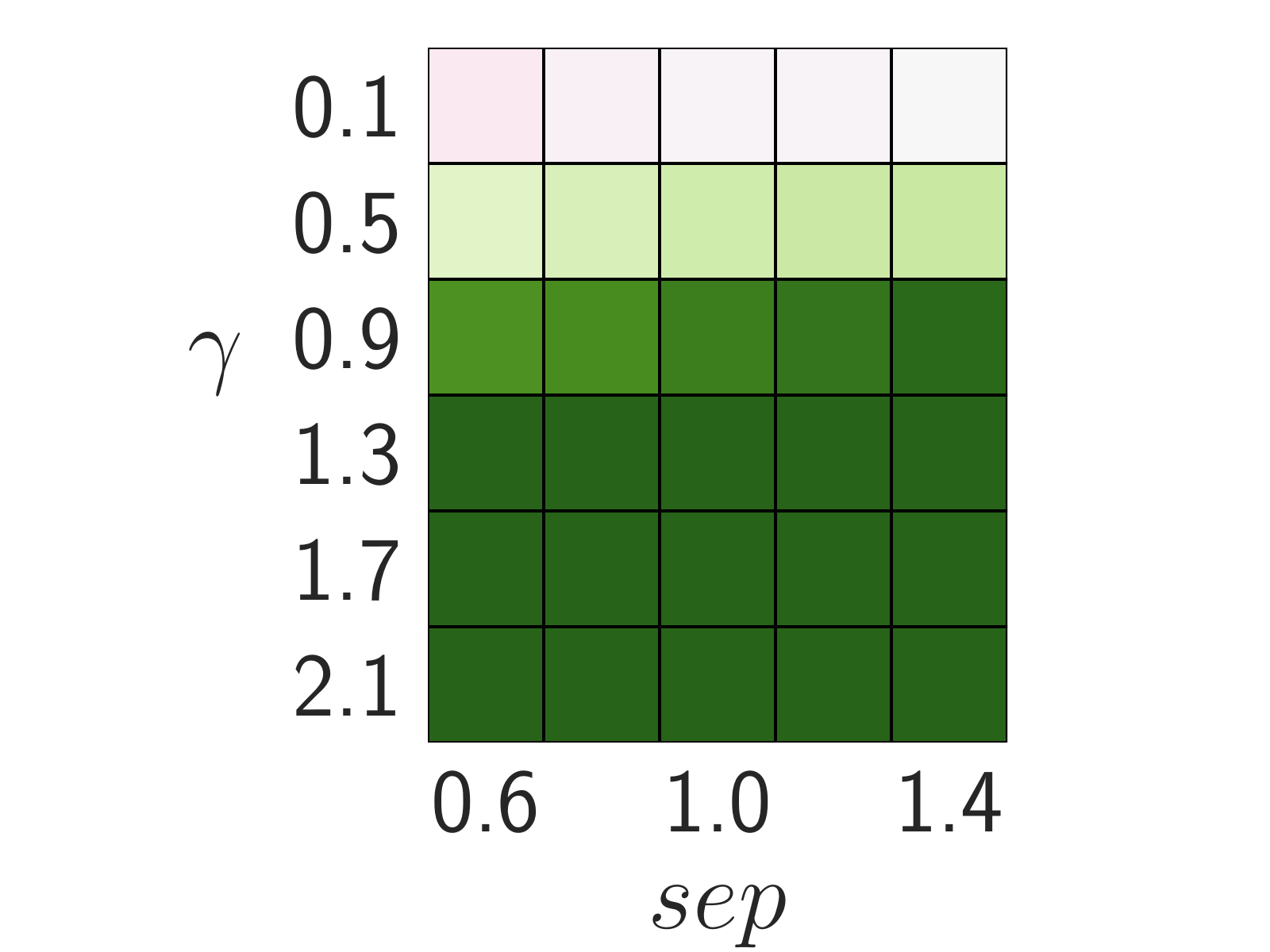} & \\
		7 & \includegraphics[width=1\linewidth]{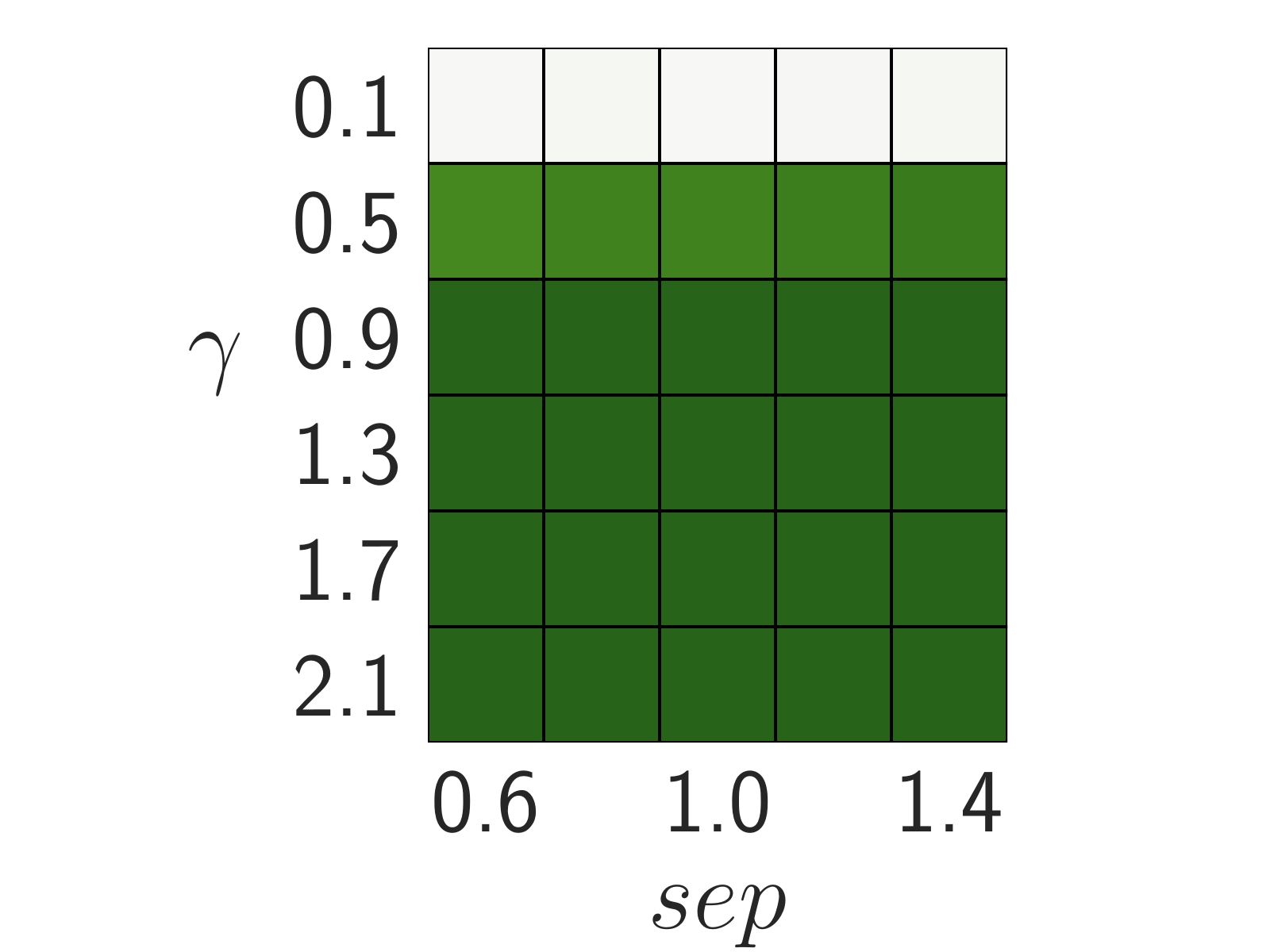} &
		\includegraphics[width=1\linewidth]{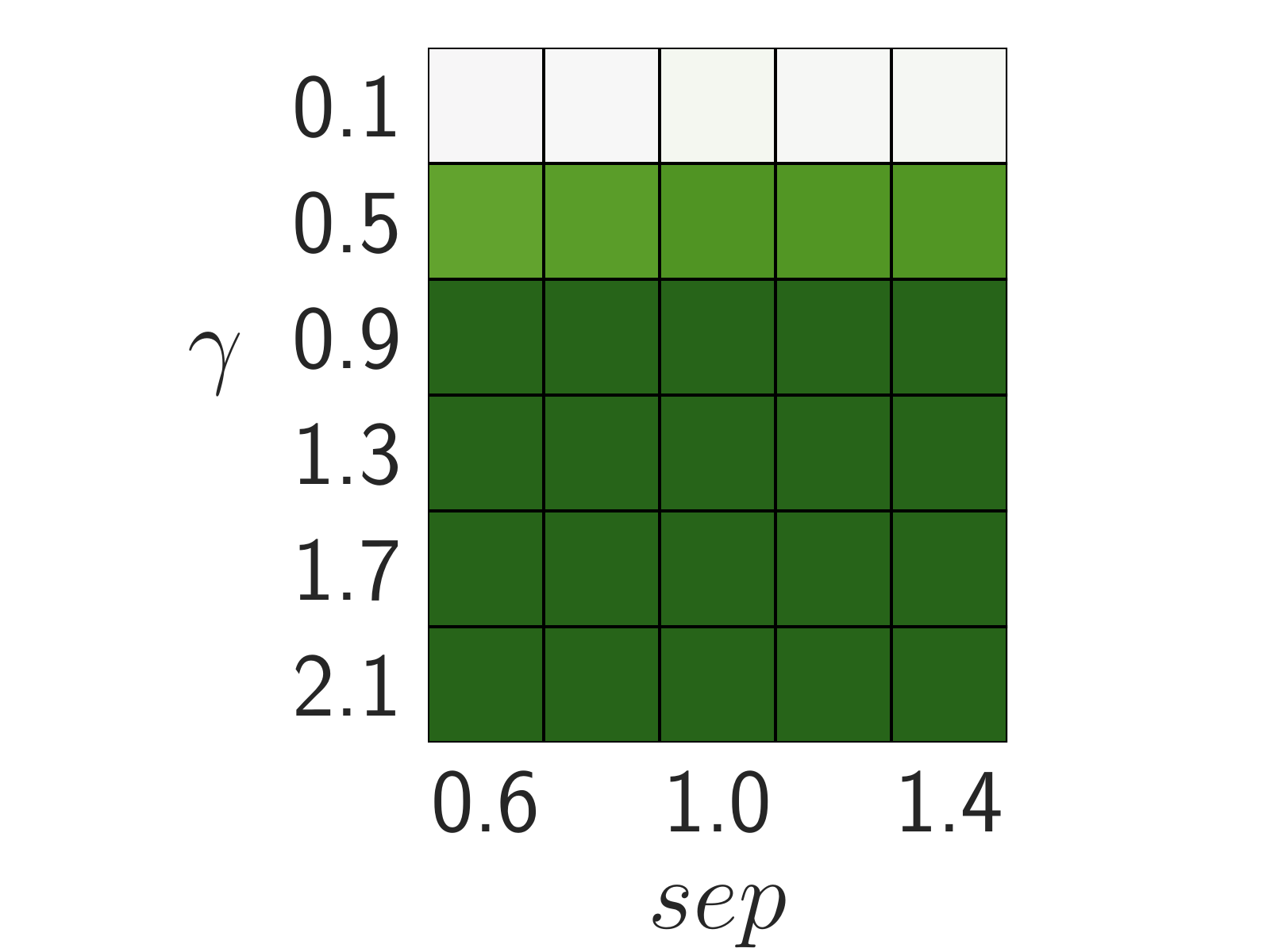} &
		\includegraphics[width=1\linewidth]{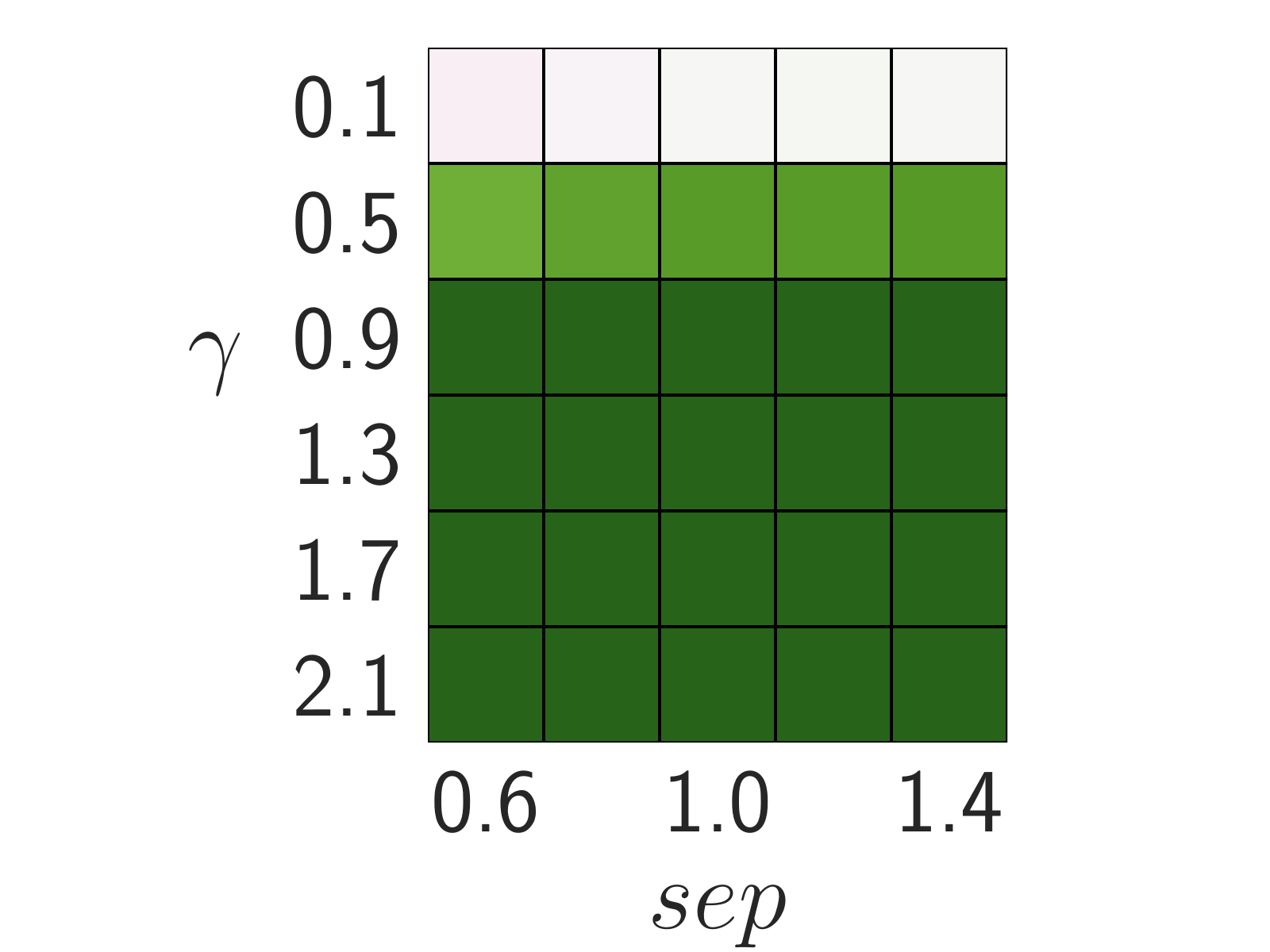} & \\
	\end{tabular}
\\
\caption{Performance improvement relation of \textit{OKSVM} against \textit{SVM}. Heatmaps are organized in the figure according to their tuned configuration where each row indicated the number of dimensions $dim$ of the datasets while columns indicated the parameter $C$. Each heatmap exhibits the average performance of the tuned configuration where each row of the heatmap represents the parameter $gamma$ and columns represent the separation $sep$ of the clusters of the datasets. Higher is better.}
\label{fig:heatmap_Fm}
\end{figure}

Next, the performance improvement heatmaps of \textit{OKSVM} against \textit{SVM} for dimensions 3, 5, 6, and 7 are presented in \figurename\,\ref{fig:heatmap_Fm}. $F_{diff}$ measure represents the scaled error between both methods. That is, a value $F_{diff}=10$ indicates that \textit{OKSVM} yielded an $F1$-score 0.1 higher than \textit{SVM}, which is similar to a 10\% of improvement. And this is what happens with the highest dimensions (5, 6 and 7), where the margin is quite better. These outcomes are in concordance with \figurename\,\ref{fig:heatmap_Fm_comparison2}.

\newpage

\begin{figure}
\centering
	\begin{tabular}{D|CCCc}
	    $dim$ & $C=0.5$ & $C=1.0$ & $C=1.5$ & \\
	    \midrule
		3 & \includegraphics[width=1\linewidth]{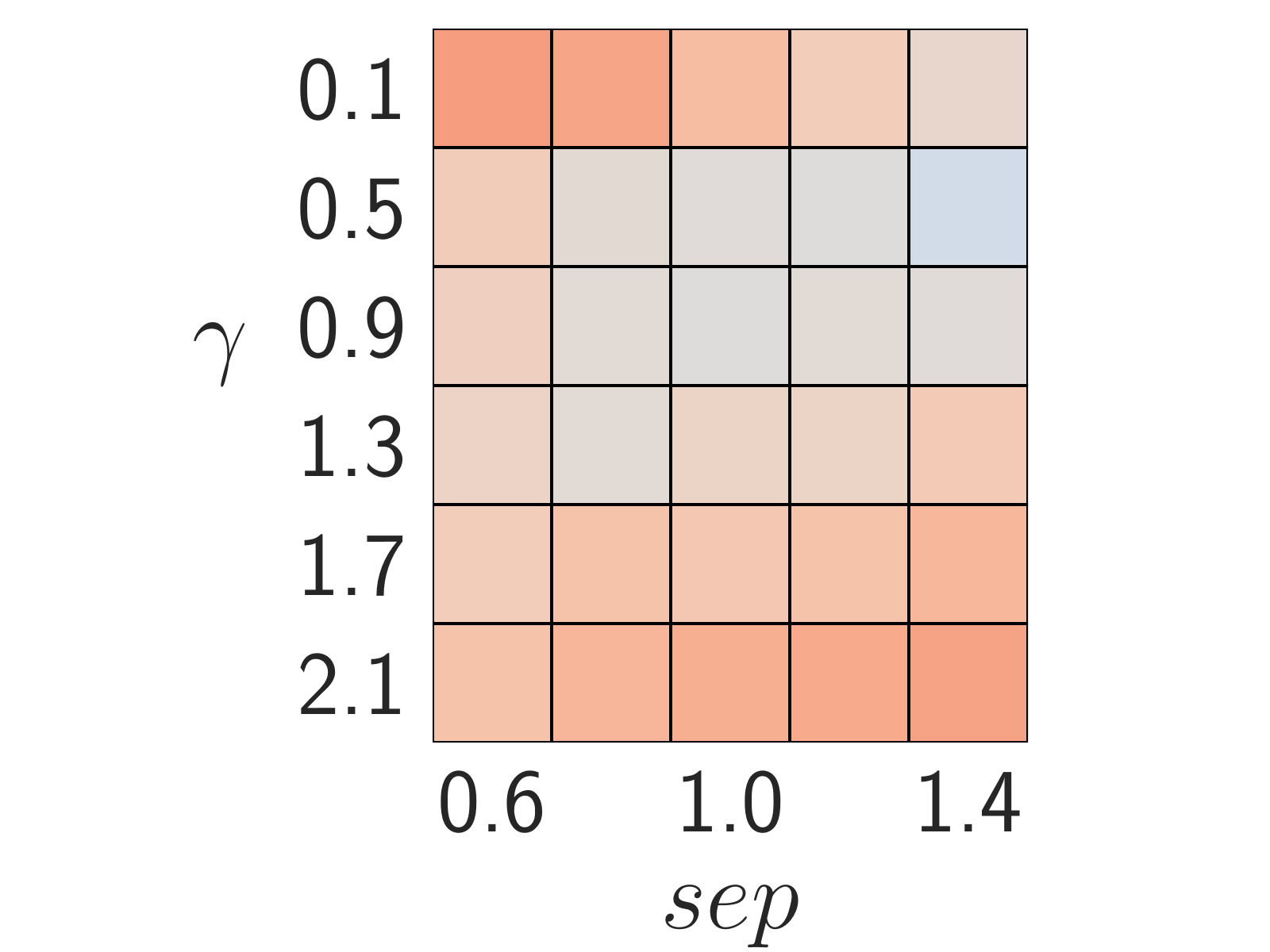} &
		\includegraphics[width=1\linewidth]{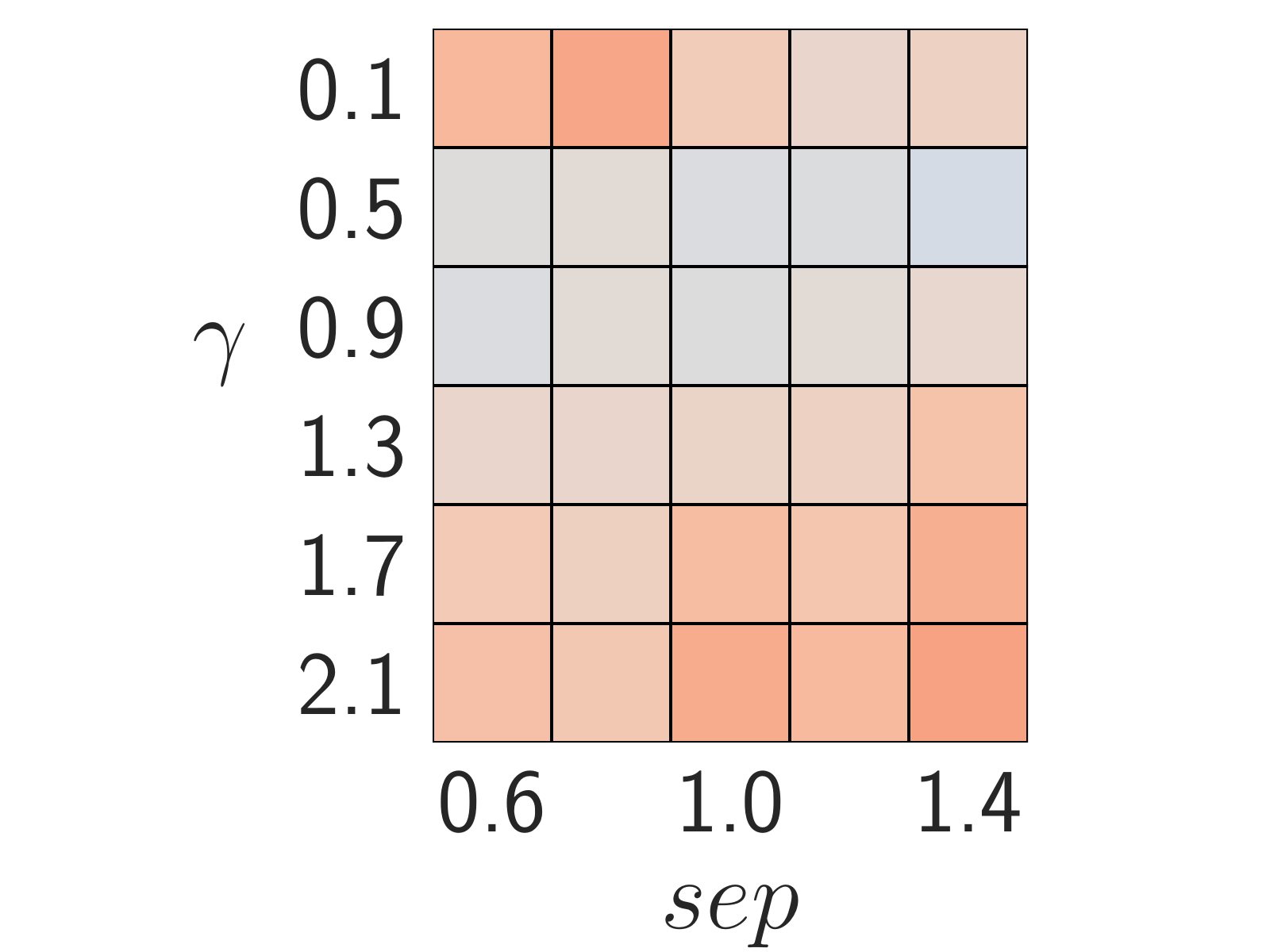} &
		\includegraphics[width=1\linewidth]{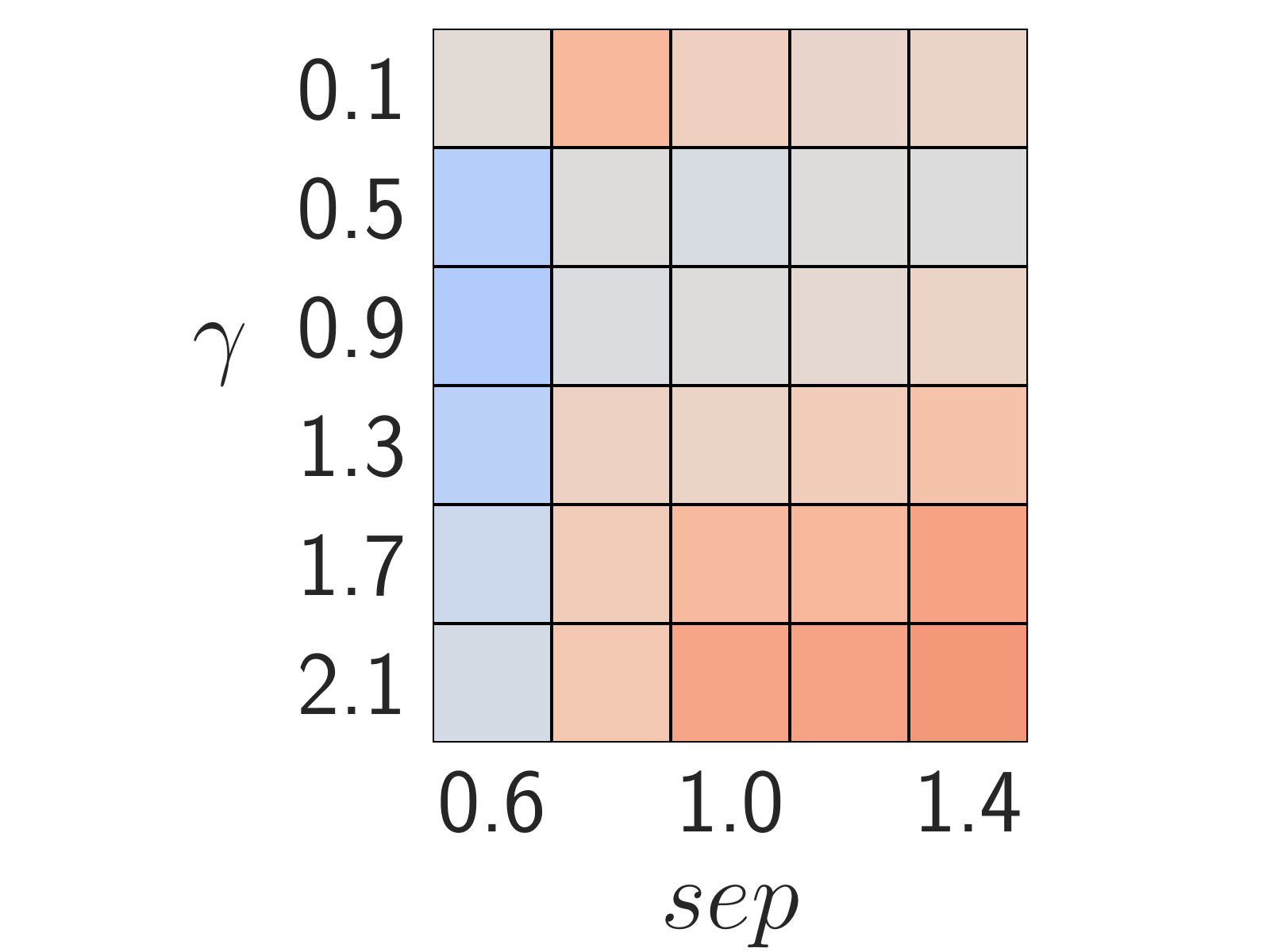} & \multirow{4}{*}{\includegraphics[width=0.1\linewidth]{images/legend_winslosses_vertical}}\\
		5 & \includegraphics[width=1\linewidth]{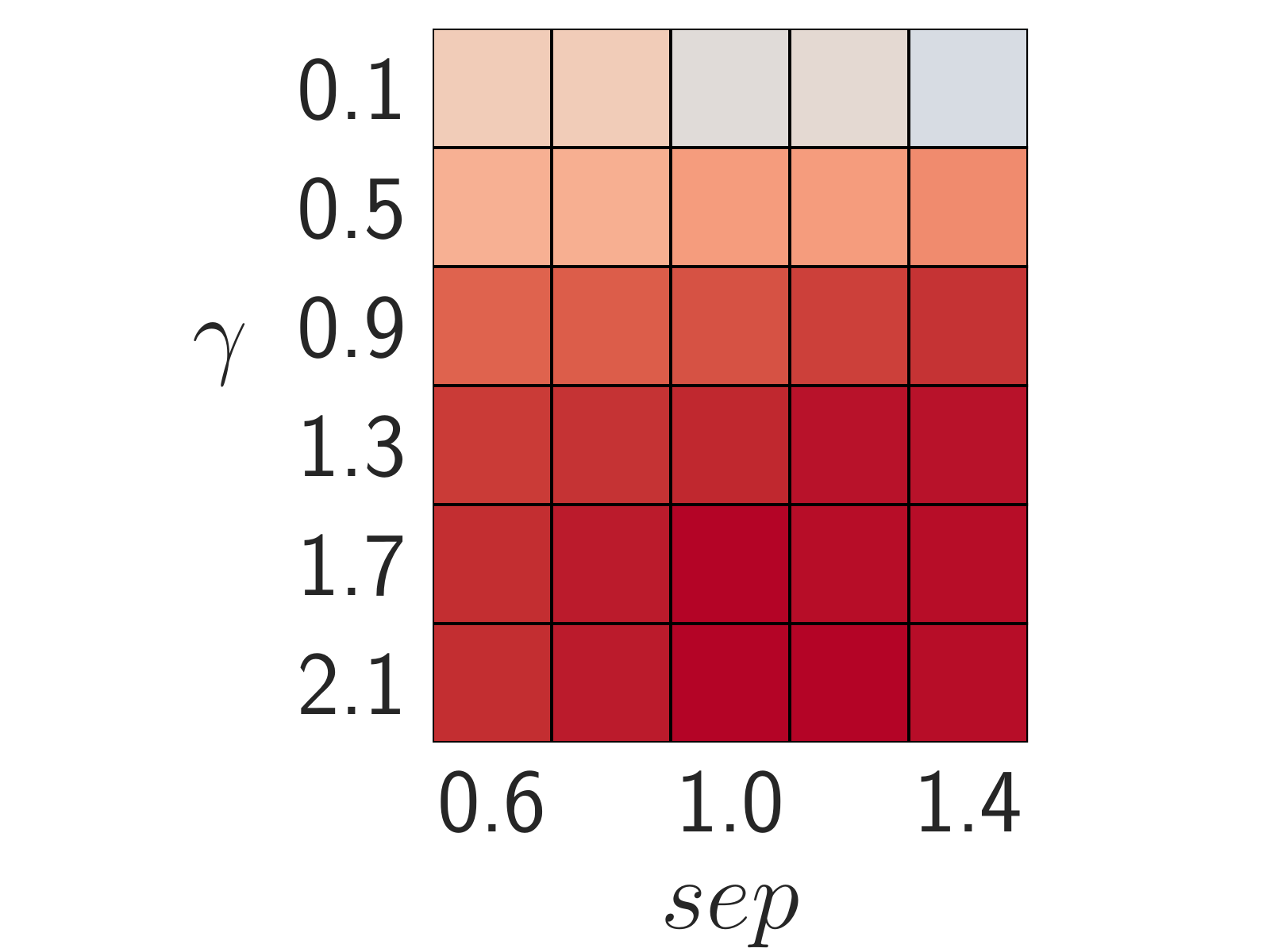} &
		\includegraphics[width=1\linewidth]{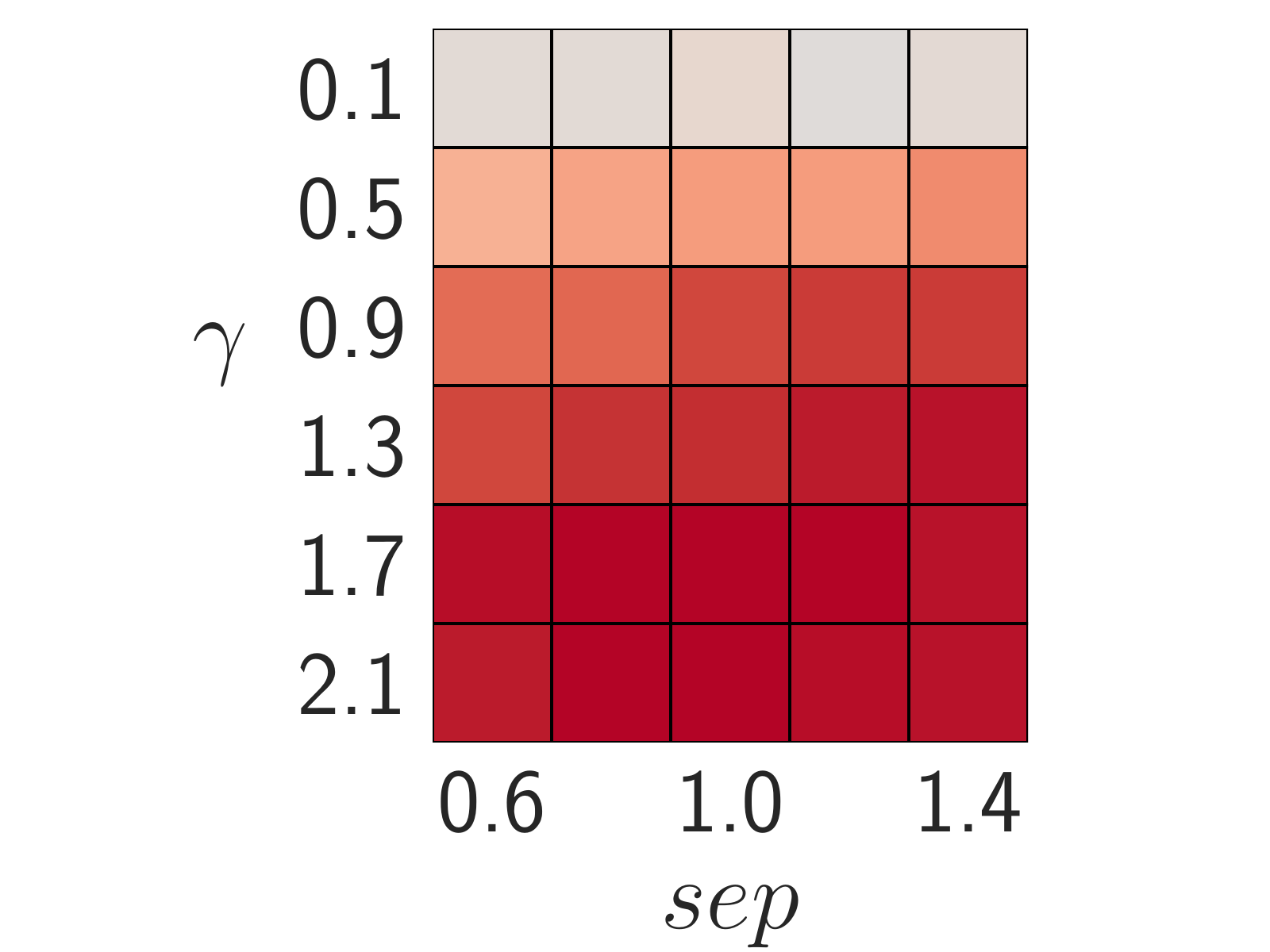} &
		\includegraphics[width=1\linewidth]{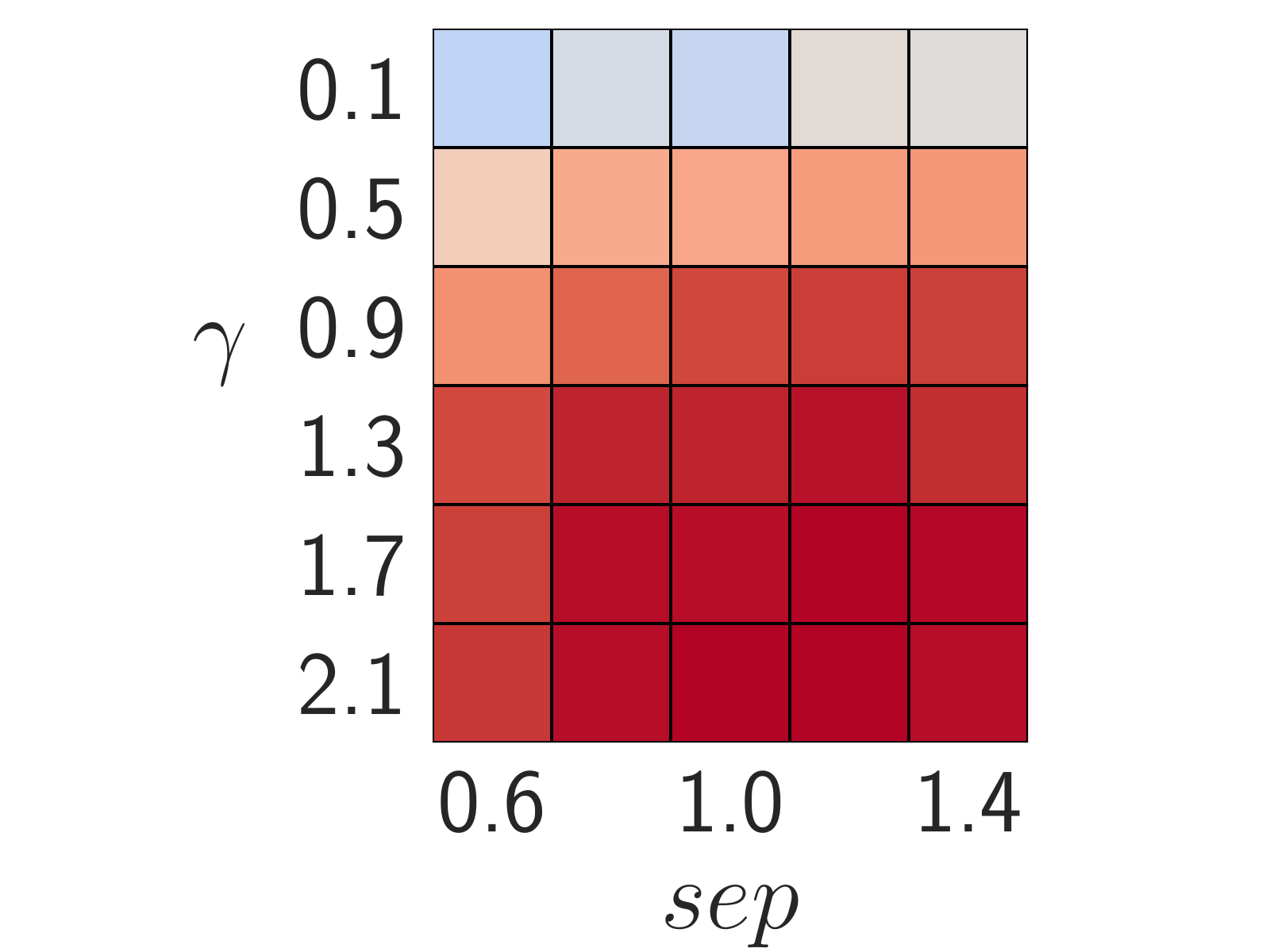} & \\
		6 & \includegraphics[width=1\linewidth]{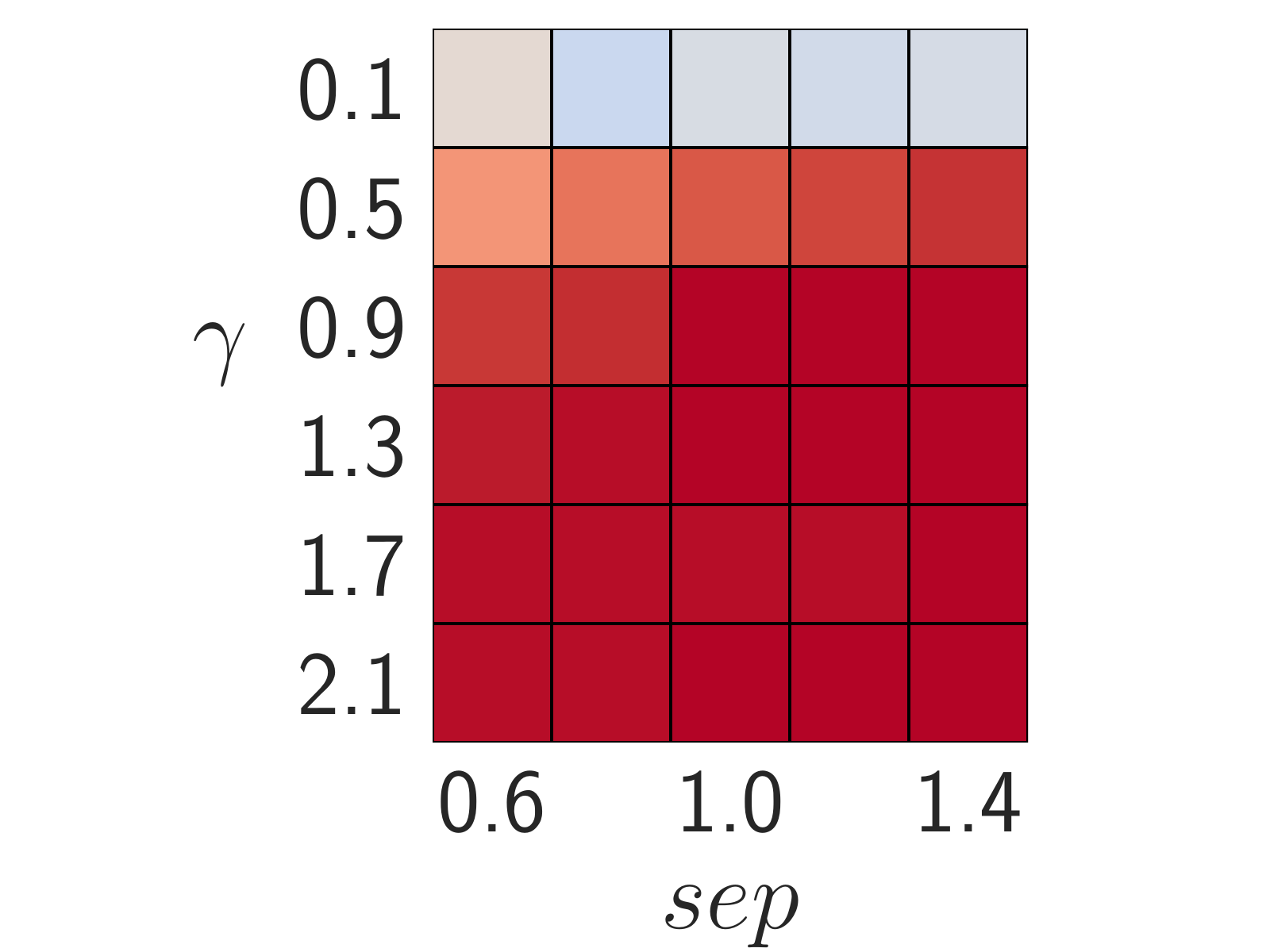} &
		\includegraphics[width=1\linewidth]{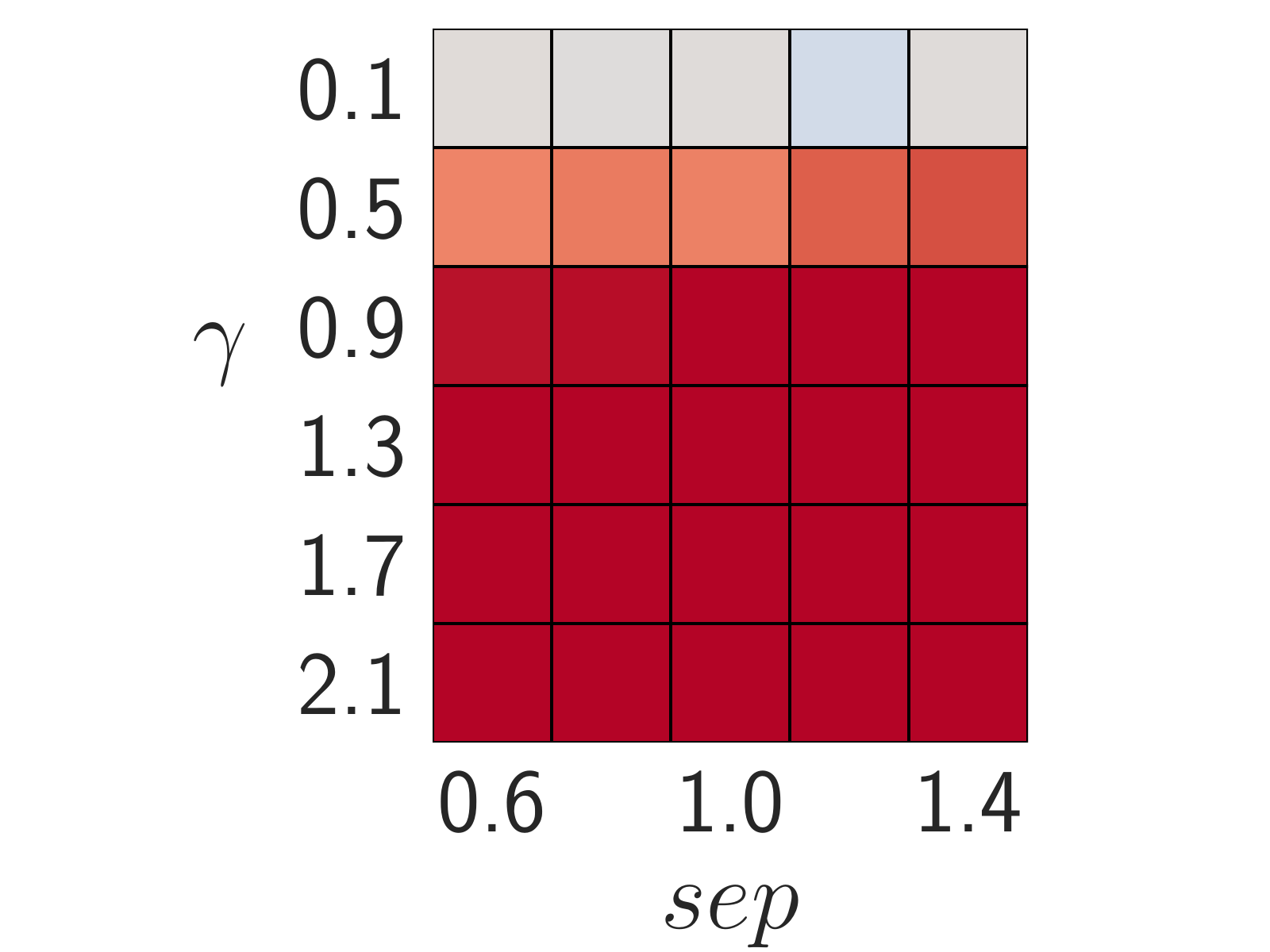} &
		\includegraphics[width=1\linewidth]{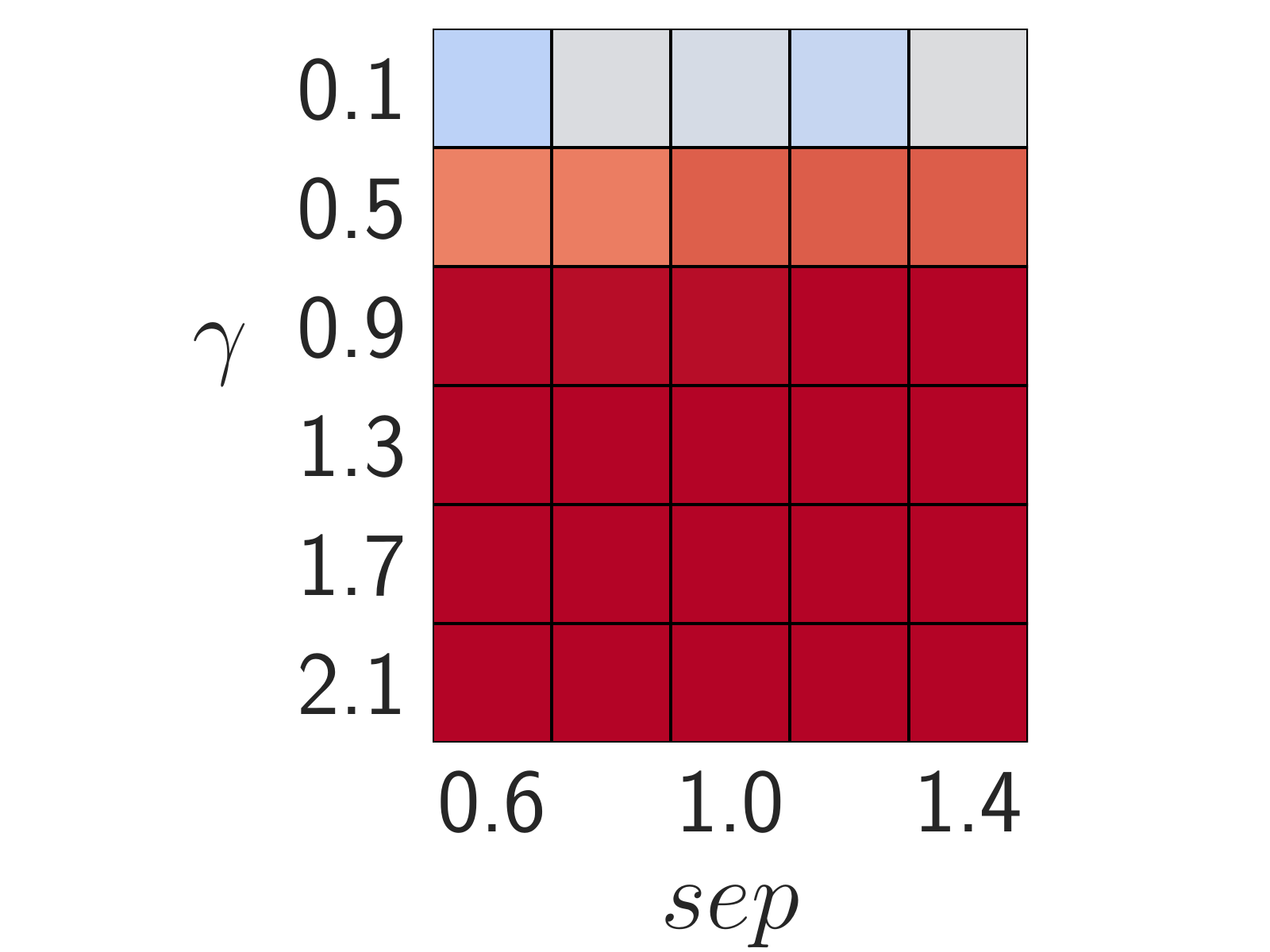} & \\
		7 & \includegraphics[width=1\linewidth]{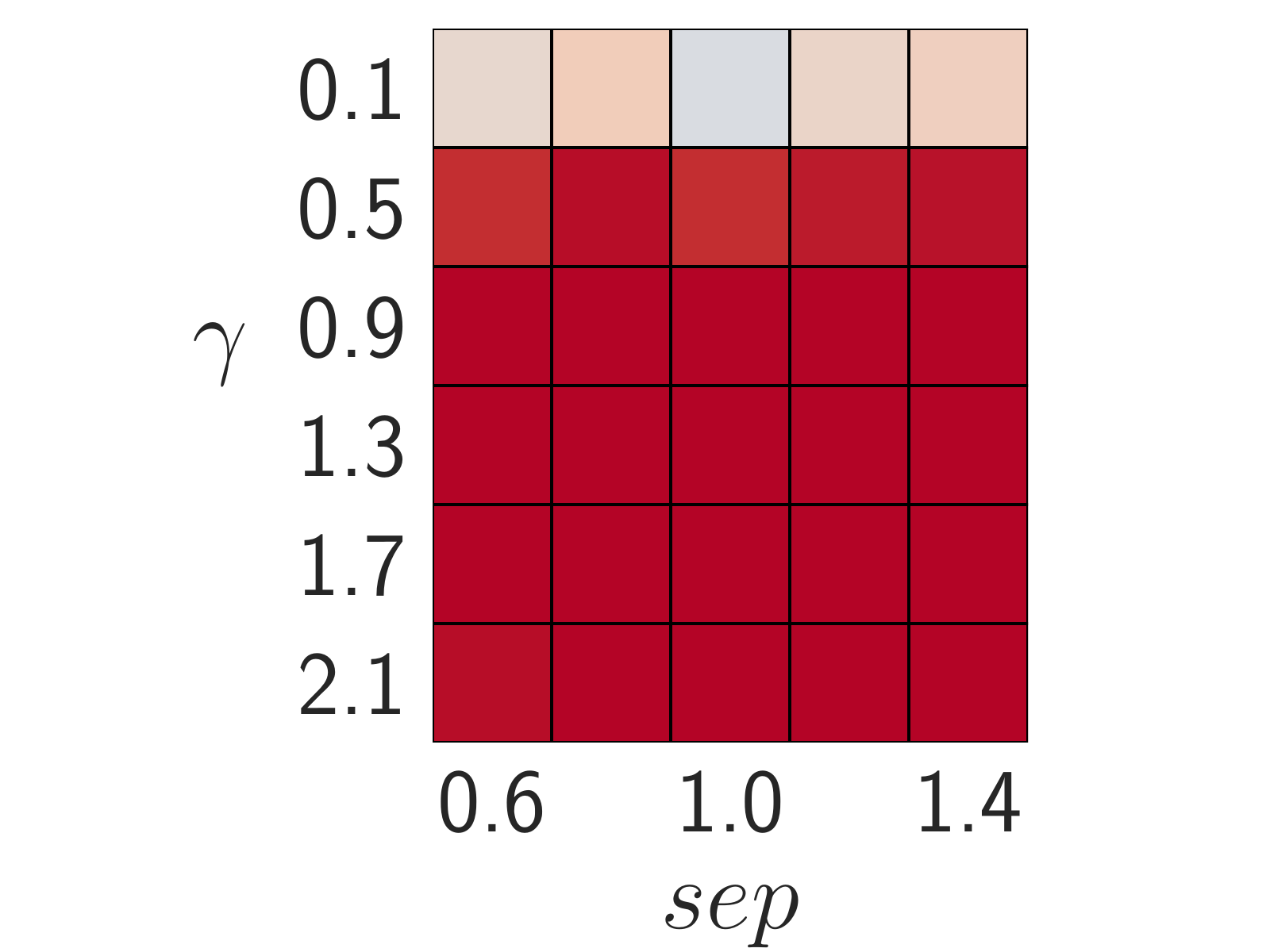} &
		\includegraphics[width=1\linewidth]{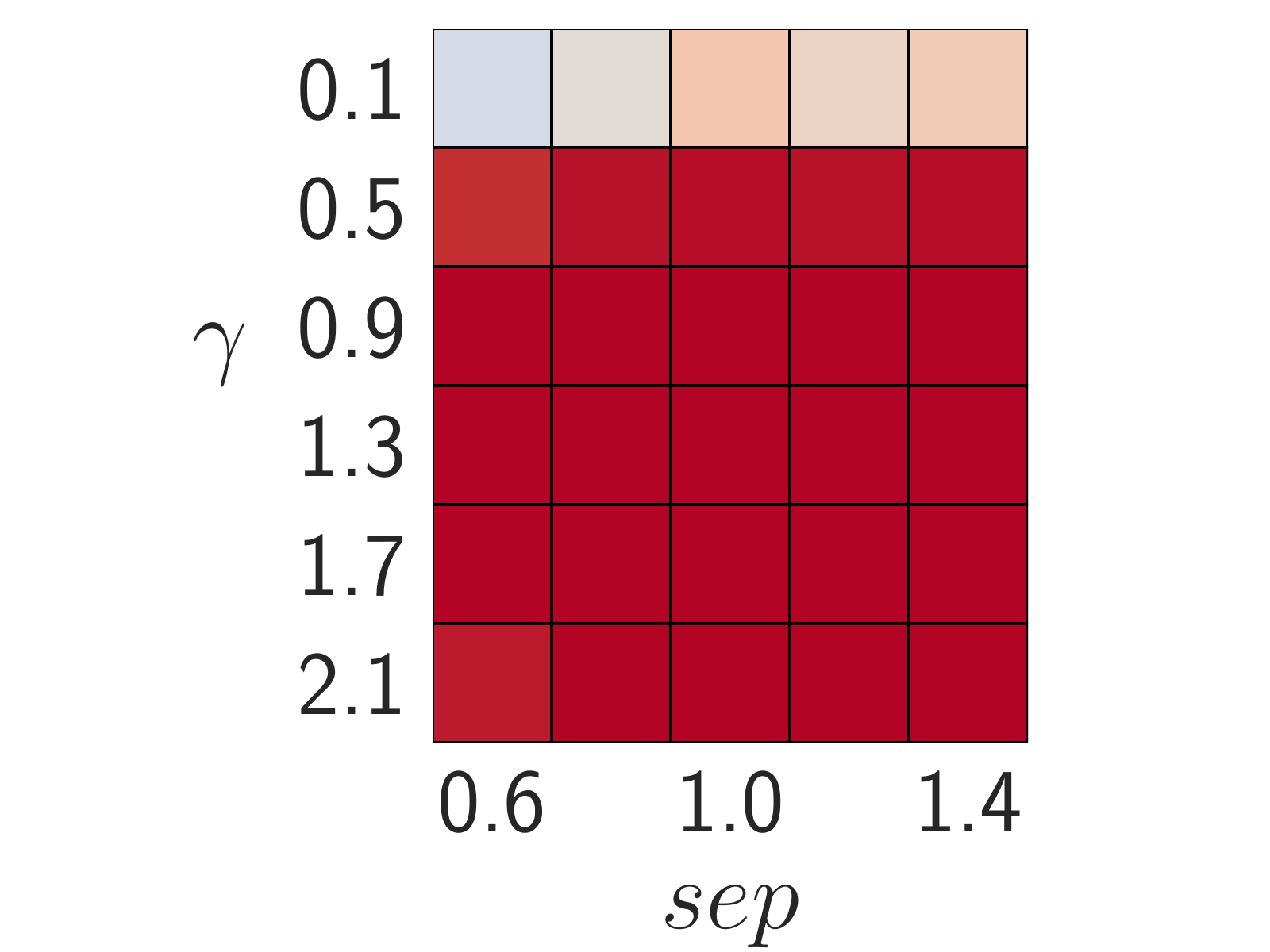} &
		\includegraphics[width=1\linewidth]{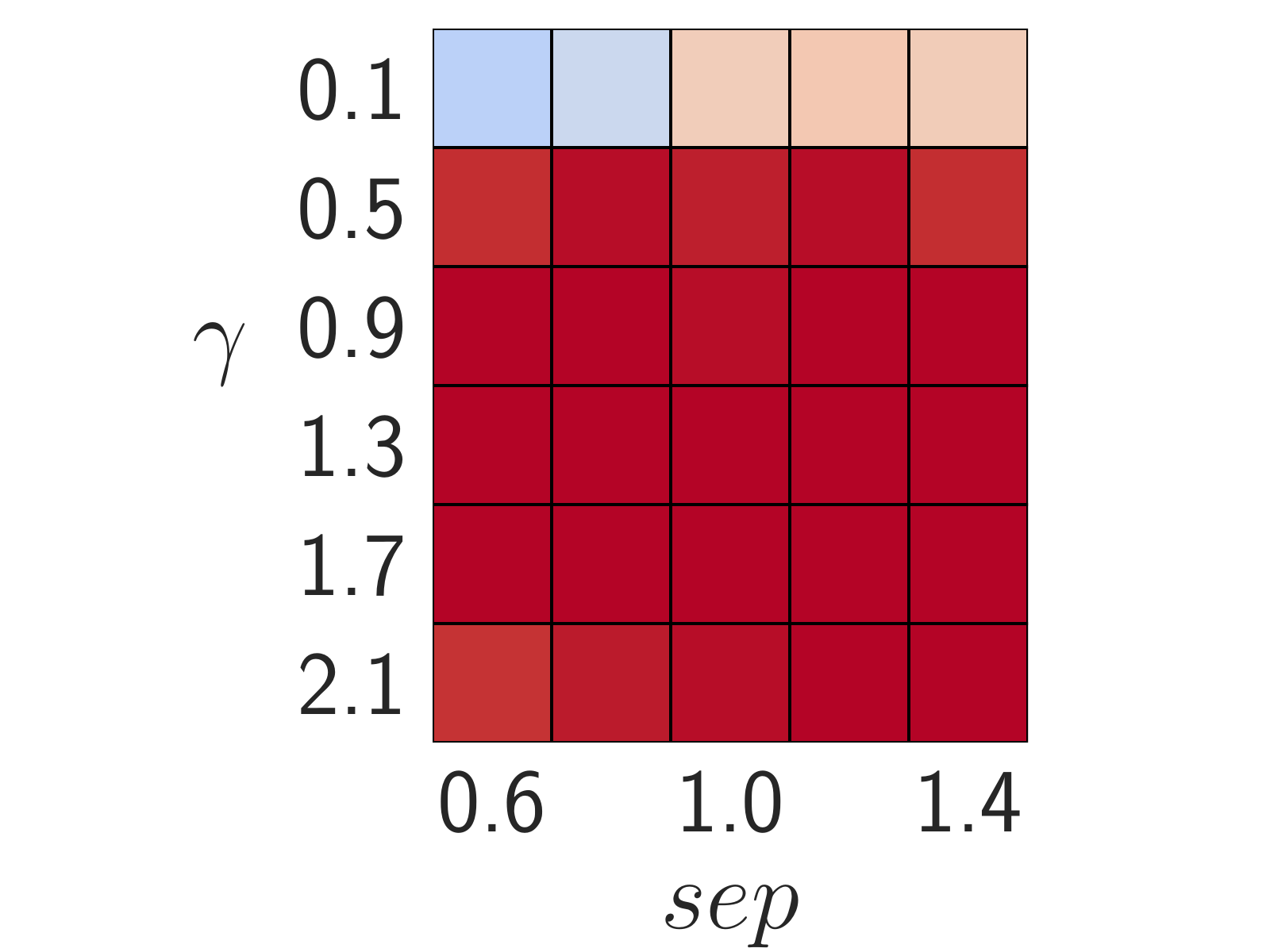} & 
	\end{tabular}
\caption{Wins-Losses ratio of \textit{OKSVM} against \textit{SVM}. Heatmaps are organized in the figure according to their tuned configuration where each row indicated the number of dimensions $dim$ of the datasets while columns indicated the parameter $C$. Each heatmap exhibits the average performance of the tuned configuration where each row of the heatmap represents the parameter $gamma$ and columns represent the separation $sep$ of the clusters of the datasets.}
\label{fig:heatmap_Wins_Losses}
\end{figure}

The wins-losses ratio comparison is reported in \figurename\,\ref{fig:heatmap_Wins_Losses}. Again, the outcomes are very similar to the ones reported in the paper. The lower the dimension is the more equality in the fitting. Thus, it can be seen that our proposal fails for some configurations with $dim=3$, but the general behavior goes in favor of \textit{OKSVM}. That is due to the simplicity of the binary classification problem of in the 3D space, even more, when more relaxation (higher $C$) is permitted. Regarding the rest of the cases, the higher is $\gamma$, the larger are the differences.

\newpage

\subsubsection*{Optimized hyperparameters comparison}

The second type of experiment consists of a previous tuning step of the hyperparameters $\gamma$ and $C$ using a validation set. Therefore, in the case of the traditional SVM, this process searches the optimal values for both parameters, and only $C$ is tuned for \textit{OKSVM} method, in order to demonstrate its no dependency on the $\gamma$ parameter.

\begin{figure}[t]
\centering
\includegraphics[width=0.48\linewidth]{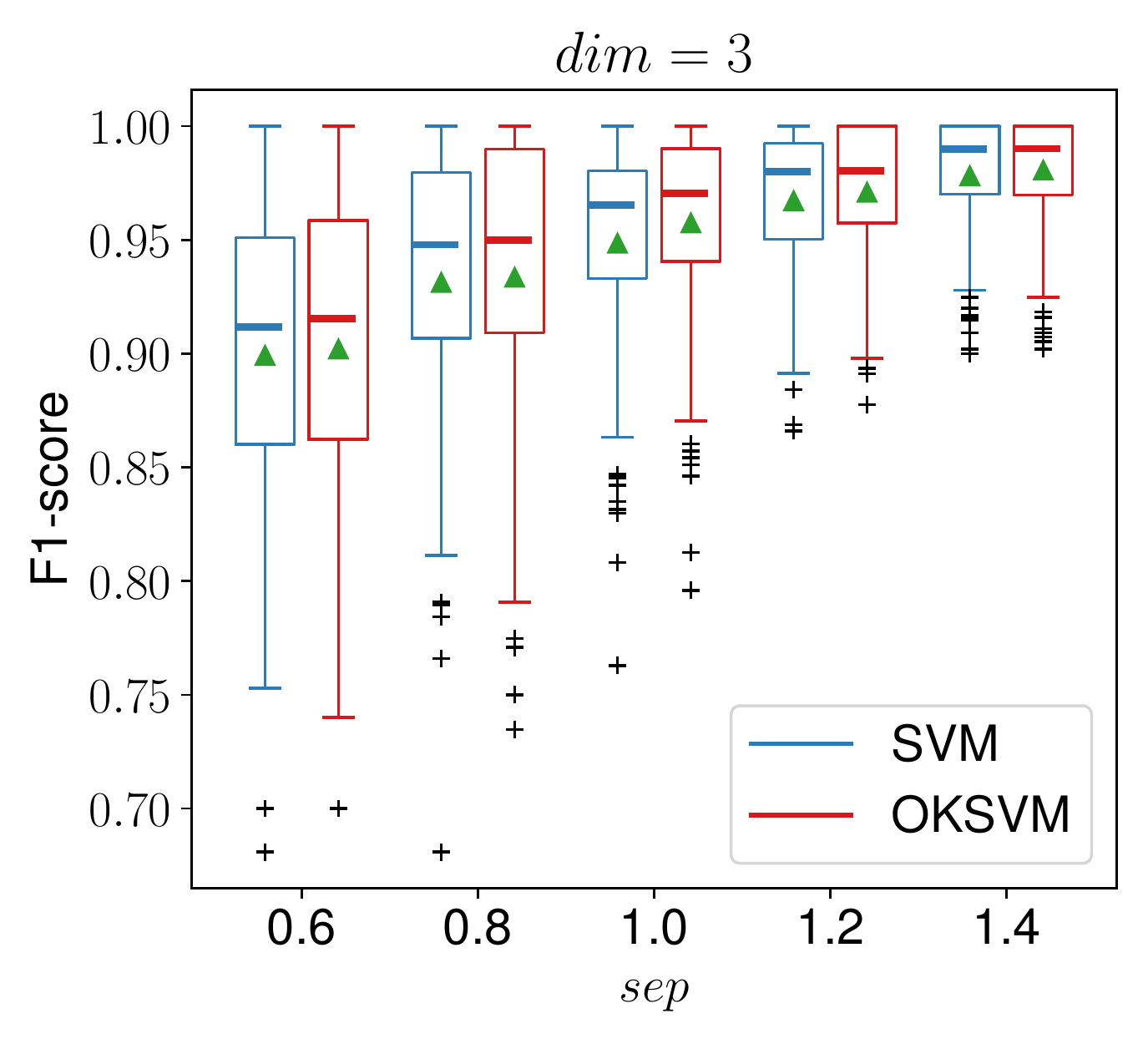}
\includegraphics[width=0.48\linewidth]{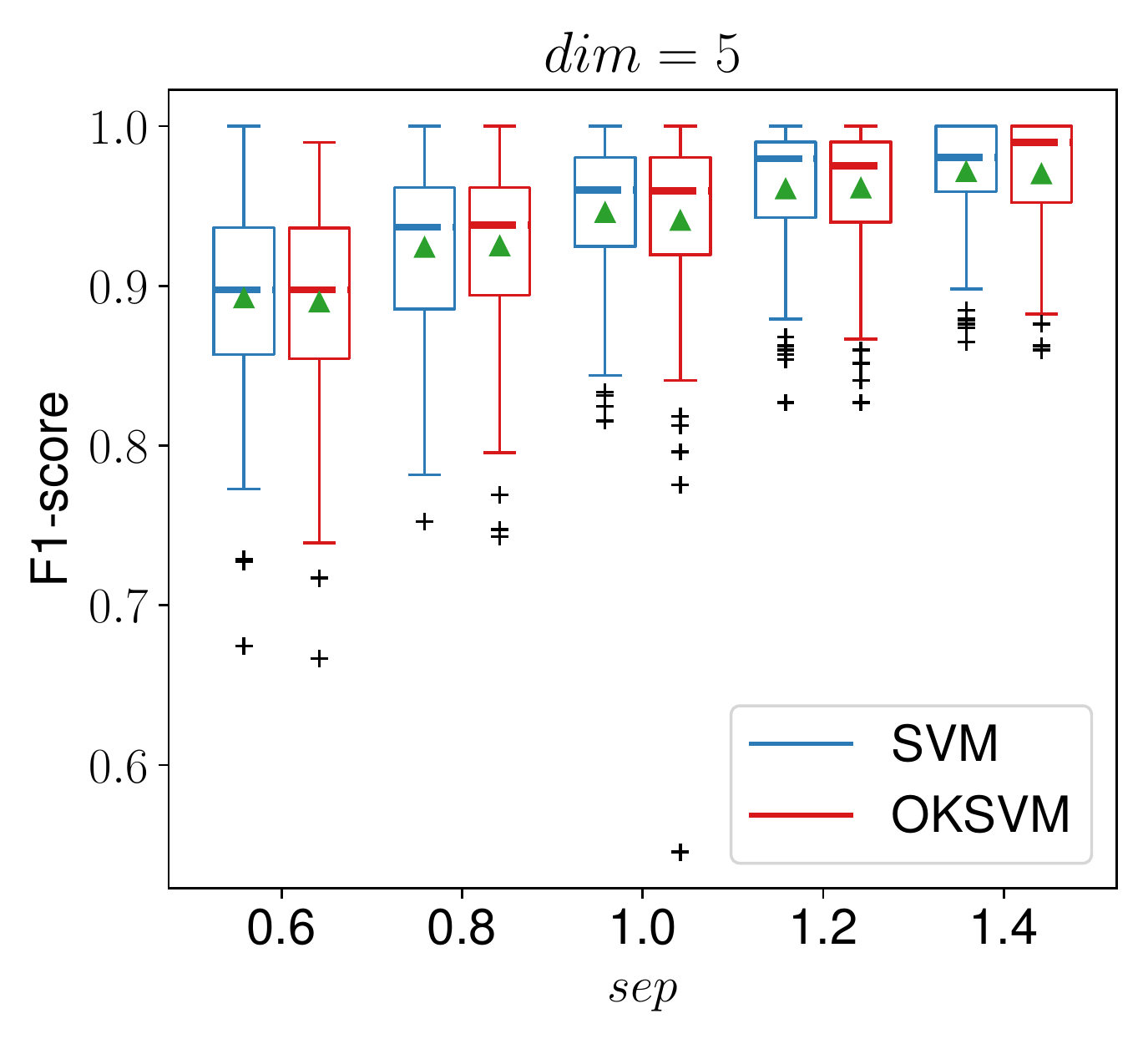} \\
\includegraphics[width=0.48\linewidth]{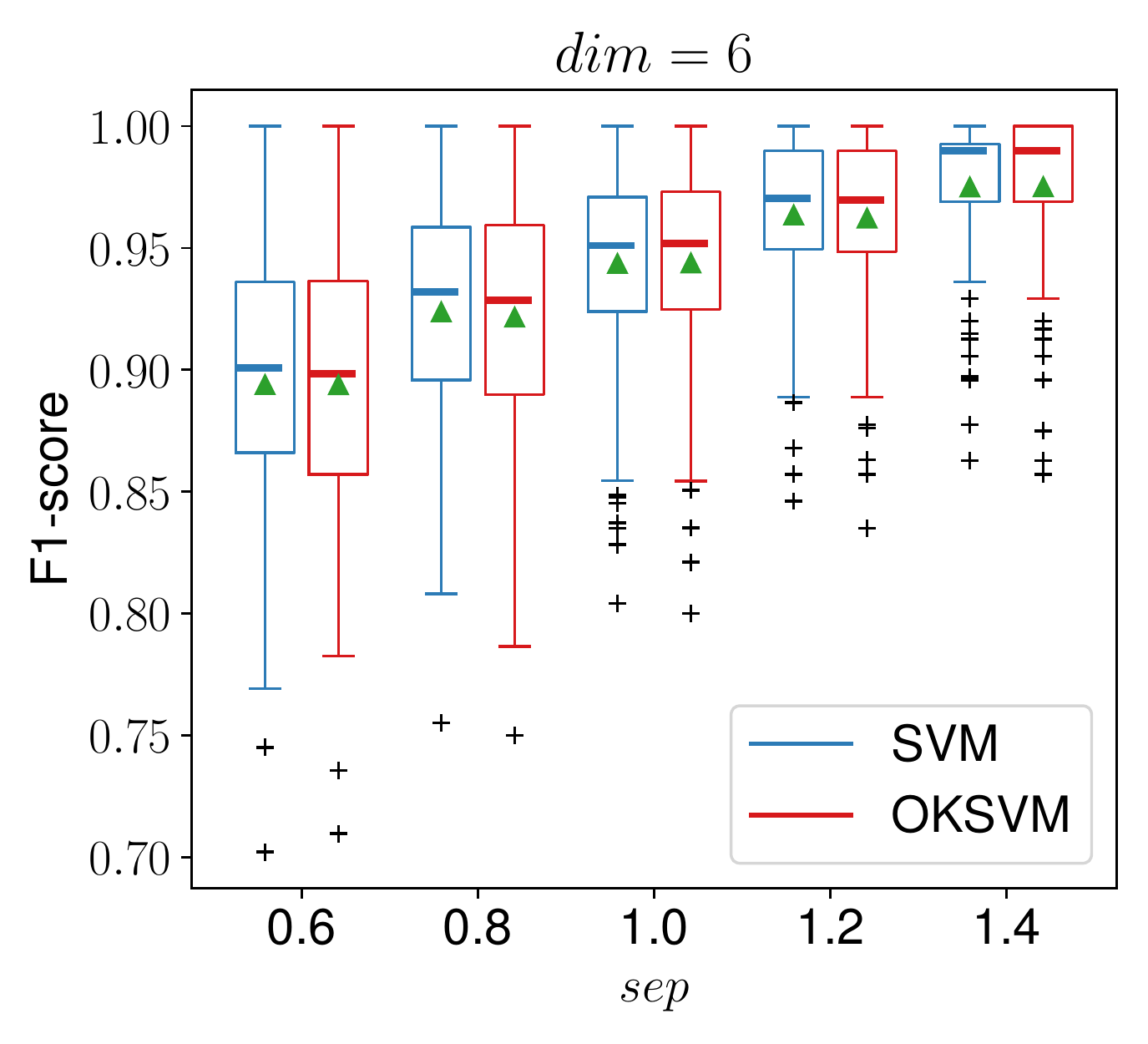}
\includegraphics[width=0.48\linewidth]{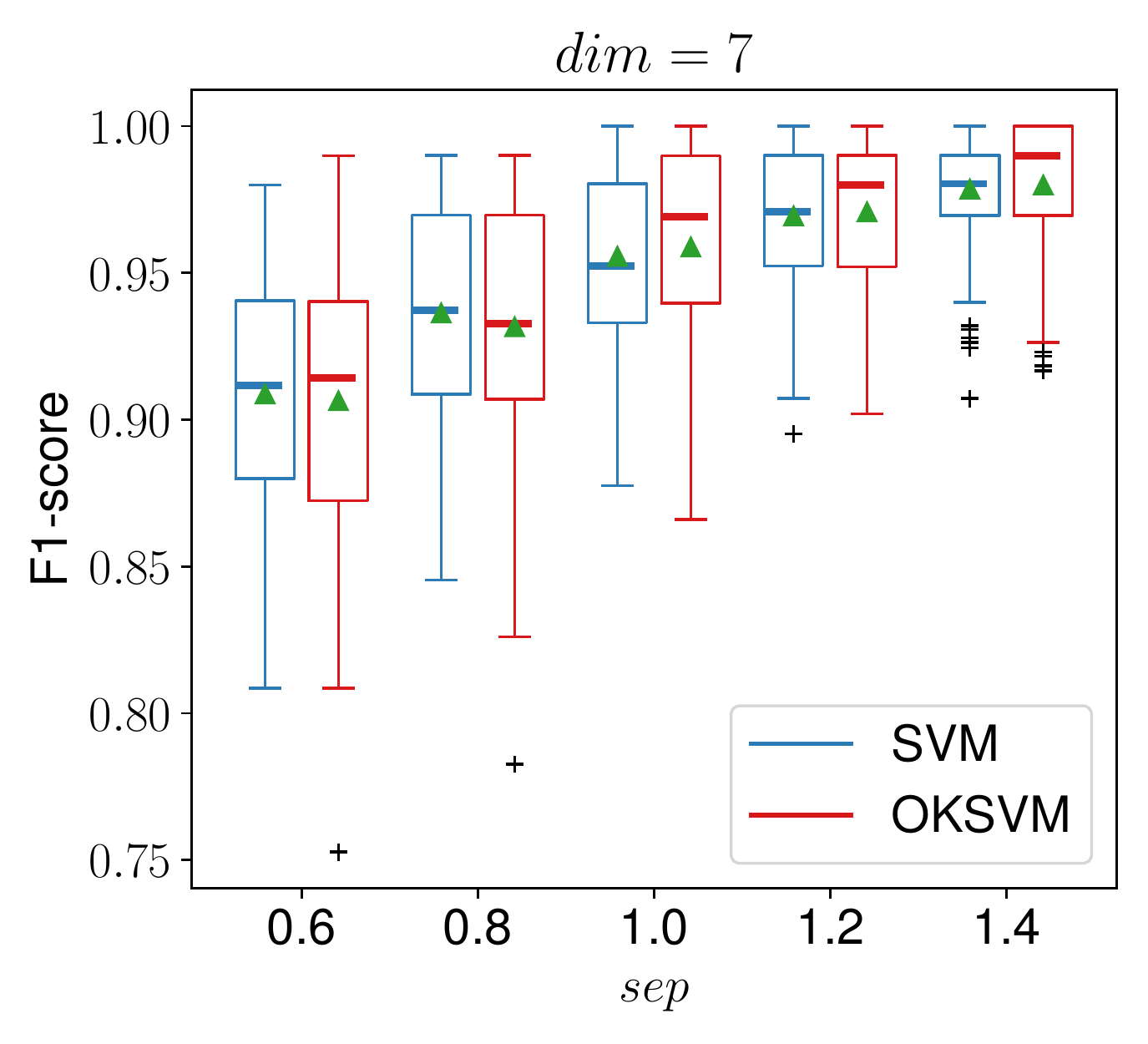}
\caption{Performance ($F1$-score) of \textit{OKSVM} and \textit{SVM}. Each subfigure represents the number of dimensions $dim$ of the datasets where each boxplot exhibits the performance of a method and they are organised by the separation $sep$ of the datasets.}
\label{fig:Boxplots_Fm_AutoC}
\end{figure}

Figure \ref{fig:Boxplots_Fm_AutoC} exhibits the performance in $F1$-score of \textit{OKSVM} and \textit{SVM} methods for dimensions 3, 5, 6 and 7. The results are very similar for both methods, although \textit{OKSVM} attains good mean and median results among the 100 runs for most of the tested configurations. It has to be pointed out that this kind of experiment runs in favor of the traditional SVMs since it is widely demonstrated their excellent efficacy when those parameters are correctly tuned for a specific dataset. Besides, as we are providing the fitted $C$ hyperparameter to the \textit{OKSVM} method, and this algorithm aims to optimize the kernel, there are more probabilities of provoking over-fitting. This effect can be seen in the dispersion represented by the boxplots.

\newpage

\subsection*{Real datasets}

Several real datasets have been selected to perform experiments for a real situation, as it is explained in the paper. Here are reported the remaining results.

\begin{figure}
\centering
	\begin{tabular}{cc}
		\includegraphics[width=0.48\linewidth]{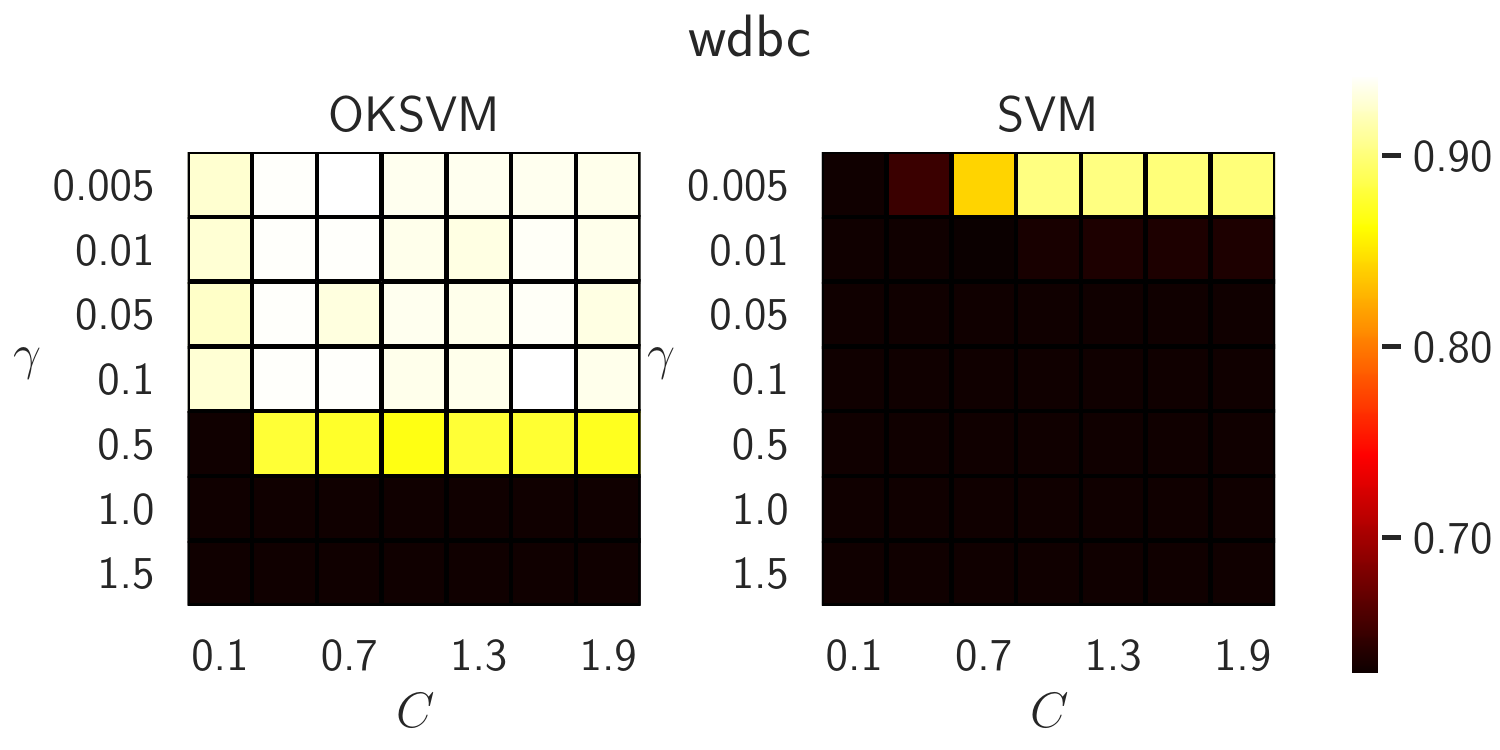} &
		\includegraphics[width=0.48\linewidth]{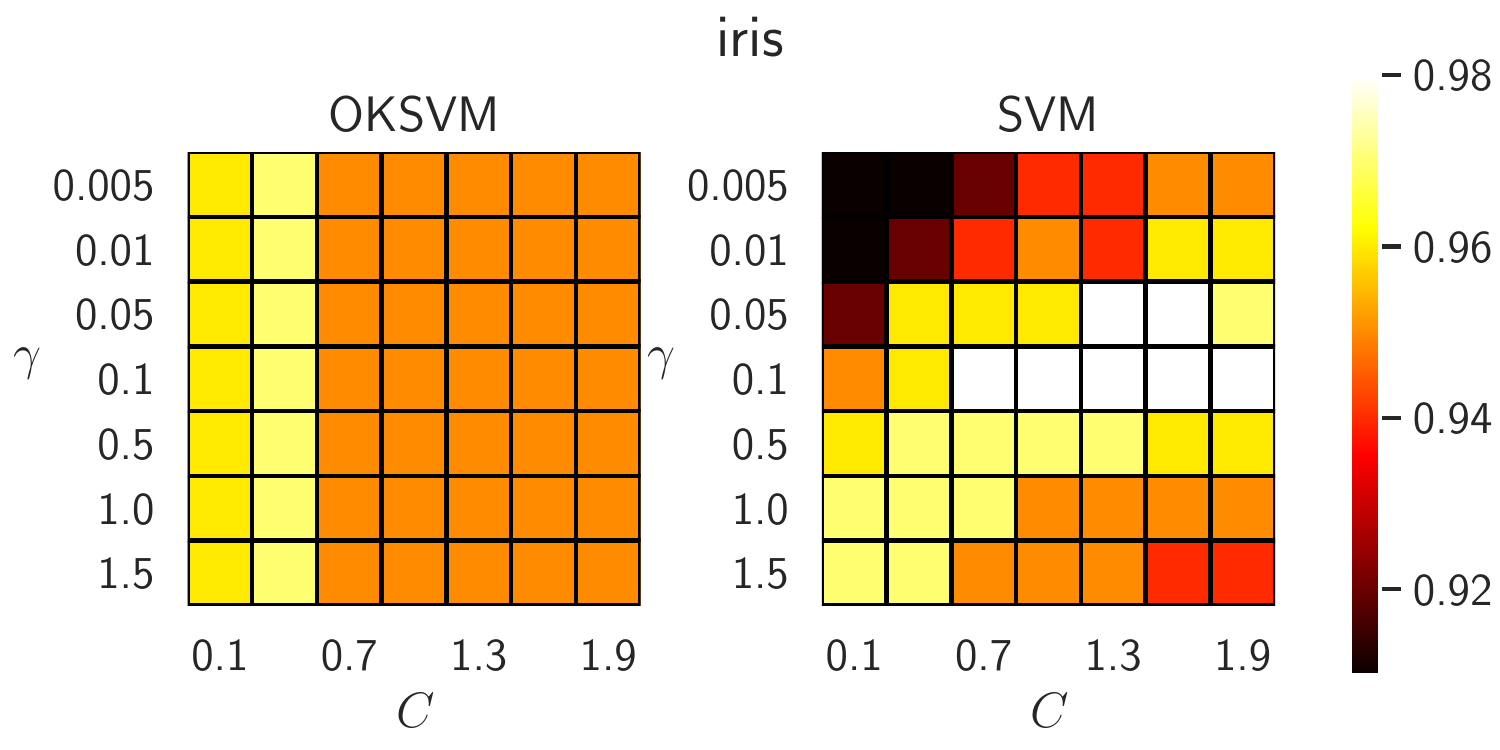} \\
	\end{tabular}
\caption{Performance (\textit{F1}) of \textit{OKSVM} and \textit{SVM}. Each heatmap exhibits the performance of the tuned configuration where each row of the heatmap represents the parameter $gamma$ and columns represent the parameter $C$. Darker is better.}
\label{fig:heatmap_Fm_comparison_realData}
\end{figure}

Datasets were split into stratified training and testing sets, where 80\% of the samples go for training and the remaining 20\% for testing. After five runs varying the hyperparameter configuration, the average results on the testing set are reported in \figurename\,\ref{fig:heatmap_Fm_comparison_realData}. 

In the case of \textit{wdbc} dataset, there is a high sensibility on the $\gamma$ parameter, more than $C$. Thus, the precise election of $\gamma$ is essential in order to generalize the model. Nevertheless, this choice is wider in the case of \textit{OKSVM}, where \textit{SVM} method only worked well with $\gamma=0.005$. 

The \textit{iris} dataset contains three balanced classes, although \textit{virginica} and \textit{versicolor} are the only ones that are not linearly separable. RBF kernels are oriented for complex data distribution (non-linear), so that is the reason why they have been chosen for the binary classification task. The output heatmaps reveal that \textit{SVM} method can achieve better results in the classification (center part of the map), although the rest of the configurations yield quite worse results. Then, the choice of the hyperparameters is important here too. \textit{OKSVM} does not overcome those best outcomes, although it is more stable. 

\newpage

\tablename s\,\ref{tab:real_data_cv} and \ref{tab:real_data_cv2}, additionally to the average measures, they include the standard deviations obtained in the 5-fold cross validation procedure. In \tablename\,\ref{tab:real_data_cv} is shown that \textit{OKSVM} outperforms \textit{SVM} method in Accuracy for most of the datasets, and the established difference is considerable. Nevertheless, the deviations are equal or higher too, which are in concordance with the analysis depicted in \figurename\,\ref{fig:Boxplots_Fm_AutoC} where it is shown that there is a bit of more dispersion in the case of \textit{OKSVM} method.

\setlength\tabcolsep{6pt}
\begin{table}
    \caption{Performance comparisons of \textit{OKSVM} and \textit{SVM} with real datasets. The class distribution and the average measures of 5-fold cross validation are shown. Best results are marked in \textbf{bold}.}
    \label{tab:real_data_cv}
    \centering
    \scalebox{0.75}{
    \begin{tabular}{l|cc|ccc|ccc}
        \toprule
        Method & \multicolumn{2}{c|}{Classes} & \multicolumn{3}{c}{\textit{SVM}} & \multicolumn{3}{c}{\textit{OKSVM}}\\
        \midrule
        Dataset & $N_-$ & $N_+$ & \textit{Acc} & \textit{RC} & \textit{PR} & \textit{Acc} & \textit{RC} & \textit{PR}\\
        \midrule
        \textit{cleveland} & 160 & 137 & 0.623$\pm$0.044 & 0.579$\pm$0.027 & 0.603$\pm$0.055 & \textbf{0.687$\pm$0.039} & \textbf{0.643$\pm$0.060} & \textbf{0.675$\pm$0.056}\\
        \textit{wdbc} & 357 & 212 & 0.895$\pm$0.034 & \textbf{0.924$\pm$0.041} & 0.819$\pm$0.058 & \textbf{0.928$\pm$0.014} & 0.900$\pm$0.059 & \textbf{0.908$\pm$0.041}\\
        \textit{breast-cancer} & 444 & 239 & \textbf{0.965$\pm$0.005} & \textbf{0.975$\pm$0.020} & \textbf{0.929$\pm$0.018} & 0.962$\pm$0.007 & \textbf{0.975$\pm$0.020} & 0.922$\pm$0.025\\
        \textit{iris} & 50 & 50 & 0.900$\pm$0.032 & 0.920$\pm$0.040 & 0.894$\pm$0.073 & \textbf{0.930$\pm$0.040} & \textbf{0.940$\pm$0.049} & \textbf{0.927$\pm$0.065}\\
        \textit{banknote} & 761 & 610 & 0.992$\pm$0.004 & \textbf{1.000$\pm$0.000} & 0.982$\pm$0.008 & \textbf{1.000$\pm$0.000} & \textbf{1.000$\pm$0.000} & \textbf{1.000$\pm$0.000}\\
        \textit{haberman} & 81 & 224 & 0.725$\pm$0.022 & 0.916$\pm$0.026 & \textbf{0.760$\pm$0.013} & \textbf{0.728$\pm$0.013} & \textbf{0.982$\pm$0.017} & 0.737$\pm$0.006\\
        \textit{winequality-red} & 744 & 855 & 0.643$\pm$0.015 & \textbf{0.995$\pm$0.004} & 0.600$\pm$0.010 & \textbf{0.654$\pm$0.016} & 0.970$\pm$0.029 & \textbf{0.611$\pm$0.013}\\
        \bottomrule
    \end{tabular}
    }
    
\end{table}

The average and standard deviations of the $F1$-score and $AUC$ are summarized in \tablename\,\ref{tab:real_data_cv2}. In this case, \textit{OKSVM} yielded better results for the $F1$ measure with lower dispersion. However, although the behavior of $AUC$ is similar in some cases, for others where the $F1$ is good, the $AUC$ is worse is average but better in standard deviation. This difference is notable for the \textit{haberman} dataset and may be caused by the high level of class imbalance (proportion 3-1).

\begin{table}
    \caption{Performance comparisons of \textit{OKSVM} and \textit{SVM} with real datasets. The class distribution and the average measures of 5-fold cross validation are shown. Best results are marked in \textbf{bold}.}
    \label{tab:real_data_cv2}
    \centering
    \scalebox{0.8}{
    \begin{tabular}{l|cc|cc|cc}
        \toprule
        Method & \multicolumn{2}{c|}{Classes} & \multicolumn{2}{c}{\textit{SVM}} & \multicolumn{2}{c}{\textit{OKSVM}}\\
        \midrule
        Dataset & $N_-$ & $N_+$ & \textit{F1} & \textit{AUC} & \textit{F1} & \textit{AUC}\\
        \midrule
        \textit{cleveland} & 160 & 137 & 0.590$\pm$0.037 & 0.673$\pm$0.063 & \textbf{0.656$\pm$0.044} & \textbf{0.736$\pm$0.055}\\
        \textit{wdbc} & 357 & 212 & 0.867$\pm$0.041 & 0.940$\pm$0.017 & \textbf{0.902$\pm$0.022} & \textbf{0.969$\pm$0.022}\\
        \textit{breast-cancer} & 444 & 239 & \textbf{0.951$\pm$0.008} & \textbf{0.993$\pm$0.003} & 0.947$\pm$0.009 & 0.988$\pm$0.006\\
        \textit{iris} & 50 & 50 & 0.903$\pm$0.025 & 0.968$\pm$0.013 & \textbf{0.931$\pm$0.039} & \textbf{0.986$\pm$0.010}\\
        \textit{banknote} & 761 & 610 & 0.991$\pm$0.004 & \textbf{1.000$\pm$0.000} & \textbf{1.000$\pm$0.000} & \textbf{1.000$\pm$0.000}\\
        \textit{haberman} & 81 & 224 & 0.831$\pm$0.014 & \textbf{0.624$\pm$0.056} & \textbf{0.842$\pm$0.008} & 0.598$\pm$0.042\\
        \textit{winequality-red} & 744 & 855 & 0.749$\pm$0.008 & \textbf{0.788$\pm$0.026} & \textbf{0.750$\pm$0.009} & 0.785$\pm$0.022\\
        \bottomrule
    \end{tabular}
    }
\end{table}